\documentclass[twoside,11pt]{article}
\usepackage{amsmath}
\usepackage{amssymb}
\usepackage{epsfig}
\usepackage{float}
\usepackage{graphics}
\usepackage{jair}
\usepackage{theapa}

\newcommand{\commentout}[1]{}

\newtheorem{theorem}{Theorem}
\newtheorem{corollary}{Corollary}

\newtheorem{proposition}{Proposition}

\newtheorem{definition}{Definition}

\newtheorem{example}{Example}
\newcommand{\qed}{$\square$}

\newcommand{\ba}{{\bf a}}
\newcommand{\bA}{{\bf A}}

\newcommand{\bw}{{\bf w}}
\newcommand{\bwbar}{\overline{\bw}}
\newcommand{\bwhat}{\widehat{\bw}}
\newcommand{\bwstar}{\bw^\ast}
\newcommand{\bwtilde}{\widetilde{\bw}}

\newcommand{\bx}{{\bf x}}
\newcommand{\bX}{{\bf X}}
\newcommand{\by}{{\bf y}}

\newcommand{\bz}{{\bf z}}
\newcommand{\bZ}{{\bf Z}}

\newcommand{\cA}{\mathcal{A}}
\newcommand{\cC}{\mathcal{C}}
\newcommand{\cL}{\mathcal{L}}
\newcommand{\cM}{\mathcal{M}}
\newcommand{\cN}{\mathcal{N}}
\newcommand{\cO}{\mathcal{O}}
\newcommand{\cS}{\mathcal{S}}
\newcommand{\cT}{\mathcal{T}}
\newcommand{\eps}{\varepsilon}

\newcommand{\pistar}{\pi^\ast}

\newcommand{\Vbw}{V^\bw}
\newcommand{\Vbwbar}{V^{\bwbar}}
\newcommand{\Vbwhat}{V^{\bwhat}}
\newcommand{\Vbwstar}{V^{\bwstar}}
\newcommand{\Vbwtilde}{V^{\bwtilde}}
\newcommand{\Vstar}{V^\ast}
\newcommand{\what}{\widehat{w}}

\newcommand{\Real}{\mathbb{R}}

\newcommand{\betapdf}{P_{\mathrm{beta}}}
\newcommand{\betacdf}{F_{\mathrm{beta}}}

\newcommand{\normalpdf}{\cN}

\newcommand{\unifpdf}[2]{\mathrm{U}_{[#1, #2]}}

\newcommand{\Dom}{\mathrm{Dom}}
\newcommand{\Parents}{\mathsf{Par}}

\newcommand{\lyapunov}{L}
\newcommand{\lyapunovfactor}{\kappa}

\newcommand{\abs}[1]{\left|#1\right|}
\newcommand{\ceils}[1]{\left\lceil#1\right\rceil}
\newcommand{\E}[2]{\mathrm{E}_{#1} \! \left[#2\right]}
\newcommand{\Eabs}[2]{\mathrm{E}_{#1} \! \abs{#2}}

\newcommand{\I}[2]{\mathbf{1}_{#1}(#2)}
\newcommand{\intin}[2]{\int_{#1} \! \! \! #2 \ud #1}
\newcommand{\maxnorm}[1]{\left\|#1\right\|_\infty}
\newcommand{\maxnormw}[2]{\left\|#1\right\|_{\infty, #2}}
\newcommand{\mode}[1]{\widehat{#1}}

\newcommand{\normw}[2]{\left\|#1\right\|_{#2}}
\newcommand{\set}[1]{\left\{#1\right\}}

\newcommand{\transpose}{^\mathsf{\scriptscriptstyle T}}
\newcommand{\ud}{\, \mathrm{d}}

\jairheading{27}{2006}{153--201}{05/06}{10/06}

\ShortHeadings{Solving Factored MDPs with Hybrid State and Action
Variables}{Kveton, Hauskrecht, \& Guestrin}

\firstpageno{153}

\begin{document}

\title{\vspace{-0.12in} Solving Factored MDPs with Hybrid State and Action \linebreak Variables}

\author{\name Branislav Kveton \email bkveton@cs.pitt.edu \\
\addr Intelligent Systems Program \\
5406 Sennott Square \\
University of Pittsburgh \\
Pittsburgh, PA 15260 \AND
\name Milos Hauskrecht \email milos@cs.pitt.edu \\
\addr Department of Computer Science \\
5329 Sennott Square \\
University of Pittsburgh \\
Pittsburgh, PA 15260 \AND
\name Carlos Guestrin \email guestrin@cs.cmu.edu \\
\addr Machine Learning Department \emph{and} \\
Computer Science Department \\
5313 Wean Hall \\
Carnegie Mellon University \\
Pittsburgh, PA 15213}

\maketitle

\begin{abstract}
Efficient representations and solutions for large decision
problems with continuous and discrete variables are among the most
important challenges faced by the designers of automated decision
support systems. In this paper, we describe a novel hybrid
factored Markov decision process (MDP) model that allows for a
compact representation of these problems, and a new hybrid
approximate linear programming (HALP) framework that permits their
efficient solutions. The central idea of HALP is to approximate
the optimal value function by a linear combination of basis
functions and optimize its weights by linear programming. We
analyze both theoretical and computational aspects of this
approach, and demonstrate its scale-up potential on several hybrid
optimization problems.
\end{abstract}

\section{Introduction}
\label{sec:introduction}

A dynamic decision problem with components of uncertainty can be
very often \mbox{formulated as} a Markov decision process (MDP). An
MDP represents a controlled stochastic process whose dynamics is
described by state transitions. Objectives of the control are
modeled by rewards (or costs), which are assigned to state-action
configurations. In the simplest form, the states and actions of an
MDP are discrete and unstructured. These models can be solved
efficiently by standard dynamic programming methods
\shortcite{bellman57dynamic,puterman94markov,bertsekas96neurodynamic}.

Unfortunately, textbook models rarely meet the practice and its
needs. First, real-world decision problems are naturally described
in a factored form and may involve a combination of discrete and
continuous variables. Second, there are no guarantees that compact
forms of the optimal value function or policy for these problems
exist. Therefore, hybrid optimization problems are usually
discretized and solved approximately by the methods for
discrete-state MDPs. The contribution of this work is a
principled, sound, and efficient approach to solving large-scale
factored MDPs that avoids this discretization step.

Our framework is based on approximate linear programming (ALP)
\shortcite{schweitzer85generalized}, which has been already
applied to solve decision problems with discrete state and action
variables efficiently
\shortcite{schuurmans02direct,defarias03linear,guestrin03efficient}.
These applications include context-specific planning
\shortcite{guestrin02context}, multiagent planning
\shortcite{guestrin02multiagent}, relational MDPs
\shortcite{guestrin03generalizing}, and first-order MDPs
\shortcite{sanner05approximate}. In this work, we show how to
adapt ALP to solving large-scale factored MDPs in hybrid state and
action spaces.

The presented approach combines factored MDP representations
(Sections \ref{sec:factored MDPs} and \ref{sec:HMDPs}) and
optimization techniques for solving large-scale structured linear
programs (Section \ref{sec:HALP constraint space}). This leads to
various benefits. First, the quality and complexity of value
function approximations is controlled by using basis functions
(Section \ref{sec:solving factored MDPs}). Therefore, we can
prevent an exponential blowup in the complexity of computations
when other techniques cannot. Second, we always guarantee that
HALP returns a solution. Its quality naturally depends on the
choice of basis functions. As analyzed in Section \ref{sec:HALP
error bounds}, if these are selected appropriately, we achieve a
close approximation to the optimal value function $\Vstar$. Third,
a well-chosen class of basis functions yields closed-form
solutions to the backprojections of our value functions (Section
\ref{sec:HALP expectation terms}). This step is important for
solving hybrid optimization problems more efficiently. Finally,
solving hybrid factored MDPs reduces to building and satisfying
relaxed formulations of the original problem (Section
\ref{sec:HALP constraint space}). The formulations can be solved
efficiently by the cutting plane method, which has been studied
extensively in applied mathematics and operations research.

For better readability of the paper, our proofs are deferred to
Appendix \ref{sec:proofs}. The following notation is adopted
throughout the work. Sets and their members are represented by
capital and small italic letters as $\cS$ and $s$, respectively.
Sets of variables, their subsets, and members of these sets are
denoted by capital letters as $\bX$, $\bX_i$, and $X_i$. In
general, corresponding small letters represent value assignments
to these objects. The subscripted indices $D$ and $C$ denote the
discrete and continuous variables in a variable set and its value
assignment. The function $\Dom(\cdot)$ computes the domain of a
variable or the domain of a function. The function
$\Parents(\cdot)$ returns the parent set of a variable in a
graphical model \shortcite{howard84influence,dean89model}.

\section{Markov Decision Processes}
\label{sec:MDPs}

Markov decision processes \shortcite{bellman57dynamic} provide an
elegant mathematical framework for modeling and solving sequential
decision problems in the presence of uncertainty. Formally, a
\emph{finite-state Markov decision process (MDP)} is given by a
4-tuple $\cM = (\cS, \cA, P, R)$, where $\cS = \set{s_1, \dots,
s_n}$ is a set of states, $\cA = \set{a_1, \dots, a_m}$ is a set
of actions, $P: \cS \times \cA \times \cS \rightarrow [0, 1]$ is a
stochastic transition function of state dynamics conditioned on
the preceding state and action, and $R: \cS \times \cA \rightarrow
\Real$ is a reward function assigning immediate payoffs to
state-action configurations. Without loss of generality, the
reward function is assumed to be nonnegative and bounded from
above by a constant $R_\mathrm{max}$ \shortcite{puterman94markov}.
Moreover, we assume that the transition and reward models are
stationary and known a priori.

Once a decision problem is formulated as an MDP, the goal is to
find a policy $\pi: \cS \rightarrow \cA$ that maximizes some
objective function. In this paper, the quality of a policy $\pi$
is measured by the \emph{infinite horizon discounted reward}:
\begin{align}
  \E{\pi}{\left.\sum_{t = 0}^\infty
  \gamma^t R(s^{(t)}, \pi(s^{(t)})) \right| s^{(0)} \sim \varphi},
  \label{eq:infinite horizon discounted reward}
\end{align}
where $\gamma \in [0, 1)$ is a \emph{discount factor}, $s^{(t)}$
is the state at the time step $t$, and the \mbox{expectation is}
taken with respect to all state-action trajectories that start in
the states $s^{(0)}$ and follow the policy $\pi$ thereafter. The
states $s^{(0)}$ are chosen according to a distribution $\varphi$.
This optimality criterion assures that there exists an
\emph{optimal policy} $\pistar$ which is stationary and
deterministic \shortcite{puterman94markov}. The policy is greedy
with respect to the \emph{optimal value function} $\Vstar$, which
is a fixed point of the Bellman equation
\shortcite{bellman57dynamic}:
\begin{align}
  \Vstar(s) = \max_a \left[R(s, a) +
  \gamma \sum_{s'} P(s' \mid s, a) \Vstar(s')\right].
  \label{eq:Bellman equation}
\end{align}
The Bellman equation plays a fundamental role in all dynamic
programming (DP) methods for solving MDPs
\shortcite{puterman94markov,bertsekas96neurodynamic}, including
value iteration, policy iteration, and linear programming. The
focus of this paper is on linear programming methods and their
refinements. Briefly, it is well known that the optimal value
function $\Vstar$ is a solution to the \emph{linear programming
(LP)} formulation \shortcite{manne60linear}:
\begin{align}
  \textrm{minimize} & \quad
  \sum_s \psi(s) V(s) \label{eq:LP} \\
  \textrm{subject to:} & \quad
  V(s) \geq R(s, a) + \gamma \sum_{s'} P(s' \mid s, a) V(s')
  \quad \forall \ s \in \cS, a \in \cA; \nonumber
\end{align}
where $V(s)$ represents the variables in the LP, one for each
state $s$, and $\psi(s) > 0$ is a strictly positive weighting on
the state space $\cS$. The number of constraints equals to the
cardinality of the cross product of the state and action spaces
$\abs{\cS \times \cA}$.

Linear programming and its efficient solutions have been studied
extensively in applied mathematics and operations research
\shortcite{bertsimas97introduction}. The simplex algorithm is a
common way of solving LPs. Its worst-case time complexity is
exponential in the number of variables. The ellipsoid method
\shortcite{khachiyan79polynomial} offers polynomial time
guarantees but it is impractical for solving LPs of even moderate
size.

The LP formulation (\ref{eq:LP}) can be solved compactly by the
\emph{cutting plane method} \shortcite{bertsimas97introduction} if
its objective function and constraint space are structured.
Briefly, this method searches for violated constraints in relaxed
formulations of the original LP. In every step, we start with a
relaxed solution $V^{(t)}$, find a violated constraint given
$V^{(t)}$, add it to the LP, and resolve for a new vector $V^{(t +
1)}$. The method is iterated until no violated constraint is
found, so that $V^{(t)}$ is an optimal solution to the LP. The
approach has a potential to solve large structured linear programs
if we can identify violated constraints efficiently
\shortcite{bertsimas97introduction}. The violated constraint and
the method that found it are often referred to as a
\emph{separating hyperplane} and a \emph{separation oracle},
respectively.

Delayed column generation is based on a similar idea as the
cutting plane method, which is applied to the column space of
variables instead of the row space of constraints. Bender's and
Dantzig-Wolfe decompositions reflect the structure in the
constraint space and are often used for solving large structured
linear programs.

\section{Discrete-State Factored MDPs}
\label{sec:factored MDPs}

Many real-world decision problems are naturally described in a
factored form. Discrete-state factored MDPs
\shortcite{boutilier95exploiting} allow for a compact
representation of this structure.

\subsection{Factored Transition and Reward Models}
\label{sec:factored representation}

A \emph{discrete-state factored MDP}
\shortcite{boutilier95exploiting} is a 4-tuple $\cM = (\bX, \cA,
P, R)$, where $\bX = \set{X_1, \dots, X_n}$ is a state space
described by a set of state variables, $\cA = \set{a_1, \dots,
a_m}$ is a set of actions\footnote{For simplicity of exposition,
we discuss a simpler model, which assumes a single action variable
$\cA$ instead of the factored action space $\bA = \set{A_1, \dots,
A_m}$. Our conclusions in Sections \ref{sec:factored
representation} and \ref{sec:ALP} extend to MDPs with factored
action spaces \shortcite{guestrin02multiagent}.}, $P(\bX' \mid
\bX, \cA)$ is a stochastic transition model of state dynamics
conditioned on the preceding state and action, and $R$ is a reward
function assigning immediate payoffs to state-action
configurations. The state of the system is completely observed and
represented by a vector of value assignments $\bx = (x_1, \dots,
x_n)$. We assume that the values of every state variable $X_i$ are
restricted to a finite domain $\Dom(X_i)$.

\bigskip \noindent {\bf Transition model:} The transition model is
given by the conditional probability distribution $P(\bX' \mid
\bX, \cA)$, where $\bX$ and $\bX'$ denote the state variables at
two successive time steps. Since the complete tabular
representation of $P(\bX' \mid \bX, \cA)$ is infeasible, we assume
that the transition model factors along $\bX'$ as:
\begin{align}
  P(\bX' \mid \bX, a) =
  \prod_{i = 1}^n P(X_i' \mid \Parents(X_i'), a)
  \label{eq:transition model}
\end{align}
and can be described compactly by a \emph{dynamic Bayesian network
(DBN)} \shortcite{dean89model}. This DBN representation captures
independencies among the state variables $\bX$ and $\bX'$ given an
action $a$. One-step dynamics of every state variable is modeled
by its conditional probability distribution $P(X_i' \mid \!
\Parents(X_i'), a)$, where $\Parents(X_i') \! \subseteq \bX$
denotes the parent set of $X_i'$. Typically, the parent set is a
subset of state variables which simplifies the parameterization of
the model. In principle, the parent set can be extended to the
state variables $\bX'$. Such an extension poses only few new
challenges when solving the new problems efficiently
\shortcite{guestrin03thesis}. Therefore, we omit the discussion on
the modeling of intra-layer dependencies in this paper.

\bigskip \noindent {\bf Reward model:} The reward model factors
similarly to the transition model. In particular, the reward
function $R(\bx, a) = \sum_j R_j(\bx_j, a)$ is an additive
function of local reward functions defined on the subsets $\bX_j$
and $\cA$. In graphical models, the local functions can be
described compactly by reward nodes $R_j$, which are conditioned
on their parent sets $\Parents(R_j) = \bX_j \cup \cA$. To allow
this representation, we formally extend our DBN to an influence
diagram \shortcite{howard84influence}.

\begin{figure}[t]
  \centering
  \begin{tabular}{@{\!\!\!}c@{\!\!\!}c@{\!\!\!}c@{\!\!\!}}
    \includegraphics[width=1.71in, bb=0in 0in 12.361in 11.736in]{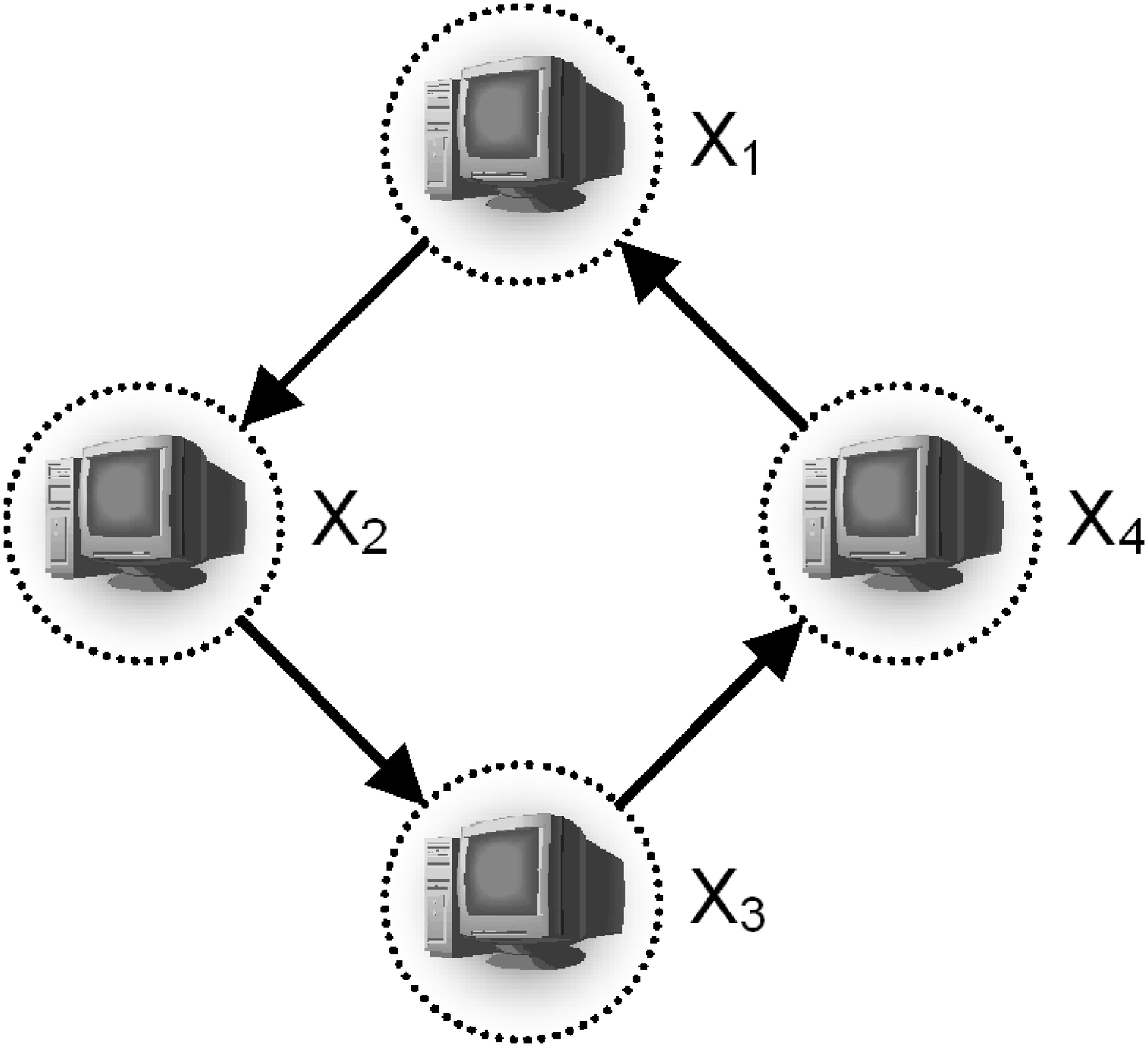} &
    \includegraphics[width=2.88in, bb=0in 0in 20.833in 11.736in]{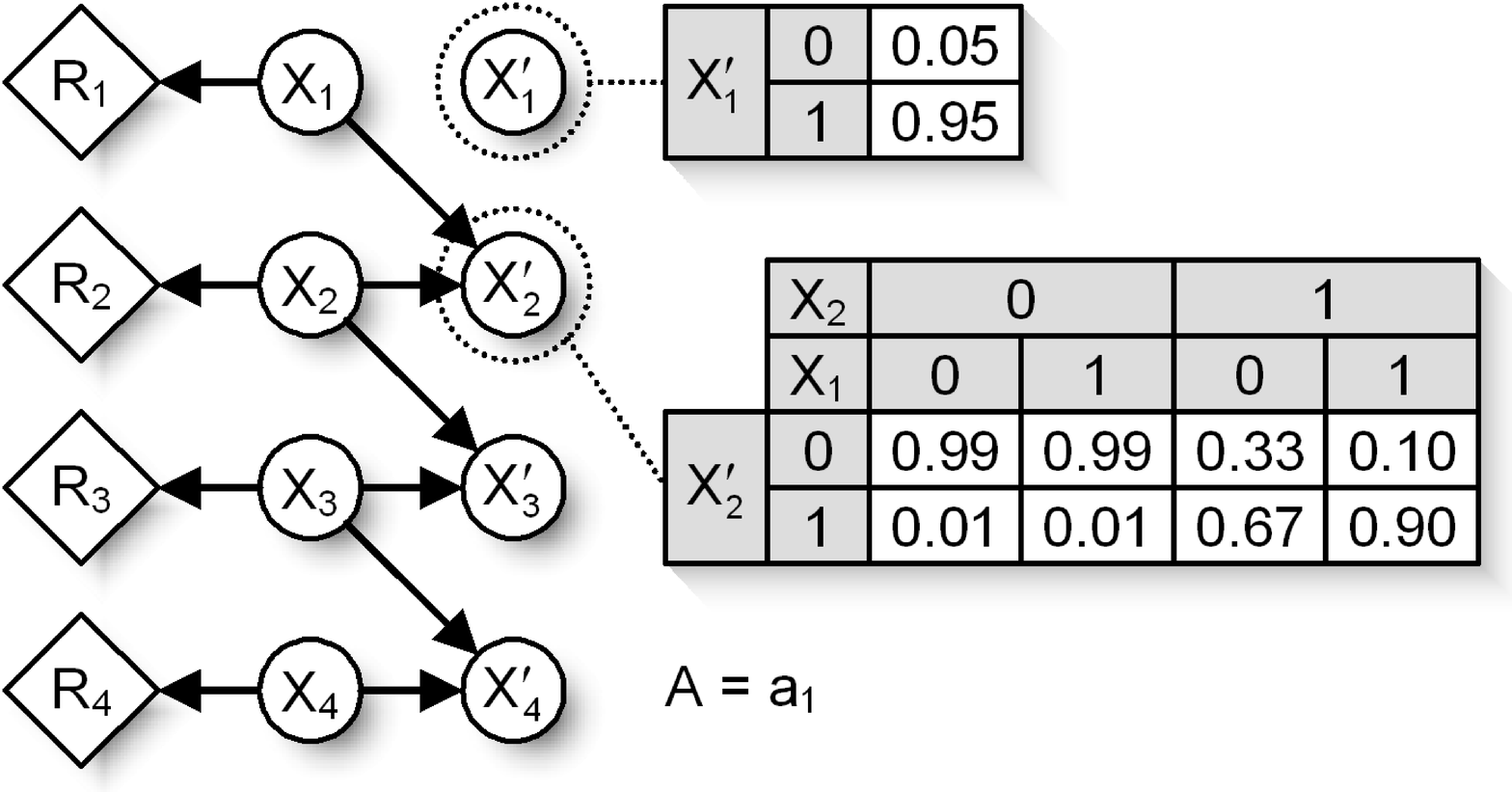} &
    \includegraphics[width=1.35in, bb=0in 0in 9.792in 11.736in]{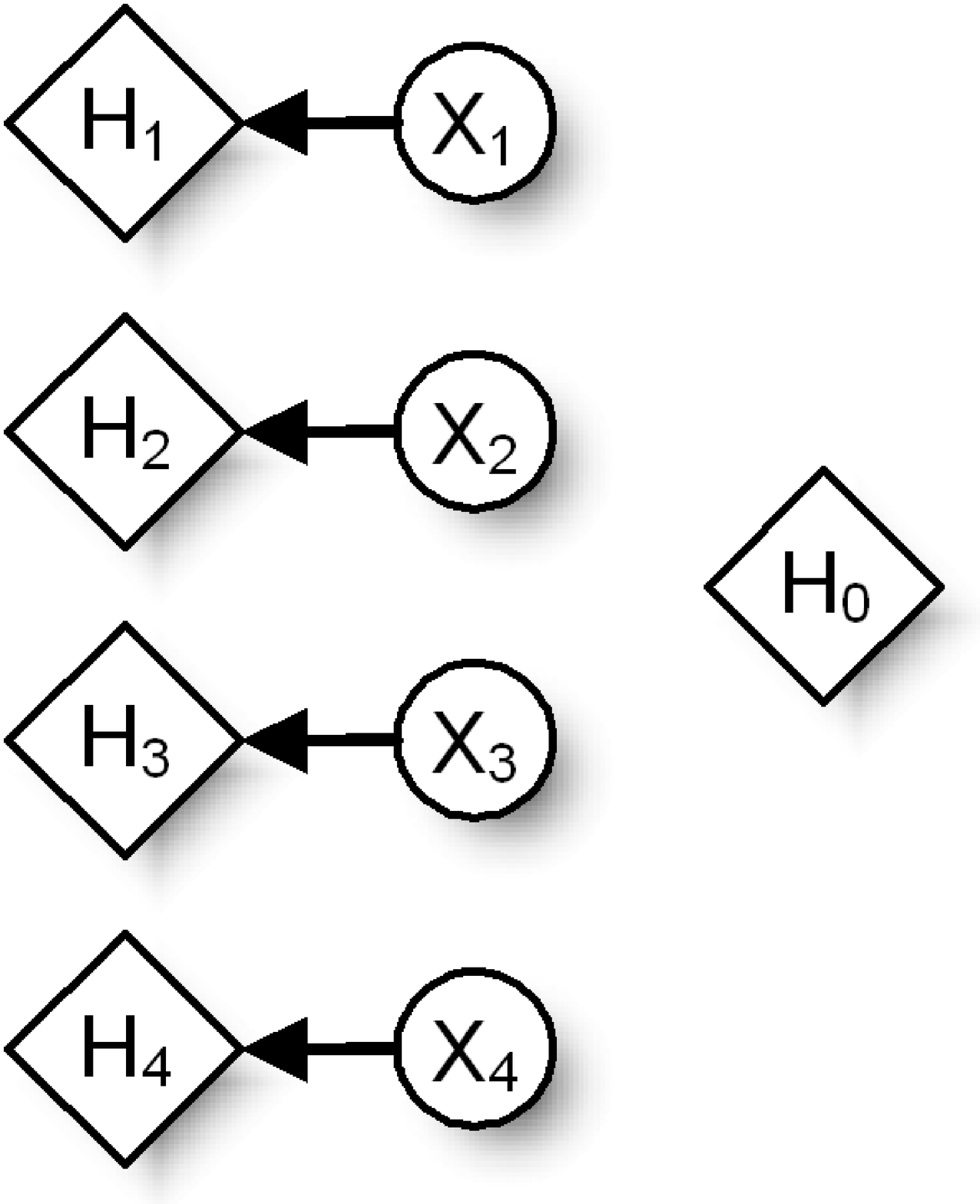} \\
    \textbf{(a)}$\quad$ &
    \textbf{(b)}$\qquad \quad \ \ $ &
    \textbf{(c)}$\ \ $
  \end{tabular}
  \caption{\textbf{a.} Four computers in a ring
  topology. Direction of propagating failures is denoted
  by arrows. \textbf{b.} A graphical representation of
  factored transition and reward models after taking an action
  $a_1$ in the 4-ring topology. The future state of the server
  $X_1'$ is independent of the rest of the network because the
  server is rebooted. Reward nodes $R_1$ and $R_j$ ($j \geq 2$)
  denote the components $2 x_1$ and $x_j$ ($j \geq 2$) of the
  reward model. \textbf{c.} A graphical representation of the
  linear value function approximation
  $\Vbw(\bx) = w_0 + \sum_{i = 1}^4 w_i x_i$ in the
  4-ring topology. Reward nodes $H_0$ and $H_i$ ($i \geq 1$)
  denote the value function components $w_0$ and $w_i x_i$
  ($i \geq 1$).}
  \label{fig:admin}
\end{figure}

\begin{example}[\shortciteR{guestrin01maxnorm}]
\label{ex:admin} To illustrate the concept of a factored MDP, we
consider a network administration problem, in which the computers
are unreliable and fail. The failures of these computers propagate
through network connections to the whole network. For instance, if
the server $X_1$ (Figure \ref{fig:admin}a) is down, the chance
that the neighboring computer $X_2$ crashes increases. The
administrator can prevent the propagation of the failures by
rebooting computers that have already crashed.

This network administration problem can be formulated as a
factored MDP. The state of the network is completely observable
and represented by $n$ binary variables $\bX = \set{X_1, \dots,
X_n}$, where the variable $X_i$ denotes the state of the $i$-th
computer: 0 (being down) or 1 (running). At each time step, the
administrator selects an action from the set $\cA = \set{a_1,
\dots, a_{n + 1}}$. The action $a_i$ ($i \leq n$) corresponds to
rebooting the $i$-th computer. The last action $a_{n + 1}$ is
dummy. The transition function reflects the propagation of
failures in the network and can be encoded locally by conditioning
on the parent set of every computer. A natural metric for
evaluating the performance of an administrator is the total number
of running computers. This metric factors along the computer
states $x_i$ and can be represented compactly by an additive
reward function:
\begin{align*}
  R(\bx, a) = 2 x_1 + \sum_{j = 2}^n x_j.
\end{align*}
The weighting of states establishes our preferences for
maintaining the server $X_1$ and workstations $X_2, \dots, X_n$.
An example of transition and reward models after taking an action
$a_1$ in the 4-ring topology (Figure \ref{fig:admin}a) is given in
Figure \ref{fig:admin}b.
\end{example}

\subsection{Solving Discrete-State Factored MDPs}
\label{sec:solving factored MDPs}

Markov decision processes can be solved by exact DP methods in
polynomial time in the size of the state space $\bX$
\shortcite{puterman94markov}. Unfortunately, factored state spaces
are exponential in the number of state variables. Therefore, the
DP methods are unsuitable for solving large factored MDPs. Since a
factored representation of an MDP (Section \ref{sec:factored
representation}) may not guarantee a structure in the optimal
value function or policy \shortcite{koller99computing}, we resort
to value function approximations to alleviate this concern.

Value function approximations have been successfully applied to a
variety of real-world domains, including backgammon
\shortcite{tesauro92practical,tesauro94tdgammon,tesauro95temporal},
elevator dispatching \shortcite{crites96improving}, and job-shop
scheduling
\shortcite{zhang95reinforcement,zhang96highperformance}. These
partial successes suggest that the approximate dynamic programming
is a powerful tool for solving large optimization problems.

In this work, we focus on \emph{linear value function
approximation} \shortcite{bellman63polynomial,vanroy98thesis}:
\begin{align}
  \Vbw(\bx) = \sum_i w_i f_i(\bx).
  \label{eq:linear value function}
\end{align}
The approximation restricts the form of the value function $\Vbw$
to the linear combination of $\abs{\bw}$ basis functions
$f_i(\bx)$, where $\bw$ is a vector of optimized weights. Every
basis function can be defined over the complete state space $\bX$,
but usually is limited to a small subset of state variables
$\bX_i$ \shortcite{bellman63polynomial,koller99computing}. The
role of basis functions is similar to features in machine
learning. They are often provided by domain experts, although
there is a growing amount of work on learning basis functions
automatically
\shortcite{patrascu02greedy,mahadevan05samuel,kveton06learning,mahadevan06value,mahadevan06learning}.

\begin{example}
\label{ex:admin value function} To demonstrate the concept of the
linear value function model, we consider the network
administration problem (Example \ref{ex:admin}) and assume a low
chance of a single computer failing. Then the value function in
Figure \ref{fig:admin}c is sufficient to derive a close-to-optimal
policy on the 4-ring topology (Figure \ref{fig:admin}a) because
the indicator functions $f_i(\bx) = x_i$ capture changes in the
states of individual computers. For instance, if the computer
$X_i$ fails, the linear policy:
\begin{align*}
  u(\bx) = \arg\max_a \left[R(\bx, a) +
  \gamma \sum_{\bx'} P(\bx' \mid \bx, a) \Vbw(\bx')\right]
\end{align*}
immediately leads to rebooting it. If the failure has already
propagated to the \mbox{computer $X_{i + 1}$}, the policy recovers
it in the next step. This procedure is repeated until the spread
of the initial failure is stopped.
\end{example}

\subsection{Approximate Linear Programming}
\label{sec:ALP}

Various methods for fitting of the linear value function
approximation have been proposed and analyzed
\shortcite{bertsekas96neurodynamic}. We focus on \emph{approximate
linear programming (ALP)} \shortcite{schweitzer85generalized},
which recasts this problem as a linear program:
\begin{align}
  \textrm{minimize}_\bw & \quad
  \sum_\bx \psi(\bx) \sum_i w_i f_i(\bx) \label{eq:ALP} \\
  \textrm{subject to:} & \quad
  \sum_i w_i f_i(\bx) \geq R(\bx, a) +
  \gamma \sum_{\bx'} P(\bx' \mid \bx, a) \sum_i w_i f_i(\bx')
  \quad \forall \ \bx \in \bX, a \in \cA; \nonumber
\end{align}
where $\bw$ represents the variables in the LP, $\psi(\bx) \geq 0$
are \emph{state relevance weights} weighting the quality of the
approximation, and $\gamma \sum_{\bx'} P(\bx' \mid \bx, a) \sum_i
w_i f_i(\bx')$ is a discounted \emph{backprojection} of the value
function $\Vbw$ (Equation \ref{eq:linear value function}). The ALP
formulation can be easily derived from the standard LP formulation
(\ref{eq:LP}) by substituting $\Vbw(\bx)$ for $V(\bx)$. The
formulation is feasible if the set of basis functions contains a
constant function $f_0(\bx) \equiv 1$. We assume that such a basis
function is always present. Note that the state relevance weights
are no longer enforced to be strictly positive (Section
\ref{sec:introduction}). Comparing to the standard LP formulation
(\ref{eq:LP}), which is solved by the optimal value function
$\Vstar$ for arbitrary weights $\psi(s) > 0$, a solution
$\bwtilde$ to the ALP formulation depends on the weights
$\psi(\bx)$. Intuitively, the higher the weights, the higher the
quality of the approximation $\Vbwtilde$ in a corresponding state.

Since our basis functions are usually restricted to subsets of
state variables (Section \ref{sec:solving factored MDPs}),
summation terms in the ALP formulation can be computed efficiently
\shortcite{guestrin01maxnorm,schuurmans02direct}. For example, the
order of summation in the backprojection term can be rearranged as
$\gamma \sum_i w_i \sum_{\bx_i'} P(\bx_i' \mid \bx, a)
f_i(\bx_i')$, which allows its aggregation in the space of $\bX_i$
instead of $\bX$. Similarly, a factored form of $\psi(\bx)$ yields
an efficiently computable objective function
\shortcite{guestrin03thesis}.

The number of constraints in the ALP formulation is exponential in
the number of state variables. Fortunately, the constraints are
structured. This results from combining factored transition and
reward models (Section \ref{sec:factored representation}) with the
linear approximation (Equation \ref{eq:linear value function}). As
a consequence, the constraints can be satisfied without
enumerating them exhaustively.

\begin{figure}[t]
  \centering
  \begin{tabular}{@{\!\!\!}c@{\!\!\!}c@{\!\!\!}}
    \includegraphics[width=3.06in, bb=0in 0in 22.153in 13.681in]{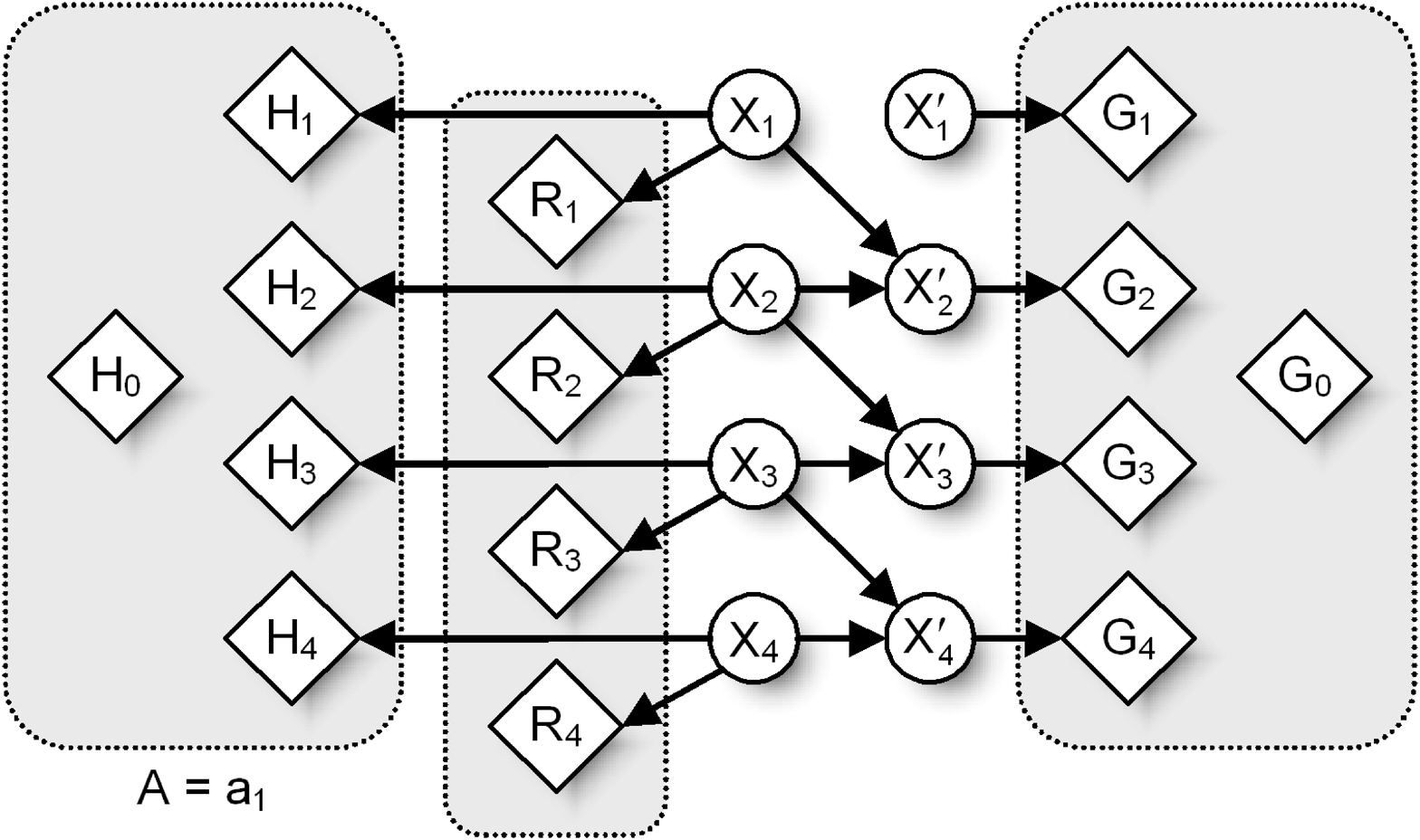} &
    \includegraphics[width=3.06in, bb=0in 0in 22.153in 13.681in]{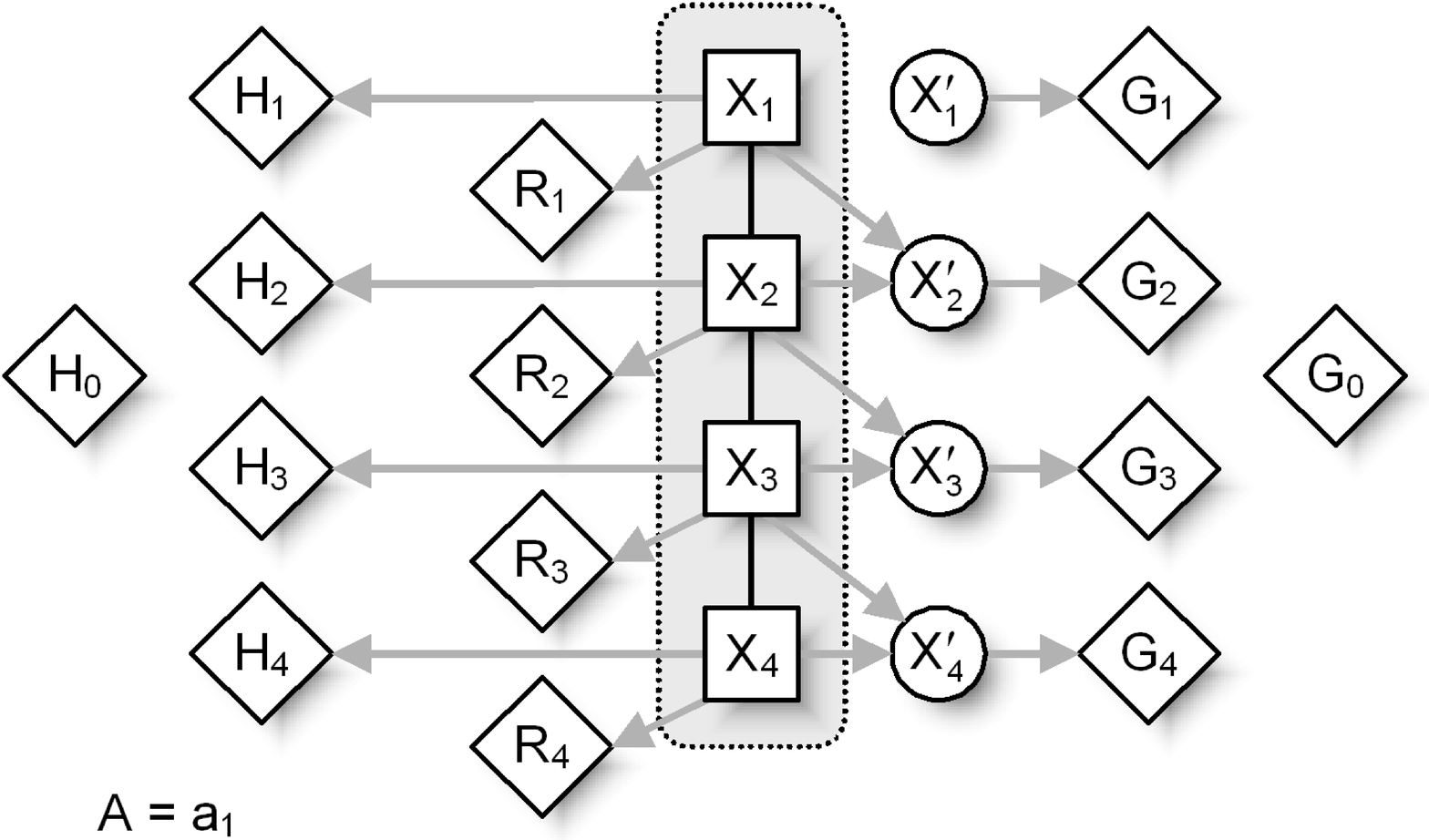} \\
    $\quad \ $\textbf{(a)} &
    $\quad \ $\textbf{(b)}
  \end{tabular}
  \caption{\textbf{a.} A graphical representation of combining
  factored transition and reward models (Figure \ref{fig:admin}b)
  with the linear approximation (Figure \ref{fig:admin}c).
  Reward nodes $G_0$ and $G_i$ ($i \geq 1$) represent the
  discounted backprojection terms $- \gamma w_0$ and
  $- \gamma w_i x_i'$ ($i \geq 1$). Gray regions are the
  cost components of the constraint space. \textbf{b.}
  A cost network corresponding to our factored constraint space
  (Figure \ref{fig:admin constraint space}a). The network captures
  pairwise dependencies $X_1 \! - \! X_2$, $X_2 \! - \! X_3$, and
  $X_3 \! - \! X_4$. The treewidth of the cost network is 1.}
  \label{fig:admin constraint space}
\end{figure}

\begin{example}
\label{ex:admin constraint space} The notion of a \emph{factored
constraint space} is important for compact satisfaction of
exponentially many constraints. To illustrate this concept, let us
consider the linear value function (Example \ref{ex:admin value
function}) on the 4-ring network administration problem (Example
\ref{ex:admin}). \hspace{-0.1in} Intuitively, by combining the
graphical representations of $P(\bx' \mid \bx, a_1)$, $R(\bx,
a_1)$ (Figure \ref{fig:admin}b), and $\Vbw(\bx)$ (Figure
\ref{fig:admin}c), we obtain a factored model of constraint
violations:
\begin{align*}
  \tau^\bw(\bx, a_1)
  \ = & \ \ \Vbw(\bx) -
  \gamma \sum_{\bx'} P(\bx' \mid \bx, a_1) \Vbw(\bx') -
  R(\bx, a_1) \\
  \ = & \ \ \sum_i w_i f_i(\bx) -
  \gamma \sum_i w_i \sum_{\bx_i'} P(\bx_i' \mid \bx, a_1) f_i(\bx_i') -
  R(\bx, a_1) \\
  \ = & \ \ w_0 + \sum_{i = 1}^4 w_i x_i -
  \gamma w_0 - \gamma w_1 P(x_1' = 1 \mid a_1) - \\
  & \ \ \gamma \sum_{i = 2}^4 w_i P(x_i' = 1 \mid x_i, x_{i - 1}, a_1) -
  2 x_1 - \sum_{j = 2}^4 x_j.
\end{align*}
for an arbitrary solution $\bw$ (Figure \ref{fig:admin constraint
space}a). Note that this cost function:
\begin{align*}
  \tau^\bw(\bx, a_1) = \phi^\bw + \sum_{i = 1}^4 \phi^\bw(x_i) +
  \sum_{i = 2}^4 \phi^\bw(x_i, x_{i - 1})
\end{align*}
is a linear combination of a constant $\phi^\bw$ in $\bx$, and
univariate and bivariate functions $\phi^\bw(x_i)$ and
$\phi^\bw(x_i, x_{i - 1})$. It can be represented compactly by a
\emph{cost network} \shortcite{guestrin01maxnorm}, which is an
undirected graph over a set of variables $\bX$. Two nodes in the
graph are connected if any of the cost terms depends on both
variables. Therefore, the cost network corresponding to the
function $\tau^\bw(\bx, a_1)$ must contain edges $X_1 \! - \!
X_2$, $X_2 \! - \! X_3$, and $X_3 \! - \! X_4$ (Figure
\ref{fig:admin constraint space}b).

Savings achieved by the compact representation of constraints are
related to the efficiency of computing $\arg\min_{\bx} \tau^\bw(\bx,
a_1)$ \shortcite{guestrin03thesis}. This computation can be done by
variable elimination and its complexity increases exponentially in
the width of the \emph{tree decomposition} of the cost network. The
smallest width of all tree decompositions is referred to as
\emph{treewidth}.
\end{example}

\noindent Inspired by the factorization,
\shortciteA{guestrin01maxnorm} proposed a variable-elimination
method \shortcite{dechter96bucket} that rewrites the constraint
space in ALP compactly. \shortciteA{schuurmans02direct} solved the
same problem by the cutting plane method. The method iteratively
searches for the most violated constraint:
\begin{align}
  \arg\min_{\bx, a} \left[\sum_i w^{(t)}_i \bigg[f_i(\bx_i) -
  \gamma \sum_{\bx_i'} P(\bx_i' \mid \bx, a) f_i(\bx_i')\bigg] -
  R(\bx, a)\right]
  \label{eq:ALP most violated constraint}
\end{align}
with respect to the solution $\bw^{(t)}$ of a relaxed ALP. The
constraint is added to the LP, which is resolved for a new
solution $\bw^{(t + 1)}$. This procedure is iterated until no
violated constraint is found, so that $\bw^{(t)}$ is an optimal
solution to the ALP.

The quality of the ALP formulation has been studied by
\shortciteA{defarias03linear}. Based on their work, we conclude
that ALP yields a close approximation $\Vbwtilde$ to the optimal
value function $\Vstar$ if the weighted max-norm error
$\maxnormw{\Vstar - \Vbw}{1 / \lyapunov}$ can be minimized. We
return to this theoretical result in Section \ref{sec:HALP error
bounds}.

\begin{theorem}[\shortciteR{defarias03linear}]
\label{thm:ALP bound} Let $\bwtilde$ be a solution to the ALP
formulation (\ref{eq:ALP}). Then the expected error of the value
function $\Vbwtilde$ can be bounded as:
\begin{align*}
  \normw{\Vstar - \Vbwtilde}{1, \psi} \leq
  \frac{2 \psi\transpose \lyapunov}{1 - \lyapunovfactor}
  \min_\bw \maxnormw{\Vstar - \Vbw}{1 / \lyapunov},
\end{align*}
where $\normw{\cdot}{1, \psi}$ is an $\cL_1$-norm weighted by the
state relevance weights $\psi$, $\lyapunov(\bx) \! = \! \sum_i
w_i^\lyapunov f_i(\bx)$ is a \emph{Lyapunov function} such that
the inequality $\lyapunovfactor \lyapunov(\bx) \geq \gamma
\sup_\ba \E{P(\bx' \mid \bx, \ba)}{\lyapunov(\bx')}$ holds,
$\lyapunovfactor \in [0, 1)$ denotes its \emph{contraction
factor}, and $\maxnormw{\cdot}{1 / \lyapunov}$ is a max-norm
reweighted by the reciprocal $1 / \lyapunov$.
\end{theorem}

\noindent Note that the $\cL_1$-norm distance $\normw{\Vstar -
\Vbwtilde}{1, \psi}$ equals to the expectation $\Eabs{\psi}{\Vstar
- \Vbwtilde}$ over the state space $\bX$ with respect to the state
relevance weights $\psi$. Similarly to Theorem \ref{thm:ALP
bound}, we utilize the $\cL_1$ and $\cL_\infty$ norms in the rest
of the work to measure the expected and worst-case errors of value
functions. These norms are defined as follows.

\begin{definition}
\label{def:norms} The \emph{$\cL_1$ (Manhattan)} and
\emph{$\cL_\infty$ (infinity) norms} are typically defined as
$\normw{f}{1} = \sum_\bx \abs{f(\bx)}$ and $\maxnorm{f} = \max_\bx
\abs{f(\bx)}$. If the state space $\bX$ is represented by both
discrete and continuous variables $\bX_D$ and $\bX_C$, the
definition of the norms changes accordingly:
\begin{align}
  \normw{f}{1} = \sum_{\bx_D} \intin{\bx_C}{\abs{f(\bx)}}
  \quad \emph{and} \quad
  \maxnorm{f} = \sup_\bx \abs{f(\bx)}.
  \label{eq:norms}
\end{align}
The following definitions:
\begin{align}
  \normw{f}{1, \psi} = \sum_{\bx_D} \intin{\bx_C}{\psi(\bx) \abs{f(\bx)}}
  \quad \emph{and} \quad
  \maxnormw{f}{\psi} = \sup_\bx \psi(\bx) \abs{f(\bx)}
  \label{eq:weighted norms}
\end{align}
correspond to the $\cL_1$ and $\cL_\infty$ norms reweighted by a
function $\psi(\bx)$.
\end{definition}

\section{Hybrid Factored MDPs}
\label{sec:HMDPs}

Discrete-state factored MDPs (Section \ref{sec:factored MDPs})
permit a compact representation of decision problems with discrete
states. However, real-world domains often involve continuous
quantities, such as temperature and pressure. A sufficient
discretization of these quantities may require hundreds of points in
a single dimension, which renders the representation of our
transition model (Equation \ref{eq:transition model}) infeasible. In
addition, rough and uninformative discretization impacts the quality
of policies. Therefore, we want to avoid discretization or defer it
until necessary. As a step in this direction, we discuss a formalism
for representing hybrid decision problems in the domains of discrete
and continuous variables.

\subsection{Factored Transition and Reward Models}
\label{sec:hybrid factored representation}

A \emph{hybrid factored MDP (HMDP)} is a 4-tuple $\cM = (\bX, \bA,
P, R)$, where $\bX = \set{X_1, \dots, X_n}$ is a state space
described by state variables, $\bA = \set{A_1, \dots, A_m}$ is an
action space described by action variables, $P(\bX' \mid \bX,
\bA)$ is a stochastic transition model of state dynamics
conditioned on the preceding state and action, and $R$ is a reward
function assigning immediate \mbox{payoffs to} state-action
configurations.\footnote{\emph{General state and action space MDP}
is an alternative term for a hybrid MDP. The term \emph{hybrid}
does not refer to the dynamics of the model, which is
discrete-time.}

\bigskip \noindent {\bf State variables:} State variables are
either discrete or continuous. Every discrete variable $X_i$ takes
on values from a finite domain $\Dom(X_i)$. Following
\shortciteA{hauskrecht04linear}, we assume that every continuous
variable is bounded to the $[0, 1]$ subspace. In general, this
assumption is very mild and permits modeling of any closed
interval on $\Real$. The state of the system is completely
observed and described by a vector of value assignments $\bx =
(\bx_D, \bx_C)$ which partitions along its discrete and continuous
components $\bx_D$ and $\bx_C$.

\bigskip \noindent {\bf Action variables:} The action space is
distributed and represented by action variables $\bA$. The
composite action is defined by a vector of individual action
choices $\ba = (\ba_D, \ba_C)$ which partitions along its discrete
and continuous components $\ba_D$ and $\ba_C$.

\bigskip \noindent {\bf Transition model:} The transition model is
given by the conditional probability distribution $P(\bX' \mid
\bX, \bA)$, where $\bX$ and $\bX'$ denote the state variables at
two successive time steps. We assume that this distribution
factors along $\bX'$ as $P(\bX' \mid \bX, \bA) = \prod_{i = 1}^n
P(X_i' \mid \Parents(X_i'))$ and can be described compactly by a
DBN \shortcite{dean89model}. Typically, the parent set
$\Parents(X_i') \subseteq \bX \cup \bA$ is a small subset of state
and action variables which allows for a local parameterization of
the transition model.

\begin{figure}[t]
  \centering
  \includegraphics[width=6.8in, bb=0.55in 4.65in 9.05in 6.65in]{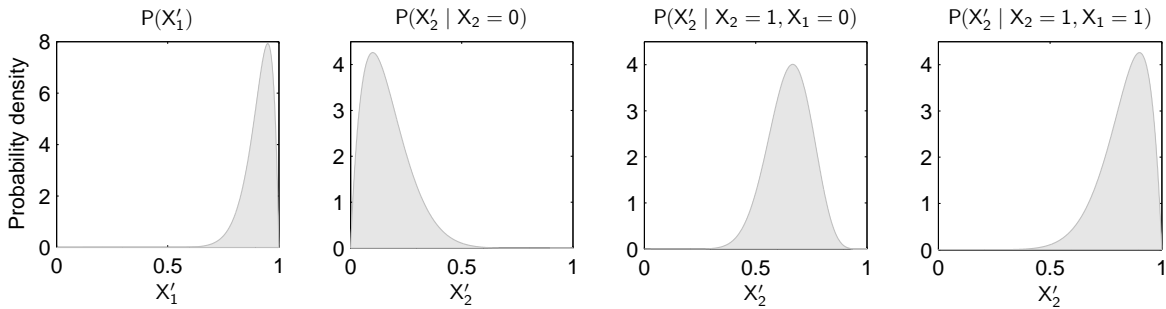}
  \caption{Transition functions for continuous
  variables $X_1'$ and $X_2'$ after taking an action $a_1$ in the
  4-ring topology (Example \ref{ex:CMDP admin}). The
  densities are shown for extreme values of their parent variables
  $X_1$ and $X_2$.}
  \label{fig:CMDP admin transition model}
\end{figure}

\bigskip \noindent {\bf Parameterization of our transition model:}
One-step dynamics of every state variable is described by its
conditional probability distribution $P(X_i' \mid
\Parents(X_i'))$. If $X_i'$ is a continuous variable, its
transition function is represented by a mixture of beta
distributions \shortcite{hauskrecht04linear}:
\begin{align}
  P(X_i' = x \mid \Parents(X_i')) \ & = \
  \sum_j \pi_{ij} \betapdf(x \mid \alpha_j, \beta_j)
  \label{eq:mixture of beta} \\
  \betapdf(x \mid \alpha, \beta) \ & = \
  \frac{\Gamma(\alpha + \beta)}{\Gamma(\alpha) \Gamma(\beta)}
  x^{\alpha - 1} (1 - x)^{\beta - 1}, \nonumber
\end{align}
where $\pi_{ij}$ is the weight assigned to the $j$-th component of
the mixture, and $\alpha_j = \phi_{ij}^\alpha(\Parents(X_i'))$ and
$\beta_j = \phi_{ij}^\beta(\Parents(X_i'))$ are arbitrary positive
functions of the parent set. The mixture of beta distributions
provides a very general class of transition functions and yet
allows closed-form solutions\footnote{The term \emph{closed-form}
refers to a generally accepted set of closed-form operations and
functions extended by the gamma and incomplete beta functions.} to
the expectation terms in HALP (Section \ref{sec:HALP}). If every
$\beta_j = 1$, Equation \ref{eq:mixture of beta} turns into a
polynomial in $X_i'$. Due to the Weierstrass approximation theorem
\shortcite{jeffreys88methods}, such a polynomial is sufficient to
approximate any continuous transition density over $X_i'$ with any
precision. If $X_i'$ is a discrete variable, its transition model
is parameterized by $\abs{\Dom(X_i')}$ nonnegative discriminant
functions $\theta_j = \phi_{ij}^\theta(\Parents(X_i'))$
\shortcite{guestrin04solving}:
\begin{align}
  P(X_i' = j \mid \Parents(X_i')) = \frac{\theta_j}
  {\sum_{j = 1}^{\abs{\Dom(X_i')}} \theta_j}.
  \label{eq:discriminant}
\end{align}
Note that the parameters $\alpha_j$, $\beta_j$, and $\theta_j$
(Equations \ref{eq:mixture of beta} and \ref{eq:discriminant}) are
functions instantiated by value assignments to the variables
$\Parents(X_i') \subseteq \bX \cup \bA$. We keep separate
parameters for every state variable $X_i'$ although our indexing
does not reflect this explicitly. The only restriction on the
functions is that they return valid parameters for all
state-action pairs $(\bx, \ba)$. Hence, we assume that
$\alpha_j(\bx, \ba) \geq 0$, $\beta_j(\bx, \ba) \geq 0$,
$\theta_j(\bx, \ba) \geq 0$, and $\sum_{j = 1}^{\abs{\Dom(X_i')}}
\theta_j(\bx, \ba) > 0$.

\bigskip \noindent {\bf Reward model:} The reward model factors
similarly to the transition model. In particular, the reward
function $R(\bx, \ba) = \sum_j R_j(\bx_j, \ba_j)$ is an additive
function of local reward functions defined on the subsets $\bX_j$
and $\bA_j$. In graphical models, the local functions can be
described compactly by reward nodes $R_j$, which are conditioned
on their parent sets $\Parents(R_j) \! = \! \bX_j \cup \bA_j$. To
allow this representation, we formally extend our DBN to an
influence diagram \shortcite{howard84influence}. Note that the
form of the reward functions $R_j(\bx_j, \ba_j)$ is not
restricted.

\bigskip \noindent {\bf Optimal value function and policy:} The
\emph{optimal policy} $\pistar$ can be defined greedily with
respect to the \emph{optimal value function} $\Vstar$, which is a
fixed point of the Bellman equation:
\begin{align}
  \Vstar(\bx)
  \ & = \ \sup_\ba \left[R(\bx, \ba) +
  \gamma \E{P(\bx' \mid \bx, \ba)}{\Vstar(\bx')}\right]
  \label{eq:hybrid Bellman equation} \\
  \ & = \ \sup_\ba \left[R(\bx, \ba) +
  \gamma \sum_{\bx_D'} \intin{\bx_C'}{P(\bx' \mid \bx, \ba)
  \Vstar(\bx')}\right]. \nonumber
\end{align}
Accordingly, the \emph{hybrid Bellman operator} $\cT^\ast$ is
given by:
\begin{align}
  \cT^\ast V(\bx) = \sup_\ba \left[R(\bx, \ba) +
  \gamma \E{P(\bx' \mid \bx, \ba)}{V(\bx')}\right].
  \label{eq:hybrid Bellman operator}
\end{align}
In the rest of the paper, we denote expectation terms over
discrete and continuous variables in a unified form:
\begin{align}
  \E{P(\bx)}{f(\bx)} =
  \sum_{\bx_D} \int_{\bx_C} P(\bx) f(\bx) \ud \bx_C.
  \label{eq:E}
\end{align}

\begin{example}[\shortciteR{hauskrecht04linear}]
\label{ex:CMDP admin} Continuous-state network administration is a
variation on Example \ref{ex:admin}, where the computer states are
represented by continuous variables on the interval between 0
(being down) and 1 (running). At each time step, the administrator
selects a single action from the set $\cA = \set{a_1, \dots, a_{n
+ 1}}$. The action $a_i$ ($i \leq n$) corresponds to rebooting the
$i$-th computer. The last action $a_{n + 1}$ is dummy. The
transition model captures the propagation of failures in the
network and is encoded locally by beta distributions:
\begin{align*}
  P(X_i' = x \mid \Parents(X_i')) =
  \betapdf(x \mid \alpha, \beta) \quad
  \begin{array}{|l l l}
    & \alpha = 20 & a = i \\
    & \beta = 2 & \\
    & \alpha = 2 + 13 x_i - 5 x_i \E{}{\Parents(X_i')} & a \neq i \\
    & \beta = 10 - 2 x_i - 6 x_i \E{}{\Parents(X_i')} &
  \end{array}
\end{align*}
where the variables $x_i$ and $\E{}{\Parents(X_i')}$ denote the
state of the $i$-th computer and the expected state of its
parents. Note that this transition function is similar to Example
\ref{ex:admin}. For instance, in the 4-ring topology, the modes of
transition densities for continuous variables $X_1'$ and $X_2'$
after taking an action $a_1$ (Figure \ref{fig:CMDP admin
transition model}):
\begin{align*}
  \begin{array}{r r}
    \mode{P}(X_1' \mid a = a_1) = 0.95 &
    \mode{P}(X_2' \mid X_2 = 1, X_1 = 0, a = a_1) \approx 0.67 \\
    \mode{P}(X_2' \mid X_2 = 0, a = a_1) = 0.10 &
    \mode{P}(X_2' \mid X_2 = 1, X_1 = 1, a = a_1) = 0.90
  \end{array}
\end{align*}
equal to the expected values of their discrete counterparts
(Figure \ref{fig:admin}b). The reward function is additive:
\begin{align*}
  R(\bx, a) = 2 x_1^2 + \sum_{j = 2}^n x_j^2
\end{align*}
and establishes our preferences for maintaining the server $X_1$
and workstations $X_2, \dots, X_n$.
\end{example}

\subsection{Solving Hybrid Factored MDPs}
\label{sec:solving HMDPs}

Value iteration, policy iteration, and linear programming are the
most fundamental dynamic programming methods for solving MDPs
\shortcite{puterman94markov,bertsekas96neurodynamic}.
Unfortunately, none of these techniques is suitable for solving
hybrid factored MDPs. First, their complexity is exponential in
the number of state variables if the variables are discrete.
Second, the methods assume a finite support for the optimal value
function or policy, which may not exist if continuous variables
are present. Therefore, any feasible approach to solving arbitrary
HMDPs is likely to be approximate. In the rest of the section, we
review two major classes of methods for approximating value
functions in hybrid domains.

\bigskip \noindent {\bf Grid-based approximation:} Grid-based
methods \shortcite{chow91optimal,rust97using} transform the
initial state space $\bX$ into a set of grid points $G =
\set{\bx^{(1)}, \dots, \bx^{(N)}}$. The points are used to
estimate the optimal value function $V_G^\ast$ on the grid, which
in turn approximates $\Vstar$. The Bellman operator on the grid is
defined as \shortcite{rust97using}:
\begin{align}
  \cT_G^\ast V(\bx^{(i)}) = \max_\ba \left[R(\bx^{(i)}, \ba) +
  \gamma \sum_{j = 1}^N P_G(\bx^{(j)} \mid \bx^{(i)}, \ba)
  V(\bx^{(j)})\right],
  \label{eq:grid Bellman operator}
\end{align}
where $P_G(\bx^{(j)} \mid \bx^{(i)}, \ba) =
\Psi_\ba^{-1}(\bx^{(i)}) P(\bx^{(j)} \mid \bx^{(i)}, \ba)$ is a
transition function, which is normalized by the term
$\Psi_\ba(\bx^{(i)}) = \sum_{j = 1}^N P(\bx^{(j)} \mid \bx^{(i)},
a)$. The operator $\cT_G^\ast$ allows the computation of the value
function $V_G^\ast$ by standard techniques for solving
discrete-state MDPs.

\shortciteA{rust97using} analyzed the convergence of these methods
for random and pseudo-random samples. Clearly, a uniform
discretization of increasing precision guarantees the convergence
of $V_G^\ast$ to $\Vstar$ but causes an exponential blowup in the
state space \shortcite{chow91optimal}. To overcome this concern,
\shortciteA{munos02variable} proposed an adaptive algorithm for
non-uniform discretization based on the Kuhn triangulation.
\shortciteA{ferns05metrics} analyzed metrics for aggregating
states in continuous-state MDPs based on the notion of
bisimulation. \shortciteA{trick93linear} used linear programming
to solve low-dimensional problems with continuous variables. These
continuous variables were discretized manually.

\bigskip \noindent {\bf Parametric value function approximation:}
An alternative approach to solving factored MDPs with
continuous-state components is the approximation of the optimal
value function $\Vstar$ by some parameterized model $V^\lambda$
\shortcite{bertsekas96neurodynamic,vanroy98thesis,gordon99thesis}.
The parameters $\lambda$ are typically optimized iteratively by
applying the backup operator $\cT^\ast$ to a finite set of states.
The least-squares error $\normw{V^\lambda - \cT^\ast
V^\lambda}{2}$ is a commonly minimized error metric (Figure
\ref{fig:L2 VI}). Online updating by gradient methods
\shortcite{bertsekas96neurodynamic,sutton98reinforcement} is
another way of optimizing value functions. The limitation of these
techniques is that their solutions are often unstable and may
diverge \shortcite{bertsekas95counterexample}. On the other hand,
they generate high-quality approximations.

\begin{figure}[t]
  \centering
  \rule{\textwidth}{0.01in}
  \vspace{-0.25in}
  \small{
  \begin{tabbing}
    \hspace{0.2in} \= \hspace{0.2in} \= \hspace{0.2in} \= \hspace{0.2in} \= \kill
    {\bf Inputs:} \\
    \> a hybrid factored MDP $\cM = (\bX, \bA, P, R)$ \\
    \> basis functions $f_0(\bx), f_1(\bx), f_2(\bx), \dots$ \\
    \> initial basis function weights $\bw^{(0)}$ \\
    \> a set of states $G = \set{\bx^{(1)}, \dots, \bx^{(N)}}$ \\
    \\
    {\bf Algorithm:} \\
    \> $t = 0$ \\
    \> while a stopping criterion is not met \\
    \>\> for every state $\bx^{(j)}$ \\
    \>\>\> for every basis function $f_i(\bx)$ \\
    \>\>\>\> $\bX_{ji} = f_i(\bx^{(j)})$ \\
    \>\>\> $\by_j = \max_\ba \left[R(\bx^{(j)}, \ba) +
    \gamma \E{P(\bx' \mid \bx^{(j)}, \ba)}{V^{\bw^{(t)}}(\bx')}\right]$ \\
    \>\> $\bw^{(t + 1)} = (\bX\transpose \bX)^{-1} \bX\transpose \by$ \\
    \>\> $t = t + 1$ \\
    \\
    {\bf Outputs:} \\
    \> basis function weights $\bw^{(t)}$
  \end{tabbing}
  }
  \vspace{-0.15in}
  \rule{\textwidth}{0.01in}
  \caption{Pseudo-code implementation of the least-squares
  value iteration ($\cL_2$ VI) with the linear value function
  approximation (Equation \ref{eq:linear value function}). The
  stopping criterion is often based on the number of steps or the
  $\cL_2$-norm error $\normw{V^{\bw^{(t)}} - \cT^\ast V^{\bw^{(t)}}}{2}$
  measured on the set $G$. Our discussion in Sections
  \ref{sec:HALP expectation terms} and \ref{sec:HALP constraint space}
  provides a recipe for an efficient implementation of the backup
  operation $\cT^\ast V^{\bw^{(t)}}(\bx^{(j)})$.}
  \label{fig:L2 VI}
\end{figure}

\newpage

Parametric approximations often assume fixed value function
models. However, in some cases, it is possible to derive flexible
forms of $V^\lambda$ that combine well with the backup operator
$\cT^\ast$. For instance, \shortciteA{sondik71thesis} showed that
convex piecewise linear functions are sufficient to represent
value functions and their DP backups in partially-observable MDPs
(POMDPs) \shortcite{astrom65optimal,hauskrecht00valuefunction}.
Based on this idea, \shortciteA{feng04dynamic} proposed a method
for solving MDPs with continuous variables. To obtain full DP
backups, the value function approximation is restricted to
\emph{rectangular piecewise linear and convex (RPWLC)} functions.
Further restrictions are placed on the transition and reward
models of MDPs. The advantage of the approach is its adaptivity.
The major disadvantages are restrictions on solved MDPs and the
complexity of RPWLC value functions, which may grow exponentially
in the number of backups. As a result, without further
modifications, this approach is less likely to succeed in solving
high-dimensional and distributed decision problems.

\section{Hybrid Approximate Linear Programming}
\label{sec:HALP}

To overcome the limitations of existing methods for solving HMDPs
(Section \ref{sec:solving HMDPs}), we extend the discrete-state
ALP (Section \ref{sec:ALP}) to hybrid state and action spaces. We
refer to this novel framework as \emph{hybrid approximate linear
programming (HALP)}.

Similarly to the discrete-state ALP, HALP optimizes the linear
value function approximation (Equation \ref{eq:linear value
function}). Therefore, it transforms an initially intractable
problem of computing $\Vstar$ in the hybrid state space $\bX$ into
a lower dimensional space of $\bw$. The HALP formulation is given
by a linear program\footnote{More precisely, the HALP formulation
(\ref{eq:HALP}) is a \emph{linear semi-infinite optimization}
problem with an infinite number of constraints. The number of
basis functions is finite. For brevity, we refer to this
optimization problem as linear programming.}:
\begin{align}
  \textrm{minimize}_\bw & \quad
  \sum_i w_i \alpha_i \label{eq:HALP} \\
  \textrm{subject to:} & \quad
  \sum_i w_i F_i(\bx, \ba) - R(\bx, \ba) \geq 0
  \quad \forall \ \bx \in \bX, \ba \in \bA; \nonumber
\end{align}
where $\bw$ represents the variables in the LP, $\alpha_i$ denotes
\emph{basis function relevance weight}:
\begin{align}
  \alpha_i
  \ & = \ \E{\psi(\bx)}{f_i(\bx)}
  \label{eq:basis function relevance weight} \\
  \ & = \ \sum_{\bx_D} \intin{\bx_C}{\psi(\bx) f_i(\bx)}, \nonumber
\end{align}
$\psi(\bx) \geq 0$ is a \emph{state relevance density function}
that weights the quality of the approximation, and $F_i(\bx, \ba)
= f_i(\bx) - \gamma g_i(\bx, \ba)$ denotes the difference between
the basis function $f_i(\bx)$ and its discounted
\emph{backprojection}:
\begin{align}
  g_i(\bx, \ba)
  \ & = \ \E{P(\bx' \mid \bx, \ba)}{f_i(\bx')}
  \label{eq:hybrid backprojection} \\
  \ & = \ \sum_{\bx_D'} \intin{\bx_C'}{P(\bx' \mid \bx, \ba) f_i(\bx')}.
  \nonumber
\end{align}
Vectors $\bx_D$ ($\bx_D'$) and $\bx_C$ ($\bx_C'$) are the discrete
and continuous components of value assignments $\bx$ ($\bx'$) to
all state variables $\bX$ ($\bX'$). The linear program can be
rewritten compactly:
\begin{align}
  \textrm{minimize}_\bw & \quad
  \E{\psi}{\Vbw} \label{eq:HALP Bellman operator} \\
  \textrm{subject to:} & \quad
  \Vbw - \cT^\ast \Vbw \geq 0 \nonumber
\end{align}
by using the Bellman operator $\cT^\ast$.

The HALP formulation reduces to the discrete-state ALP (Section
\ref{sec:ALP}) if the state and action variables are discrete, and
to the continuous-state ALP \shortcite{hauskrecht04linear} if the
state variables are continuous. The formulation is feasible if the
set of basis functions contains a constant function $f_0(\bx)
\equiv 1$. We assume that such a basis function is present.

In the rest of the paper, we address several concerns related to
the HALP formulation. First, we analyze the quality of this
approximation and relate it to the minimization of the max-norm
error $\maxnorm{\Vstar - \Vbw}$, which is a commonly-used metric
(Section \ref{sec:HALP error bounds}). Second, we present rich
classes of basis functions that lead to closed-form solutions to
the expectation terms in the objective function and constraints
(Equations \ref{eq:basis function relevance weight} and
\ref{eq:hybrid backprojection}). These terms involve sums and
integrals over the complete state space $\bX$ (Section
\ref{sec:HALP expectation terms}), and therefore are hard to
evaluate. Finally, we discuss approximations to the constraint
space in HALP and introduce a framework for solving HALP
formulations in a unified way (Section \ref{sec:HALP constraint
space}). Note that complete satisfaction of this constraint space
may not be possible since every state-action pair $(\bx, \ba)$
induces a constraint.

\subsection{Error Bounds}
\label{sec:HALP error bounds}

The quality of the ALP approximation (Section \ref{sec:ALP}) has
been studied by \shortciteA{defarias03linear}. We follow up on
their work and extend it to structured state and action spaces
with continuous variables. Before we proceed, we demonstrate that
a solution to the HALP formulation (\ref{eq:HALP}) constitutes an
upper bound on the optimal value function $\Vstar$.

\begin{proposition}
\label{prop:HALP lower bound} Let $\bwtilde$ be a solution to the
HALP formulation (\ref{eq:HALP}). Then $\Vbwtilde \geq \Vstar$.
\end{proposition}

\noindent This result allows us to restate the objective
$\E{\psi}{\Vbw}$ in HALP.

\begin{proposition}
\label{prop:HALP L1} Vector $\bwtilde$ is a solution to the HALP
formulation (\ref{eq:HALP}):
\begin{align*}
  \emph{\textrm{minimize}}_\bw & \quad \E{\psi}{\Vbw} \\
  \emph{\textrm{subject to:}} & \quad \Vbw - \cT^\ast \Vbw \geq 0
\end{align*}
if and only if it solves:
\begin{align*}
  \emph{\textrm{minimize}}_\bw & \quad \normw{\Vstar - \Vbw}{1, \psi} \\
  \emph{\textrm{subject to:}} & \quad \Vbw - \cT^\ast \Vbw \geq 0;
\end{align*}
where $\normw{\cdot}{1, \psi}$ is an $\cL_1$-norm weighted by the
state relevance density function $\psi$ and $\cT^\ast$ \mbox{is the}
hybrid Bellman operator.
\end{proposition}

\noindent Based on Proposition \ref{prop:HALP L1}, we conclude
that HALP optimizes the linear value function approximation with
respect to the reweighted $\cL_1$-norm error $\normw{\Vstar -
\Vbw}{1, \psi}$. The following theorem draws a parallel between
minimizing this objective and max-norm error $\maxnorm{\Vstar -
\Vbw}$. More precisely, the theorem says that HALP yields a close
approximation $\Vbwtilde$ to the optimal value function $\Vstar$
if $\Vstar$ is close to the span of basis functions $f_i(\bx)$.

\begin{theorem}
\label{thm:HALP simple bound} Let $\bwtilde$ be an optimal
solution to the HALP formulation (\ref{eq:HALP}). Then the
expected error of the value function $\Vbwtilde$ can be bounded
as:
\begin{align*}
  \normw{\Vstar - \Vbwtilde}{1, \psi} \leq
  \frac{2}{1 - \gamma} \min_\bw \maxnorm{\Vstar - \Vbw},
\end{align*}
where $\normw{\cdot}{1, \psi}$ is an $\cL_1$-norm weighted by the
state relevance density function $\psi$ and $\maxnorm{\cdot}$
\mbox{is a} max-norm.
\end{theorem}

\noindent Unfortunately, Theorem \ref{thm:HALP simple bound}
rarely yields a tight bound on $\normw{\Vstar - \Vbwtilde}{1,
\psi}$. First, it is hard to guarantee a uniformly low max-norm
error $\maxnorm{\Vstar - \Vbw}$ if the dimensionality of a problem
grows but the basis functions $f_i(\bx)$ are local. Second, the
bound ignores the state relevance density function $\psi(\bx)$
although this one impacts the quality of HALP solutions. To
address these concerns, we introduce non-uniform weighting of the
max-norm error in Theorem \ref{thm:HALP bound}.

\begin{theorem}
\label{thm:HALP bound} Let $\bwtilde$ be an optimal solution to
the HALP formulation (\ref{eq:HALP}). Then the expected error of
the value function $\Vbwtilde$ can be bounded as:
\begin{align*}
  \normw{\Vstar - \Vbwtilde}{1, \psi} \leq
  \frac{2 \E{\psi}{\lyapunov}}{1 - \lyapunovfactor}
  \min_\bw \maxnormw{\Vstar - \Vbw}{1 / \lyapunov},
\end{align*}
where $\normw{\cdot}{1, \psi}$ is an $\cL_1$-norm weighted by the
state relevance density $\psi$, $\lyapunov(\bx) \! = \! \sum_i
w_i^\lyapunov f_i(\bx)$ is a \emph{Lyapunov function} such that
the inequality $\lyapunovfactor \lyapunov(\bx) \geq \gamma
\sup_\ba \E{P(\bx' \mid \bx, \ba)}{\lyapunov(\bx')}$ holds,
$\lyapunovfactor \in [0, 1)$ denotes its \emph{contraction
factor}, and $\maxnormw{\cdot}{1 / \lyapunov}$ is a max-norm
reweighted by the reciprocal $1 / \lyapunov$.
\end{theorem}

\noindent Note that Theorem \ref{thm:HALP simple bound} is a
special form of Theorem \ref{thm:HALP bound} when $\lyapunov(\bx)
\equiv 1$ and $\lyapunovfactor = \gamma$. Therefore, the Lyapunov
function $\lyapunov(\bx)$ permits at least as good bounds as
Theorem \ref{thm:HALP simple bound}. To make these bounds tight,
the function $\lyapunov(\bx)$ should return large values in the
regions of the state space, which are unimportant for modeling. In
turn, the reciprocal $1 / \lyapunov(\bx)$ is close to zero in
these undesirable regions, which makes their impact on the
max-norm error $\maxnormw{\Vstar - \Vbw}{1 / \lyapunov}$ less
likely. Since the state relevance density function $\psi(\bx)$
reflects the importance of states, the term $\E{\psi}{\lyapunov}$
should remain small. These two factors contribute to tighter
bounds than those by Theorem \ref{thm:HALP simple bound}.

Since the Lyapunov function $\lyapunov(\bx) = \sum_i w_i^\lyapunov
f_i(\bx)$ lies in the span of basis functions $f_i(\bx)$, Theorem
\ref{thm:HALP bound} provides a recipe for achieving high-quality
approximations. Intuitively, a good set of basis functions always
involves two types of functions. The first type guarantees small
errors $\abs{\Vstar(\bx) - \Vbw(\bx)}$ in the important regions of
the state space, where the state relevance density $\psi(\bx)$ is
high. The second type returns high values where the state
relevance density $\psi(\bx)$ is low, and vice versa. The latter
functions allow the satisfaction of the constraint space $\Vbw
\geq \cT^\ast \Vbw$ in the unimportant regions of the state space
without impacting the optimized objective function $\normw{\Vstar
- \Vbw}{1, \psi}$. Note that a trivial value function $\Vbw(\bx) =
(1 - \gamma)^{-1} R_\mathrm{max}$ satisfies all constraints in any
HALP but unlikely leads to good policies. For a comprehensive
discussion on selecting appropriate $\psi(\bx)$ and
$\lyapunov(\bx)$, refer to the case studies of
\shortciteA{defarias03linear}.

Our discussion is concluded by clarifying the notion of the state
relevance density $\psi(\bx)$. As demonstrated by Theorem
\ref{thm:HALP policy bound}, its choice is closely related to the
quality of a greedy policy for the value function $\Vbwtilde$
\shortcite{defarias03linear}.

\begin{theorem}
\label{thm:HALP policy bound} Let $\bwtilde$ be an optimal
solution to the HALP formulation (\ref{eq:HALP}). Then the
expected error of a greedy policy:
\begin{align*}
  u(\bx) = \arg\sup_\ba \left[R(\bx, \ba) +
  \gamma \E{P(\bx' \mid \bx, \ba)}{\Vbwtilde(\bx')}\right]
\end{align*}
can be bounded as:
\begin{align*}
  \normw{\Vstar - V^u}{1, \nu} \leq
  \frac{1}{1 - \gamma}
  \normw{\Vstar - \Vbwtilde}{1, \mu_{u, \nu}},
\end{align*}
where $\normw{\cdot}{1, \nu}$ and $\normw{\cdot}{1, \mu_{u, \nu}}$
are weighted $\cL_1$-norms, $V^u$ is a value function for the
greedy policy $u$, and $\mu_{u, \nu}$ is the expected frequency of
state visits generated by following the policy $u$ given the
initial state distribution $\nu$.
\end{theorem}

\noindent Based on Theorem \ref{thm:HALP policy bound}, we may
conclude that the expected error of greedy policies for HALP
approximations is bounded when $\psi \! = \! \mu_{u, \nu}$. Note
that the distribution $\mu_{u, \nu}$ is unknown when optimizing
$\Vbwtilde$ because it is a function of the optimized quantity
itself. To break this cycle, \shortciteA{defarias03linear} suggested
an iterative procedure that solves several LPs and adapts $\mu_{u,
\nu}$ accordingly. In addition, real-world control problems exhibit
a lot of structure, which permits the guessing of $\mu_{u, \nu}$.

Finally, it is important to realize that although our bounds
(Theorems \ref{thm:HALP bound} and \ref{thm:HALP policy bound})
build a foundation for better HALP approximations, they can be
rarely used in practice \mbox{because the} optimal value function
$\Vstar$ is generally unknown. After all, if it was known, there
is no need to approximate it. Moreover, note that the optimization
of $\maxnormw{\Vstar - \Vbw}{1 / \lyapunov}$ (Theorem
\ref{thm:HALP bound}) is a hard problem and there are no methods
that would minimize this error directly
\shortcite{patrascu02greedy}. Despite these facts, both bounds
provide a loose guidance for empirical choices of basis functions.
In Section \ref{sec:experiments}, we use this intuition and
propose basis functions that should closely approximate unknown
optimal value functions $\Vstar$.

\subsection{Expectation Terms}
\label{sec:HALP expectation terms}

Since our basis functions are often restricted to small subsets of
state variables, expectation terms (Equations \ref{eq:basis
function relevance weight} and \ref{eq:hybrid backprojection}) in
the HALP formulation (\ref{eq:HALP}) should be efficiently
computable. To unify the analysis of these expectation terms,
$\E{\psi(\bx)}{f_i(\bx)}$ and $\E{P(\bx' \mid \bx,
\ba)}{f_i(\bx')}$, we show that their evaluation constitutes the
same computational problem $\E{P(\bx)}{f_i(\bx)}$, where $P(\bx)$
denotes some factored distribution.

Before we discuss expectation terms in the constraints, note that
the transition function $P(\bx' \mid \bx, \ba)$ is factored and its
parameterization is determined by the state-action pair $(\bx,
\ba)$. We keep the pair $(\bx, \ba)$ fixed in the rest of the
section, which corresponds to choosing a single constraint $(\bx,
\ba)$. Based on this selection, we rewrite the expectation terms
$\E{P(\bx' \mid \bx, \ba)}{f_i(\bx')}$ in a simpler notation
$\E{P(\bx')}{f_i(\bx')}$, where $P(\bx') \! = \! P(\bx' \! \mid \!
\bx, \ba)$ denotes a factored distribution with fixed parameters.

We also assume that the state relevance density function
$\psi(\bx)$ factors along $\bX$ as:
\begin{align}
  \psi(\bx) = \prod_{i = 1}^n \psi_i(x_i),
  \label{eq:state relevance density function}
\end{align}
where $\psi_i(x_i)$ is a distribution over the random state variable
$X_i$. Based on this assumption, we can rewrite the expectation
terms $\E{\psi(\bx)}{f_i(\bx)}$ in the objective function in a new
notation $\E{P(\bx)}{f_i(\bx)}$, where $P(\bx) = \psi(\bx)$ denotes
a factored distribution. In line with our discussion in the last two
paragraphs, efficient solutions to the expectation terms in HALP are
obtained by solving the generalized term $\E{P(\bx)}{f_i(\bx)}$
efficiently. We address this problem in the rest of the section.

Before computing the expectation term $\E{P(\bx)}{f_i(\bx)}$ over
the complete state space $\bX$, we recall that the basis function
$f_i(\bx)$ is defined on a subset of state variables $\bX_i$.
Therefore, we may conclude that $\E{P(\bx)}{f_i(\bx)} =
\E{P(\bx_i)}{f_i(\bx_i)}$, where $P(\bx_i)$ denotes a factored
distribution on a lower dimensional space $\bX_i$. If no further
assumptions are made, the local expectation term
$\E{P(\bx_i)}{f_i(\bx_i)}$ may be still hard to compute. Although
it can be estimated by a variety of numerical methods, for
instance Monte Carlo \shortcite{andrieu03introduction}, these
techniques are imprecise if the sample size is small, and quite
computationally expensive if a high precision is needed.
Consequently, we try to avoid such an approximation step. Instead,
we introduce an appropriate form of basis functions that leads to
closed-form solutions to the expectation term
$\E{P(\bx_i)}{f_i(\bx_i)}$.

In particular, let us assume that every basis function
$f_i(\bx_i)$ factors as:
\begin{align}
  f_i(\bx_i) = f_{i_D}(\bx_{i_D}) f_{i_C}(\bx_{i_C})
  \label{eq:hybrid basis function}
\end{align}
along its discrete and continuous components $f_{i_D}(\bx_{i_D})$
and $f_{i_C}(\bx_{i_C})$, where the continuous component further
decouples as a product:
\begin{align}
  f_{i_C}(\bx_{i_C}) = \prod_{X_j \in \bX_{i_C}} f_{ij}(x_j)
  \label{eq:continuous basis function}
\end{align}
of univariate basis function factors $f_{ij}(x_j)$. Note that the
basis functions remain multivariate despite the two independence
assumptions. We make these presumptions for computational purposes
and they are relaxed later in the section.

Based on Equation \ref{eq:hybrid basis function}, we conclude that
the expectation term:
\begin{align}
  \E{P(\bx_i)}{f_i(\bx_i)}
  \ & = \ \E{P(\bx_i)}{f_{i_D}(\bx_{i_D}) f_{i_C}(\bx_{i_C})}
  \nonumber \\
  \ & = \ \E{P(\bx_{i_D})}{f_{i_D}(\bx_{i_D})}
  \E{P(\bx_{i_C})}{f_{i_C}(\bx_{i_C})}
  \label{eq:hybrid expectation terms}
\end{align}
decomposes along the discrete and continuous variables $\bX_{i_D}$
and $\bX_{i_C}$, where $\bx_i = (\bx_{i_D}, \bx_{i_C})$ and
$P(\bx_i) = P(\bx_{i_D}) P(\bx_{i_C})$. The evaluation of the
discrete part $\E{P(\bx_{i_D})}{f_{i_D}(\bx_{i_D})}$ requires
aggregation in the subspace $\bX_{i_D}$:
\begin{align}
  \E{P(\bx_{i_D})}{f_{i_D}(\bx_{i_D})} =
  \sum_{\bx_{i_D}} P(\bx_{i_D}) f_{i_D}(\bx_{i_D}),
  \label{eq:discrete expectation terms}
\end{align}
which can be carried out efficiently in $O(\prod_{X_j \in
\bX_{i_D}} \abs{\Dom(X_j)})$ time (Section \ref{sec:ALP}).
Following Equation \ref{eq:continuous basis function}, the
continuous term $\E{P(\bx_{i_C})}{f_{i_C}(\bx_{i_C})}$ decouples
as a product:
\begin{align}
  \E{P(\bx_{i_C})}{f_{i_C}(\bx_{i_C})}
  \ & = \ \E{P(\bx_{i_C})}{\prod_{X_j \in \bX_{i_C}} f_{ij}(x_j)}
  \nonumber \\
  \ & = \ \prod_{X_j \in \bX_{i_C}} \E{P(x_j)}{f_{ij}(x_j)},
  \label{eq:continuous expectation terms}
\end{align}
where $\E{P(x_j)}{f_{ij}(x_j)}$ represents the expectation terms
over individual random variables $X_j$. Consequently, an efficient
solution to the local expectation term $\E{P(\bx_i)}{f_i(\bx_i)}$
is guaranteed by efficient solutions to its univariate components
$\E{P(x_j)}{f_{ij}(x_j)}$.

In this paper, we consider three univariate basis function factors
$f_{ij}(x_j)$: piecewise linear functions, polynomials, and beta
distributions. These factors support a very general class of basis
functions and yet allow closed-form solutions to the expectation
terms $\E{P(x_j)}{f_{ij}(x_j)}$. These solutions are provided in
the following propositions and demonstrated in Example
\ref{ex:CMDP admin expectation terms}.

\begin{figure}[t]
  \centering
  \includegraphics[width=5.6in, bb=0.8in 4.65in 7.8in 6.65in]{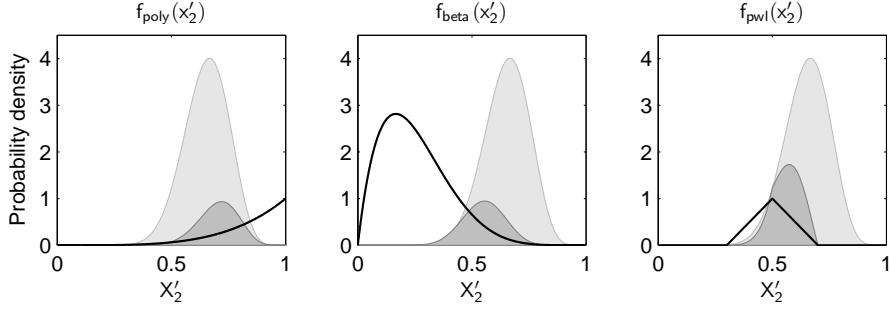}
  \caption{Expectation of three basis functions $f(x_2')$
  (Example \ref{ex:CMDP admin expectation terms}) with respect to
  the transition function $P(X_2' \mid X_2 = 1, X_1 = 0, a = a_1)$
  from Figure \ref{fig:CMDP admin transition model}. Every basis
  function $f(x_2')$ is depicted by a thick black line. The
  transition function is shown in a light gray color. Darker gray
  lines represent the values of the product $P(x_2' \! \mid \! \bx, a_1)
  f(x_2')$. The area below corresponds to the expectation terms
  $\E{P(x_2' \mid \bx, a_1)}{f(x_2')}$.}
  \label{fig:CMDP admin expectation terms}
\end{figure}

\begin{proposition}[Polynomial basis functions]
\label{prop:polynomial basis function} Let:
\begin{align*}
  P(x) = \betapdf(x \mid \alpha, \beta)
\end{align*}
be a beta distribution over $X$ and:
\begin{align*}
  f(x) = x^n (1 - x)^m
\end{align*}
be a polynomial in $x$ and $(1 - x)$. Then $\E{P(x)}{f(x)}$ has a
closed-form solution:
\begin{align*}
  \E{P(x)}{f(x)} =
  \frac{\Gamma(\alpha + \beta)}
  {\Gamma(\alpha) \Gamma(\beta)}
  \frac{\Gamma(\alpha + n) \Gamma(\beta + m)}
  {\Gamma(\alpha + \beta + n + m)}.
\end{align*}
\end{proposition}

\begin{corollary}[Beta basis functions]
\label{cor:beta basis function} Let:
\begin{align*}
  P(x) \ & = \ \betapdf(x \mid \alpha, \beta) \\
  f(x) \ & = \ \betapdf(x \mid \alpha_f, \beta_f)
\end{align*}
be beta distributions over $X$. Then $\E{P(x)}{f(x)}$ has a
closed-form solution:
\begin{align*}
  \E{P(x)}{f(x)} =
  \frac{\Gamma(\alpha + \beta)}
  {\Gamma(\alpha) \Gamma(\beta)}
  \frac{\Gamma(\alpha_f + \beta_f)}
  {\Gamma(\alpha_f) \Gamma(\beta_f)}
  \frac{\Gamma(\alpha + \alpha_f - 1) \Gamma(\beta + \beta_f - 1)}
  {\Gamma(\alpha + \alpha_f + \beta + \beta_f - 2)}.
\end{align*}
\end{corollary}
{\bf Proof:} A direct consequence of Proposition
\ref{prop:polynomial basis function}. Since integration is a
distributive operation, our claim straightforwardly generalizes to
the mixture of beta distributions $P(x)$. \qed

\begin{proposition}[Piecewise linear basis functions]
\label{prop:piecewise linear basis function} Let:
\begin{align*}
  P(x) = \betapdf(x \mid \alpha, \beta)
\end{align*}
be a beta distribution over $X$ and:
\begin{align*}
  f(x) = \sum_i \I{[l_i, r_i]}{x} (a_i x + b_i)
\end{align*}
be a piecewise linear (PWL) function in $x$, where $\I{[l_i,
r_i]}{x}$ represents the indicator function of the interval $[l_i,
r_i]$. Then $\E{P(x)}{f(x)}$ has a closed-form solution:
\begin{align*}
  \E{P(x)}{f(x)} =
  \sum_i \left[a_i \frac{\alpha}{\alpha + \beta}
  (F^+(r_i) - F^+(l_i)) + b_i (F(r_i) - F(l_i))\right],
\end{align*}
where $F(u) = \betacdf(u \mid \alpha, \beta)$ and $F^+(u) =
\betacdf(u \mid \alpha + 1, \beta)$ denote the cumulative density
functions of beta distributions.
\end{proposition}

\begin{example}
\label{ex:CMDP admin expectation terms} Efficient closed-form
solutions to the expectation terms in HALP are illustrated on the
4-ring network administration problem (Example \ref{ex:CMDP admin})
with three hypothetical univariate basis functions:
\begin{align*}
  f_\mathrm{poly}(x_2') \ & = \ x_2'^4 \\
  f_\mathrm{beta}(x_2') \ & = \ \betapdf(x_2' \mid 2, 6) \\
  f_\mathrm{pwl}(x_2') \ & = \ \I{[0.3, 0.5]}{x_2'} (5 x_2' - 1.5) +
  \I{[0.5, 0.7]}{x_2'} (-5 x_2' + 3.5)
\end{align*}
Suppose that our goal is to evaluate expectation terms in a single
constraint that corresponds to the network state $\bx \! = \! (0,
1, 0, 0)$ and the administrator rebooting the server. Based on
these assumptions, the expectation terms in the constraint $(\bx,
a_1)$ simplify as:
\begin{align*}
  \E{P(\bx' \mid \bx, a_1)}{f(x_2')} =
  \E{P(x_2' \mid \bx, a_1)}{f(x_2')},
\end{align*}
where the transition function $P(x_2' \mid \bx, a_1)$ is given by:
\begin{align*}
  P(x_2' \mid \bx, a_1)
  \ & = \ P(X_2' = x_2' \mid X_2 = 1, X_1 = 0, a = a_1) \\
  \ & = \ \betapdf(x_2' \mid 15, 8).
\end{align*}
Closed-form solutions to the simplified expectation terms
$\E{P(x_2' \mid \bx, a_1)}{f(x_2')}$ are computed as:
\begin{align*}
  \E{P(x_2' \mid \bx, a_1)}{f_\mathrm{poly}(x_2')}
  \ = & \ \ \int_{x_2'} \betapdf(x_2' \mid 15, 8)
  x_2'^4 \ud x_2' \\
  {\scriptstyle \emph{(Proposition
  \ref{prop:polynomial basis function})}}
  \ = & \ \ \frac{\Gamma(15 + 8)}{\Gamma(15) \Gamma(8)}
  \frac{\Gamma(15 + 4) \Gamma(8)}{\Gamma(15 + 8 + 4)} \\
  \approx & \ \ 0.20 \\
  \E{P(x_2' \mid \bx, a_1)}{f_\mathrm{beta}(x_2')}
  \ = & \ \ \int_{x_2'} \betapdf(x_2' \mid 15, 8)
  \betapdf(x_2' \mid 2, 6) \ud x_2' \\
  {\scriptstyle \emph{(Corollary
  \ref{cor:beta basis function})}}
  \ = & \ \ \frac{\Gamma(15 + 8)}{\Gamma(15) \Gamma(8)}
  \frac{\Gamma(2 + 6)}{\Gamma(2) \Gamma(6)}
  \frac{\Gamma(15 + 2 - 1) \Gamma(8 + 6 - 1)}
  {\Gamma(15 + 2 + 8 + 6 - 2)} \\
  \approx & \ \ 0.22 \\
  \E{P(x_2' \mid \bx, a_1)}{f_\mathrm{pwl}(x_2')}
  \ = & \ \ \int_{x_2'} \betapdf(x_2' \mid 15, 8)
  \I{[0.3, 0.5]}{x_2'} (5 x_2' - 1.5) \ud x_2' + \\
  & \ \ \int_{x_2'} \betapdf(x_2' \mid 15, 8)
  \I{[0.5, 0.7]}{x_2'} (-5 x_2' + 3.5) \ud x_2' \\
  {\scriptstyle \emph{(Proposition
  \ref{prop:piecewise linear basis function})}}
  \ = & \ \ 5 \frac{15}{15 + 8}
  (F^+(0.5) - F^+(0.3)) - 1.5 (F(0.5) - F(0.3)) - \\
  & \ \ 5 \frac{15}{15 + 8}
  (F^+(0.7) - F^+(0.5)) + 3.5 (F(0.7) - F(0.5)) \\
  \approx & \ \ 0.30
\end{align*}
where $F(u) = \betacdf(u \mid 15, 8)$ and $F^+(u) = \betacdf(u
\mid 15 + 1, 8)$ denote the cumulative density functions of beta
distributions. A graphical interpretation of these computations is
presented in Figure \ref{fig:CMDP admin expectation terms}. Brief
inspection verifies that the term $\E{P(x_2' \mid \bx,
a_1)}{f_\mathrm{pwl}(x_2')}$ is indeed the largest one.
\end{example}

\noindent Up to this point, we obtained efficient closed-form
solutions for factored basis functions and state relevance
densities. Unfortunately, the factorization assumptions in
Equations \ref{eq:state relevance density function},
\ref{eq:hybrid basis function}, and \ref{eq:continuous basis
function} are rarely justified in practice. In the rest of the
section, we show how to relax them. In Section \ref{sec:HALP
constraint space}, we apply our current results and propose
several methods that approximately satisfy the constraint space in
HALP.

\subsubsection{Factored State Relevance Density Functions}
\label{sec:HALP factored state relevance density functions}

Note that the state relevance density function $\psi(\bx)$ is very
unlikely to be completely factored (Section \ref{sec:HALP error
bounds}). Therefore, the independence assumption in Equation
\ref{eq:state relevance density function} is extremely limiting.
To relax this assumption, we approximate $\psi(\bx)$ by a linear
combination $\psi^\omega(\bx) \! = \! \sum_\ell \omega_\ell
\psi_\ell(\bx)$ of factored state relevance densities
$\psi_\ell(\bx) = \prod_{i = 1}^n \psi_{\ell i}(x_i)$. As a
result, the expectation terms in the objective function decompose
as:
\begin{align}
  \E{\psi^\omega(\bx)}{f_i(\bx)}
  \ & = \ \E{\sum_\ell \omega_\ell \psi_\ell(\bx)}{f_i(\bx)}
  \nonumber \\
  \ & = \ \sum_\ell \omega_\ell \E{\psi_\ell(\bx)}{f_i(\bx)},
  \label{eq:mixture of state relevance density functions}
\end{align}
where the factored terms $\E{\psi_\ell(\bx)}{f_i(\bx)}$ can be
evaluated efficiently (Equation \ref{eq:hybrid expectation
terms}). Moreover, if we assume the factored densities
$\psi_\ell(\bx)$ are polynomials, their linear combination
$\psi^\omega(\bx)$ is a polynomial. Due to the Weierstrass
approximation theorem \shortcite{jeffreys88methods}, this
polynomial is sufficient to approximate any state relevance
density $\psi(\bx)$ with any precision. It follows that the linear
combinations permit state relevance densities that reflect
arbitrary dependencies among the state variables $\bX$.

\subsubsection{Factored Basis Functions}
\label{sec:HALP factored basis functions}

In line with the previous discussion, note that the linear value
function $\Vbw(\bx) = \sum_i w_i f_i(\bx)$ with factored basis
functions (Equations \ref{eq:hybrid basis function} and
\ref{eq:continuous basis function}) is sufficient to approximate
the optimal value function $\Vstar$ within any max-norm error
$\maxnorm{\Vstar - \Vbw}$. Based on Theorem \ref{thm:HALP simple
bound}, we know that the same set of basis functions guarantees a
bound on the $\cL_1$-norm error $\normw{\Vstar - \Vbwtilde}{1,
\psi}$. Therefore, despite our independence assumptions (Equations
\ref{eq:hybrid basis function} and \ref{eq:continuous basis
function}), we have a potential to obtain an arbitrarily close
HALP approximation $\Vbwtilde$ to $\Vstar$.

\section{Constraint Space Approximations}
\label{sec:HALP constraint space}

An optimal solution $\bwtilde$ to the HALP formulation
(\ref{eq:HALP}) is determined by a finite set of \emph{active
constraints} at a vertex of the feasible region. Unfortunately,
identification of this active set is a hard computational problem.
In particular, it requires searching through an exponential number
of constraints, if the state and action variables are discrete,
and an infinite number of constraints, if any of the variables are
continuous. As a result, it is in general infeasible to find the
optimal solution $\bwtilde$ to the HALP formulation. Hence, we
resort to approximations to the constraint space in HALP whose
optimal solution $\bwhat$ is close to $\bwtilde$. This notion of
an approximation is formalized as follows.

\begin{figure}[t]
  \centering
  \includegraphics[width=5.6in, bb=0.8in 4.65in 7.8in 6.65in]{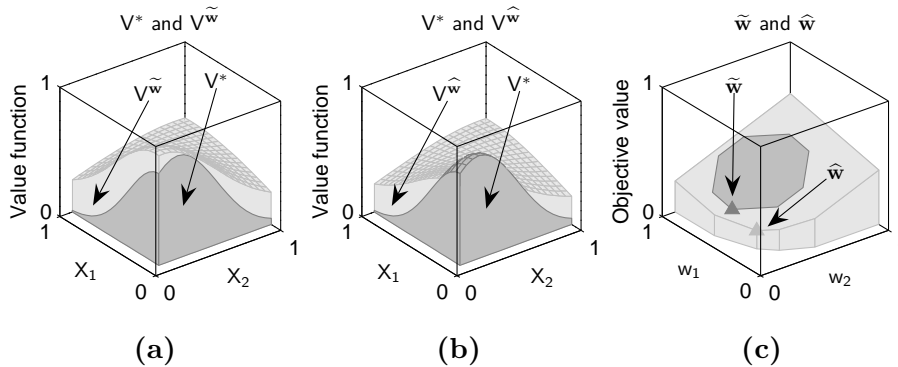}
  \vspace{0.2in} \\
  $\! \! \! \! \!$\textbf{(a)}
  $\qquad \qquad \qquad \qquad \ \, \,$\textbf{(b)}
  $\qquad \qquad \qquad \qquad \ \, \,$\textbf{(c)}
  \caption{\textbf{a.} Graphical relation between the value
  function $\Vstar$ and its HALP approximation $\Vbwtilde$.
  The function $\Vbwtilde$ is guaranteed to be an upper bound on
  $\Vstar$. \textbf{b.} The relaxed HALP approximation $\Vbwhat$
  may not lead to an upper bound. \textbf{c.} Graphical relation
  between the optimal and relaxed solutions $\bwtilde$ and
  $\bwhat$. The feasible regions of the complete and relaxed HALP
  formulations are shown in dark and light gray colors. The value
  function approximations $\Vbwtilde$ and $\Vbwhat$ are typically
  nonlinear in the state space $\bX$ but always linear in the
  space of parameters $\bw$.}
  \label{fig:HALP}
\end{figure}

\begin{definition}
\label{def:relaxed HALP} The HALP formulation is \emph{relaxed}:
\begin{align}
  \emph{\textrm{minimize}}_\bw & \quad
  \sum_i w_i \alpha_i \label{eq:relaxed HALP} \\
  \emph{\textrm{subject to:}} & \quad
  \sum_i w_i F_i(\bx, \ba) - R(\bx, \ba) \geq 0
  \quad (\bx, \ba) \in \cC; \nonumber
\end{align}
if only a subset $\cC$ of its constraints is satisfied.
\end{definition}

\noindent The HALP formulation (\ref{eq:HALP}) can be solved
approximately by solving its relaxed formulations (\ref{eq:relaxed
HALP}). Several methods for building and solving these approximate
LPs have been proposed: Monte Carlo sampling of constraints,
\shortcite{hauskrecht04linear}, $\eps$-grid discretization of the
constraint space \shortcite{guestrin04solving}, and an adaptive
search for a violated constraint \shortcite{kveton05mcmc}. In the
remainder of this section, we introduce these methods. From now
on, we denote optimal solutions to the complete and relaxed HALP
formulations by the symbols $\bwtilde$ and $\bwhat$, respectively.

Before we proceed, note that while $\Vbwtilde$ is an upper bound
on the optimal value function $\Vstar$ (Figure \ref{fig:HALP}a),
the relaxed value function $\Vbwhat$ does not have to be (Figure
\ref{fig:HALP}b). The reason is that the relaxed HALP formulation
does not guarantee that the constraint $\Vbwhat \geq \cT^\ast
\Vbwhat$ is satisfied for all states $\bx$. As a result, we cannot
simply use Proposition \ref{prop:HALP lower bound} to prove
$\Vbwhat \geq \Vstar$. Furthermore, note that the inequality
$\E{\psi}{\Vbwhat} \leq \E{\psi}{\Vbwtilde}$ always holds because
the optimal solution $\bwtilde$ is feasible in the relaxed HALP
(Figure \ref{fig:HALP}c). These observations become helpful for
understanding the rest of the section.

\subsection{MC-HALP}
\label{sec:MC-HALP}

In the simplest case, the constraint space in HALP can be
approximated by its Monte Carlo (MC) sample. In such a relaxation,
the set of constraints $\cC$ is selected with respect to some
proposal distribution $\varphi$ over state-action pairs $(\bx,
\ba)$. Since the set $\cC$ is finite, it establishes a relaxed
formulation (\ref{eq:relaxed HALP}), which can be solved by any LP
solver. An algorithm that builds and satisfies relaxed MC-HALP
formulations is outlined in Figure \ref{fig:MC-HALP solver}.

Constraint sampling is easily applied in continuous domains and
its space complexity is proportional to the number of state and
action components. \shortciteA{hauskrecht04linear} used it to
solve continuous-state factored MDPs and further refined it by
heuristics \shortcite{kveton04heuristic}. In discrete-state
domains, the quality of the sampled approximations was analyzed by
\shortciteA{defarias04constraint}. Their result is summarized by
Theorem \ref{thm:relaxed ALP bound}.

\begin{theorem}[\shortciteR{defarias04constraint}]
\label{thm:relaxed ALP bound} Let $\bwtilde$ be a solution to the
ALP formulation (\ref{eq:ALP}) and $\bwhat$ be a solution to its
relaxed formulation whose constraints are sampled with respect to
a proposal distribution $\varphi$ over state-action pairs $(\bx,
\ba)$. Then there exist a distribution $\varphi$ and sample size:
\begin{align*}
  N \geq O\left(\frac{A \theta}{(1 - \gamma) \epsilon}
  \left(K \ln \frac{A \theta}{(1 - \gamma) \epsilon} +
  \ln \frac{1}{\delta}\right)\right)
\end{align*}
such that with probability at least $1 - \delta$:
\begin{align*}
  \normw{\Vstar - \Vbwhat}{1, \psi} \leq
  \normw{\Vstar - \Vbwtilde}{1, \psi} +
  \epsilon \normw{\Vstar}{1, \psi},
\end{align*}
where $\normw{\cdot}{1, \psi}$ is an $\cL_1$-norm weighted by the
state relevance weights $\psi$, $\theta$ is a problem-specific
constant, $A$ and $K$ denote the numbers of actions and basis
functions, and $\epsilon$ and $\delta$ are scalars from the
interval $(0, 1)$.
\end{theorem}

\noindent Unfortunately, proposing a sampling distribution
$\varphi$ that guarantees this polynomial bound on the sample size
is as hard as knowing the optimal policy $\pistar$
\shortcite{defarias04constraint}. This conclusion is parallel to
those in importance sampling. Note that uniform Monte Carlo
sampling can guarantee a low probability of constraints being
violated but it is not sufficient to bound the magnitude of their
violation \shortcite{defarias04constraint}.

\subsection{$\eps$-HALP}
\label{sec:e-HALP}

\begin{figure}[t]
  \centering
  \rule{\textwidth}{0.01in}
  \vspace{-0.25in}
  \small{
  \begin{tabbing}
    \hspace{0.2in} \= \hspace{0.2in} \= \hspace{0.2in} \= \hspace{0.2in} \= \kill
    {\bf Inputs:} \\
    \> a hybrid factored MDP $\cM = (\bX, \bA, P, R)$ \\
    \> basis functions $f_0(\bx), f_1(\bx), f_2(\bx), \dots$ \\
    \> a proposal distribution $\varphi$ \\
    \\
    {\bf Algorithm:} \\
    \> initialize a relaxed HALP formulation with an empty set of constraints \\
    \> $t = 0$ \\
    \> while a stopping criterion is not met \\
    \>\> sample $(\bx, \ba) \sim \varphi$ \\
    \>\> add the constraint $(\bx, \ba)$ to the relaxed HALP \\
    \>\> $t = t + 1$ \\
    \> solve the relaxed MC-HALP formulation \\
    \\
    {\bf Outputs:} \\
    \> basis function weights $\bw$
  \end{tabbing}
  }
  \vspace{-0.15in}
  \rule{\textwidth}{0.01in}
  \caption{Pseudo-code implementation of the MC-HALP solver.}
  \label{fig:MC-HALP solver}
\end{figure}

\begin{figure}[t]
  \centering
  \rule{\textwidth}{0.01in}
  \vspace{-0.25in}
  \small{
  \begin{tabbing}
    \hspace{0.2in} \= \hspace{0.2in} \= \hspace{0.2in} \= \hspace{0.2in} \= \kill
    {\bf Inputs:} \\
    \> a hybrid factored MDP $\cM = (\bX, \bA, P, R)$ \\
    \> basis functions $f_0(\bx), f_1(\bx), f_2(\bx), \dots$ \\
    \> grid resolution $\eps$ \\
    \\
    {\bf Algorithm:} \\
    \> discretize continuous variables $\bX_C$
    and $\bA_C$ into $\ceils{1 / \eps + 1}$ equally-spaced values \\
    \> identify subsets $\bX_i$ and $\bA_i$
    ($\bX_j$ and $\bA_j$) corresponding to the domains of
    $F_i(\bx, \ba)$ ($R_j(\bx, \ba)$) \\
    \> evaluate $F_i(\bx_i, \ba_i)$ ($R_j(\bx_j, \ba_j)$) for
    all configurations $\bx_i$ and $\ba_i$ ($\bx_j$ and $\ba_j$)
    on the $\eps$-grid \\
    \> calculate basis function relevance weights $\alpha_i$ \\
    \> solve the relaxed $\eps$-HALP formulation (Section \ref{sec:ALP}) \\
    \\
    {\bf Outputs:} \\
    \> basis function weights $\bw$
  \end{tabbing}
  }
  \vspace{-0.15in}
  \rule{\textwidth}{0.01in}
  \caption{Pseudo-code implementation of the $\eps$-HALP solver.}
  \label{fig:e-HALP solver}
\end{figure}

Another way of approximating the constraint space in HALP is by
discretizing its continuous variables $\bX_C$ and $\bA_C$ on a
uniform $\eps$-grid. The new discretized constraint space
preserves its original factored structure but spans discrete
variables only. Therefore, it can be compactly satisfied by the
methods for discrete-state ALP (Section \ref{sec:ALP}). An
algorithm that builds and satisfies relaxed $\eps$-HALP
formulations is outlined in Figure \ref{fig:e-HALP solver}. Note
that the new constraint space involves exponentially many
constraints $O(\ceils{1 / \eps + 1}^{\abs{\bX_C} + \abs{\bA_C}})$
in the number of state and action variables $\bX_C$ and $\bA_C$.

\subsubsection{Error Bounds}
\label{sec:e-HALP error bounds}

Recall that the $\eps$-HALP formulation approximates the
constraint space in HALP by a finite set of equally-spaced grid
points. In this section, we study the quality of this
approximation and bound it in terms violating constraints in the
complete HALP. More precisely, we prove that if a relaxed HALP
solution $\bwhat$ violates the constraints in the complete HALP by
a small amount, the quality of the approximation $\Vbwhat$ is
close to $\Vbwtilde$. In the next section, we extend this result
and relate $\Vbwhat$ to the grid resolution $\eps$. Before we
proceed, we quantify our notion of constraint violation.

\begin{definition}
\label{def:delta-infeasible} Let $\bwhat$ be an optimal solution
to a relaxed HALP formulation (\ref{eq:relaxed HALP}). The vector
$\bwhat$ is \emph{$\delta$-infeasible} if:
\begin{align}
  \Vbwhat - \cT^\ast \Vbwhat \geq - \delta,
  \label{eq:delta-infeasible}
\end{align}
where $\cT^\ast$ is the hybrid Bellman operator.
\end{definition}

\noindent Intuitively, the lower the $\delta$-infeasibility of a
relaxed HALP solution $\bwhat$, the closer the quality of the
approximation $\Vbwhat$ to $\Vbwtilde$. Proposition \ref{prop:HALP
to relaxed HALP bound} states this intuition formally. In
particular, it says that the relaxed HALP formulation leads to a
close approximation $\Vbwhat$ to the optimal value function
$\Vstar$ if the complete HALP does and the solution $\bwhat$
violates its constraints by a small amount.

\begin{proposition}
\label{prop:HALP to relaxed HALP bound} Let $\bwtilde$ be an
optimal solution to the HALP formulation (\ref{eq:HALP}) and
$\bwhat$ be an optimal $\delta$-infeasible solution to its relaxed
formulation (\ref{eq:relaxed HALP}). Then the expected error of
the value function $\Vbwhat$ can be bounded as:
\begin{align*}
  \normw{\Vstar - \Vbwhat}{1, \psi} \leq
  \normw{\Vstar - \Vbwtilde}{1, \psi} + \frac{2 \delta}{1 - \gamma},
\end{align*}
where $\normw{\cdot}{1, \psi}$ is an $\cL_1$-norm weighted by the
state relevance density function $\psi$.
\end{proposition}

\noindent Based on Proposition \ref{prop:HALP to relaxed HALP
bound}, we can generalize our conclusions from Section
\ref{sec:HALP error bounds} to relaxed HALP formulations. For
instance, we may draw a parallel between optimizing the relaxed
objective $\E{\psi}{\Vbwhat}$ and the max-norm error
$\maxnormw{\Vstar - \Vbw}{1 / \lyapunov}$.

\begin{theorem}
\label{thm:relaxed HALP bound} Let $\bwhat$ be an optimal
$\delta$-infeasible solution to a relaxed HALP formulation
(\ref{eq:relaxed HALP}). Then the expected error of the value
function $\Vbwhat$ can be bounded as:
\begin{align*}
  \normw{\Vstar - \Vbwhat}{1, \psi} \leq
  \frac{2 \E{\psi}{\lyapunov}}{1 - \lyapunovfactor}
  \min_\bw \maxnormw{\Vstar - \Vbw}{1 / \lyapunov} +
  \frac{2 \delta}{1 - \gamma},
\end{align*}
where $\normw{\cdot}{1, \psi}$ is an $\cL_1$-norm weighted by the
state relevance density $\psi$, $\lyapunov(\bx) \! = \! \sum_i
w_i^\lyapunov f_i(\bx)$ is a \emph{Lyapunov function} such that
the inequality $\lyapunovfactor \lyapunov(\bx) \geq \gamma
\sup_\ba \E{P(\bx' \mid \bx, \ba)}{\lyapunov(\bx')}$ holds,
$\lyapunovfactor \in [0, 1)$ denotes its \emph{contraction
factor}, and $\maxnormw{\cdot}{1 / \lyapunov}$ is a max-norm
reweighted by the reciprocal $1 / \lyapunov$.
\end{theorem}
{\bf Proof:} Direct combination of Theorem \ref{thm:HALP bound}
and Proposition \ref{prop:HALP to relaxed HALP bound}. \qed

\subsubsection{Grid Resolution}
\label{sec:e-HALP grid resolution}

In Section \ref{sec:e-HALP error bounds}, we bounded the error of
a relaxed HALP formulation by its $\delta$-infeasibility (Theorem
\ref{thm:relaxed HALP bound}), a measure of constraint violation
in the complete HALP. However, it is unclear how the grid
resolution $\eps$ relates to $\delta$-infeasibility. In this
section, we analyze the relationship between $\eps$ and $\delta$.
Moreover, we show how to exploit the factored structure in the
constraint space to achieve the $\delta$-infeasibility of a
relaxed HALP solution $\bwhat$ efficiently.

First, let us assume that $\bwhat$ is an optimal
$\delta$-infeasible solution to an $\eps$-HALP formulation and
$\bZ = \bX \cup \bA$ is the joint set of state and action
variables. To derive a bound relating both $\eps$ and $\delta$, we
assume that the magnitudes of constraint violations
$\tau^{\bwhat}(\bz) = \sum_i \what_i F_i(\bz) - R(\bz)$ are
Lipschitz continuous.

\begin{definition}
\label{def:Lipschitz continuous} The function $f(\bx)$ is
\emph{Lipschitz continuous} if:
\begin{align}
  \abs{f(\bx) - f(\bx')} \leq K \maxnorm{\bx - \bx'}
  \quad \forall \ \bx, \bx' \in \bX;
  \label{eq:Lipschitz continuous}
\end{align}
where $K$ is referred to as a \emph{Lipschitz constant}.
\end{definition}

\noindent Based on the $\eps$-grid discretization of the
constraint space, we know that the distance of any point $\bz$ to
its closest grid point $\bz_G = \arg\min_{\bz'} \maxnorm{\bz -
\bz'}$ is bounded as:
\begin{align}
  \maxnorm{\bz - \bz_G} < \frac{\eps}{2}.
  \label{eq:e-grid distance}
\end{align}
From the Lipschitz continuity of $\tau^{\bwhat}(\bz)$, we
conclude:
\begin{align}
  \abs{\tau^{\bwhat}(\bz_G) - \tau^{\bwhat}(\bz)} \leq
  K \maxnorm{\bz_G - \bz} \leq \frac{K \eps}{2}.
  \label{eq:e-grid bound}
\end{align}
Since every constraint in the relaxed $\eps$-HALP formulation is
satisfied, $\tau^{\bwhat}(\bz_G)$ is nonnegative for all grid
points $\bz_G$. As a result, Equation \ref{eq:e-grid bound} yields
$\tau^{\bwhat}(\bz) > - K \eps / 2$ for every state-action pair
$\bz = (\bx, \ba)$. Therefore, based on Definition
\ref{def:delta-infeasible}, the solution $\bwhat$ is
$\delta$-infeasible for $\delta \geq K \eps / 2$. Conversely, the
$\delta$-infeasibility of $\bwhat$ is guaranteed by choosing $\eps
\leq 2 \delta / K$.

Unfortunately, $K$ may increase rapidly with the dimensionality of a
function. To address this issue, we use the structure in the
constraint space and demonstrate that this is not our case. First,
we observe that the \emph{global Lipschitz constant}
$K_\mathrm{glob}$ is additive in \emph{local Lipschitz constants}
that correspond to the terms $\what_i F_i(\bz)$ and $R_j(\bz)$.
Moreover, $K_\mathrm{glob} \leq N K_\mathrm{loc}$, where $N$ denotes
the total number of the terms and $K_\mathrm{loc}$ is the maximum
over the local constants. Finally, parallel to Equation
\ref{eq:e-grid bound}, the $\delta$-infeasibility of a relaxed HALP
solution $\bwhat$ is achieved by the discretization:
\begin{align}
  \eps \leq \frac{2 \delta}{N K_\mathrm{loc}} \leq
  \frac{2 \delta}{K_\mathrm{glob}}.
  \label{eq:e-grid delta-infeasibility}
\end{align}
Since the factors $\what_i F_i(\bz)$ and $R_j(\bz)$ are often
restricted to small subsets of state and action variables,
$K_\mathrm{loc}$ should change a little when the size of a problem
increases but its structure is fixed. To prove that
$K_\mathrm{loc}$ is bounded, we have to bound the weights
$\what_i$. If all basis functions are of unit magnitude, the
weights $\what_i$ are intuitively bounded as $\abs{\what_i} \leq
(1 - \gamma)^{-1} R_\mathrm{max}$, where $R_\mathrm{max}$ denotes
the maximum one-step reward in the HMDP.

Based on Equation \ref{eq:e-grid delta-infeasibility}, we conclude
that the number of discretization points in a single dimension
$\ceils{1 / \eps + 1}$ is bounded by a polynomial in $N$,
$K_\mathrm{loc}$, and $1 / \delta$. Hence, the constraint space in
the relaxed $\eps$-HALP formulation involves $O([N K_\mathrm{loc} (1
/ \delta)]^{\abs{\bX} + \abs{\bA}})$ constraints, where $\abs{\bX}$
and $\abs{\bA}$ denote the number of state and action variables. The
idea of variable elimination can be used to write the constraints
compactly by $O([N K_\mathrm{loc} (1 / \delta)]^{T + 1} (\abs{\bX} +
\abs{\bA}))$ constraints (Example \ref{ex:admin constraint space}),
where $T$ is the treewidth of a corresponding cost network.
Therefore, satisfying this constraint space is polynomial in $N$,
$K_\mathrm{loc}$, $1 / \delta$, $\abs{\bX}$, and $\abs{\bA}$, but
still exponential in $T$.

\subsection{Cutting Plane Method}
\label{sec:cutting plane method}

\begin{figure}[t]
  \centering
  \rule{\textwidth}{0.01in}
  \vspace{-0.25in}
  \small{
  \begin{tabbing}
    \hspace{0.2in} \= \hspace{0.2in} \= \hspace{0.2in} \= \hspace{0.2in} \= \kill
    {\bf Inputs:} \\
    \> a hybrid factored MDP $\cM = (\bX, \bA, P, R)$ \\
    \> basis functions $f_0(\bx), f_1(\bx), f_2(\bx), \dots$ \\
    \> initial basis function weights $\bw^{(0)}$ \\
    \> a separation oracle $\cO$ \\
    \\
    {\bf Algorithm:} \\
    \> initialize a relaxed HALP formulation with an empty set of constraints \\
    \> $t = 0$ \\
    \> while a stopping criterion is not met \\
    \>\> query the oracle $\cO$ for a violated constraint $(\bx_\cO, \ba_\cO)$
    with respect to $\bw^{(t)}$ \\
    \>\> if the constraint $(\bx_\cO, \ba_\cO)$ is violated \\
    \>\>\> add the constraint to the relaxed HALP \\
    \>\> resolve the LP for a new vector $\bw^{(t + 1)}$ \\
    \>\> $t = t + 1$ \\
    \\
    {\bf Outputs:} \\
    \> basis function weights $\bw^{(t)}$
  \end{tabbing}
  }
  \vspace{-0.15in}
  \rule{\textwidth}{0.01in}
  \caption{Pseudo-code implementation of a HALP solver with the
  cutting plane method.}
  \label{fig:HALP solver}
\end{figure}

Both MC and $\eps$-HALP formulations (Sections \ref{sec:MC-HALP}
and \ref{sec:e-HALP}) approximate the constraint space in HALP by
a finite set of constraints $\cC$. Therefore, they can be solved
directly by any linear programming solver. However, if the number
of constraints is large, formulating and solving LPs with the
complete set of constraints is infeasible. In this section, we
show how to build relaxed HALP approximations efficiently by the
cutting plane method.

The cutting plane method for solving HALP formulations is outlined
in Figure \ref{fig:HALP solver}. Briefly, this approach builds the
set of LP constraints incrementally by adding a violated
constraint to this set in every step. In the remainder of the
paper, we refer to any method that returns a violated constraint
for an arbitrary vector $\bwhat$ as a \emph{separation oracle}.
Formally, every HALP oracle approaches the optimization problem:
\begin{align}
  \arg\min_{\bx, \ba} \left[\Vbwhat(\bx) -
  \gamma \E{P(\bx' \mid \bx, \ba)}{\Vbwhat(\bx')} -
  R(\bx, \ba)\right].
  \label{eq:HALP most violated constraint}
\end{align}
Consequently, the problem of solving hybrid factored MDPs
efficiently reduces to the design of efficient separation oracles.
Note that the cutting plane method (Figure \ref{fig:HALP solver})
can be applied to suboptimal solutions to Equation \ref{eq:HALP
most violated constraint} if these correspond to violated
constraints.

\begin{figure}[t]
  \centering
  \rule{\textwidth}{0.01in}
  \vspace{-0.25in}
  \small{
  \begin{tabbing}
    \hspace{0.2in} \= \hspace{0.2in} \= \hspace{0.2in} \= \hspace{0.2in} \= \kill
    {\bf Inputs:} \\
    \> a hybrid factored MDP $\cM = (\bX, \bA, P, R)$ \\
    \> basis functions $f_0(\bx), f_1(\bx), f_2(\bx), \dots$ \\
    \> basis function weights $\bw$ \\
    \> grid resolution $\eps$ \\
    \\
    {\bf Algorithm:} \\
    \> discretize continuous variables $\bX_C$
    and $\bA_C$ into ($\ceils{1 / \eps + 1}$) equally-spaced values \\
    \> identify subsets $\bX_i$ and $\bA_i$
    ($\bX_j$ and $\bA_j$) corresponding to the domains of
    $F_i(\bx, \ba)$ ($R_j(\bx, \ba)$) \\
    \> evaluate $F_i(\bx_i, \ba_i)$ ($R_j(\bx_j, \ba_j)$) for
    all configurations $\bx_i$ and $\ba_i$ ($\bx_j$ and
    $\ba_j$) on the $\eps$-grid \\
    \> build a cost network for the factored cost function: \\
    \>\> $\tau^\bw(\bx, \ba) =
    \sum_i w_i F_i(\bx, \ba) - R(\bx, \ba)$ \\
    \> find the most violated constraint in the cost network: \\
    \>\> $(\bx_\cO, \ba_\cO) = \arg\min_{\bx, \ba}
    \tau^\bw(\bx, \ba)$ \\
    \\
    {\bf Outputs:} \\
    \> state-action pair $(\bx_\cO, \ba_\cO)$
  \end{tabbing}
  }
  \vspace{-0.15in}
  \rule{\textwidth}{0.01in}
  \caption{Pseudo-code implementation of the $\eps$-HALP
  separation oracle $\cO_\eps$.}
  \label{fig:e-HALP separation oracle}
\end{figure}

The presented approach can be directly used to satisfy the
constraints in relaxed $\eps$-HALP formulations
\shortcite{schuurmans02direct}. Briefly, the solver from Figure
\ref{fig:HALP solver} iterates until no violated constraint is
found and the $\eps$-HALP separation oracle $\cO_\eps$ (Figure
\ref{fig:e-HALP separation oracle}) returns the most violated
constraint in the discretized cost network given an intermediate
solution $\bw^{(t)}$. Note that although the search for the most
violated constraint is polynomial in $\abs{\bX}$ and $\abs{\bA}$
(Section \ref{sec:e-HALP grid resolution}), the running time of
our solver does not have to be \shortcite{guestrin03thesis}. In
fact, the number of generated cuts is exponential in $\abs{\bX}$
and $\abs{\bA}$ in the worst case. However, the same oracle
embedded into the ellipsoid method
\shortcite{khachiyan79polynomial} yields a polynomial-time
algorithm \shortcite{bertsimas97introduction}. Although this
technique is impractical for solving large LPs, we may conclude
that our approach is indeed polynomial-time if implemented in this
particular way.

Finally, note that searching for the most violated constraint
(Equation \ref{eq:HALP most violated constraint}) has application
beyond satisfying the constraint space in HALP. For instance,
computation of a greedy policy for the value function $\Vbwhat$:
\begin{align}
  u(\bx)
  \ & = \ \arg\max_\ba \left[R(\bx, \ba) +
  \gamma \E{P(\bx' \mid \bx, \ba)}{\Vbwhat(\bx')}\right]
  \nonumber \\
  \ & = \ \arg\min_\ba \left[- R(\bx, \ba) -
  \gamma \E{P(\bx' \mid \bx, \ba)}{\Vbwhat(\bx')}\right]
  \label{eq:HALP greedy policy}
\end{align}
is almost an identical optimization problem, where the state
variables $\bX$ are fixed. Moreover, the magnitude of the most
violated constraint is equal to the lowest $\delta$ for which the
relaxed HALP solution $\bwhat$ is $\delta$-infeasible (Equation
\ref{eq:delta-infeasible}):
\begin{align}
  \underline{\delta}
  \ & = \ \min_\bx \left[\Vbwhat(\bx) -
  \max_\ba \left[R(\bx, \ba) +
  \gamma \E{P(\bx' \mid \bx, \ba)}{\Vbwhat(\bx')}\right]\right]
  \nonumber \\
  \ & = \ \min_{\bx, \ba} \left[\Vbwhat(\bx) - R(\bx, \ba) -
  \gamma \E{P(\bx' \mid \bx, \ba)}{\Vbwhat(\bx')}\right].
  \label{eq:delta-infeasible upper bound}
\end{align}

\subsection{MCMC-HALP}
\label{sec:MCMC-HALP}

In practice, both MC and $\eps$-HALP formulations (Sections
\ref{sec:MC-HALP} and \ref{sec:e-HALP}) are built on a blindly
selected set of constraints $\cC$. More specifically, the
constraints in the MC-HALP formulation are chosen randomly (with
respect to a prior distribution $\varphi$) while the $\eps$-HALP
formulation is based on a uniform $\eps$-grid. This discretized
constraint space preserves its original factored structure, which
allows for its compact satisfaction. However, the complexity of
solving the $\eps$-HALP formulation is exponential in the treewidth
of its discretized constraint space. Note that if the discretized
constraint space is represented by binary variables only, the
treewidth increases by a multiplicative factor of $\log_2 \ceils{1 /
\eps + 1}$, where $\ceils{1 / \eps + 1}$ denotes the number of
discretization points in a single dimension. Consequently, even if
the treewidth of a problem is relatively small, solving its
$\eps$-HALP formulation becomes intractable for small values
\mbox{of $\eps$}.

To address the issues of the discussed approximations (Sections
\ref{sec:MC-HALP} and \ref{sec:e-HALP}), we propose a novel Markov
chain Monte Carlo (MCMC) method for finding the most violated
constraint of a relaxed HALP. The procedure directly operates in the
domains of continuous variables, takes into account the structure of
factored MDPs, and its space complexity is proportional to the
number of variables. This separation oracle can be easily embedded
into the ellipsoid or cutting plane method for solving linear
programs (Section \ref{sec:cutting plane method}), and therefore
constitutes a key step towards solving HALP efficiently. Before we
proceed, we represent the constraint space in HALP compactly and
state an optimization problem for finding violated constraints in
this factored representation.

\subsubsection{Compact Representation of Constraints}
\label{sec:MCMC-HALP compact representation}

In Section \ref{sec:ALP}, we showed how the factored
representation of the constraint space allows for its compact
satisfaction. Following this idea, we define \emph{violation
magnitude} $\tau^{\bw}(\bx, \ba)$:
\begin{align}
  \tau^\bw(\bx, \ba)
  \ & = \ - \left[\Vbw(\bx) -
  \gamma \E{P(\bx' \mid \bx, \ba)}{\Vbw(\bx')} -
  R(\bx, \ba)\right] \label{eq:violation magnitude} \\
  \ & = \ - \sum_i w_i [f_i(\bx) -
  \gamma g_i(\bx, \ba)] + R(\bx, \ba), \nonumber
\end{align}
which measures the amount by which the solution $\bw$ violates the
constraints in the complete HALP. We represent the magnitude of
violation $\tau^{\bw}(\bx, \ba)$ compactly by an influence diagram
(ID), where $\bX$ and $\bA$ are decision nodes, and $\bX'$ are
random variables. This representation is built on the transition
model $P(\bX' \mid \bX, \bA)$, which is factored and captures
independencies among the variables $\bX$, $\bX'$, and $\bA$. We
extend the diagram by three types of reward nodes, one for each
term in Equation \ref{eq:violation magnitude}: $H_i = - w_i
f_i(\bx)$ for every basis function, $G_i = \gamma w_i f_i(\bx')$
for every backprojection, and $R_j = R_j(\bx_j, \ba_j)$ for every
local reward function. The construction is completed by adding
arcs that graphically represent the dependencies of the reward
nodes on the variables. Finally, we can verify that:
\begin{align}
  \tau^\bw(\bx, \ba) =
  \E{P(\bx' \mid \bx, \ba)}{\sum_i (H_i + G_i) + \sum_j R_j}.
  \label{eq:violation magnitude graphical}
\end{align}
Consequently, the decision that maximizes the expected utility in
the ID corresponds to the most violated constraint. A graphical
representation of the violation magnitude $\tau^{\bw}(\bx, \ba)$
on the 4-ring network administration problem (Example \ref{ex:CMDP
admin}) is given in Figure \ref{fig:admin constraint space}a. The
structure of the constraint space is identical to Example
\ref{ex:admin constraint space} if the basis functions are
univariate.

We conclude that any algorithm for solving IDs can be applied to
find the most violated constraint. However, most of these methods
\shortcite{cooper88method,jensen94influence,ortiz02thesis} are
restricted to discrete variables. Fortunately, special properties
of the ID representation allow its further simplification. If the
basis functions are chosen conjugate to the transition model
(Section \ref{sec:HALP expectation terms}), we obtain a
closed-form solution to the expectation term $\E{P(\bx' \mid \bx,
\ba)}{G_i}$ (Equation \ref{eq:hybrid backprojection}), and the
random variables $\bX'$ are marginalized out of the diagram. The
new representation contains no random variables and is known as a
cost network (Section \ref{sec:ALP}).

Note that the problem of finding the most violated constraint in the
ID representation is also identical to finding the maximum a
posteriori (MAP) configuration of random variables in Bayesian
networks
\shortcite{dechter96bucket,park01approximating,park03solving,yuan04annealed}.
The latter problem is difficult because of the alternating summation
and maximization operators. Since we marginalized out the random
variables $\bX'$, we can solve the maximization problem by standard
large-scale optimization techniques.

\subsubsection{Separation Oracle $\cO_\mathrm{MCMC}$}
\label{sec:MCMC-HALP separation oracle}

To find the most violated constraint in the cost network, we apply
the Metropolis-Hastings (MH) algorithm
\shortcite{metropolis53equations,hastings70monte} and propose a
Markov chain whose invariant distribution converges to the
vicinity of $\arg\max_\bz \tau^\bw(\bz)$, where $\bz = (\bx, \ba)$
is a value assignment to the joint set of state and action
variables $\bZ = \bX \cup \bA$.

In short, the Metropolis-Hastings algorithm defines a Markov chain
that transits between an existing state $\bz$ and a proposed state
$\bz^\ast$ with the \emph{acceptance probability}:
\begin{align}
  A(\bz, \bz^\ast) =
  \min \set{1,
  \frac{p(\bz^\ast) q(\bz \mid \bz^\ast)}
  {p(\bz) q(\bz^\ast \mid \bz)}},
  \label{eq:acceptance probability}
\end{align}
where $q(\bz^\ast \mid \bz)$ and $p(\bz)$ are a \emph{proposal
distribution} and a \emph{target density}, respectively. Under mild
restrictions on $p(\bz)$ and $q(\bz^\ast \mid \bz)$, the frequency
of state visits generated by the Markov chain always converges to
the target function $p(\bz)$ \shortcite{andrieu03introduction}. In
the remainder of this section, we discuss the choices of $p(\bz)$
and $q(\bz^\ast \mid \bz)$ to solve our optimization
problem.\footnote{For an introduction to Markov chain Monte Carlo
(MCMC) methods, refer to the work of
\shortciteA{andrieu03introduction}.}

\bigskip \noindent {\bf Target density:} The violation magnitude
$\tau^\bw(\bz)$ is turned into a density by the transformation
$p(\bz) = \exp[\tau^\bw(\bz)]$. Due to its monotonic character,
$p(\bz)$ retains the same \mbox{set of global} maxima as
$\tau^\bw(\bz)$. Therefore, the search for $\arg\max_\bz
\tau^\bw(\bz)$ can be done on the new function $p(\bz)$. To prove
that $p(\bz)$ is a density, we demonstrate that $\sum_{\bz_D}
\int_{\bz_C} p(\bz) \ud \bz_C$ is a normalizing constant, where
$\bz_D$ and $\bz_C$ are the discrete and continuous parts of the
value assignment $\bz$. First, note that the integrand $\bz_C$ is
restricted to the space $[0, 1]^{\abs{\bZ_C}}$. As a result, the
integral $\int_{\bz_C} p(\bz) \ud \bz_C$ is proper if $p(\bz)$ is
bounded, and hence it is Riemann integrable and finite. To prove
that $p(\bz) = \exp[\tau^\bw(\bz)]$ is bounded, we bound the
magnitude of violation  $\tau^\bw(\bz)$. If all basis functions
are of unit magnitude, the weights $\what_i$ can be bounded as
$\abs{\what_i} \! \leq \! (1 - \gamma)^{-1} R_\mathrm{max}$
(Section \ref{sec:e-HALP grid resolution}), which in turn yields
the bound $\abs{\tau^\bw(\bz)} \leq (\abs{\bw} (1 - \gamma)^{-1} +
1) R_\mathrm{max}$. Therefore, $p(\bz)$ is bounded and can be
treated as a density function.

To find the mode of $p(\bz)$, we employ simulating annealing
\shortcite{kirkpatrick83optimization} and generate a non-homogeneous
Markov chain whose invariant distribution is equal to $p^{1 /
T_t}(\bz)$, where $T_t$ is a cooling schedule such that $\lim_{t
\rightarrow \infty} T_t = 0$. Under weak regularity assumptions on
$p(\bz)$, $p^\infty(\bz)$ is a probability density that concentrates
on the set of the global maxima of $p(\bz)$
\shortcite{andrieu03introduction}. If our cooling schedule $T_t$
decreases such that $T_t \geq c / \ln(t + 1)$, where $c$ is a
problem-specific constant, the chain from Equation
\ref{eq:acceptance probability} converges to the vicinity of
$\arg\max_\bz \tau^\bw(\bz)$ with the probability converging to 1
\shortcite{geman84stochastic}. However, this logarithmic cooling
schedule is slow in practice, especially for a high initial
temperature $c$. To overcome this problem, we select a smaller value
of $c$ \shortcite{geman84stochastic} than is required by the
convergence criterion. Therefore, the convergence of our chain to
the global optimum $\arg\max_\bz \tau^\bw(\bz)$ is no longer
guaranteed.

\begin{figure}[t]
  \centering
  \rule{\textwidth}{0.01in}
  \vspace{-0.25in}
  \small{
  \begin{tabbing}
    \hspace{0.2in} \= \hspace{0.2in} \= \hspace{0.2in} \= \hspace{0.2in} \= \kill
    {\bf Inputs:} \\
    \> a hybrid factored MDP $\cM = (\bX, \bA, P, R)$ \\
    \> basis functions $f_0(\bx), f_1(\bx), f_2(\bx), \dots$ \\
    \> basis function weights $\bw$ \\
    \\
    {\bf Algorithm:} \\
    \> initialize a state-action pair $\bz^{(t)}$ \\
    \> $t = 0$ \\
    \> while a stopping criterion is not met \\
    \>\> for every variable $Z_i$ \\
    \>\>\> sample $u \sim \unifpdf{0}{1}$ \\
    \>\>\> sample $z_i^\ast \sim p(Z_i \mid \bz_{- i}^{(t)})$ \\
    \>\>\> if $u < \min\left\{1,
    \frac{p^{1 / T_t - 1}(z_i^\ast \mid \bz_{- i}^{(t)})}
    {p^{1 / T_t - 1}(z_i^{(t)} \mid \bz_{- i}^{(t)})}
    \right\}$ \\
    \>\>\>\> $z_i^{(t + 1)} = z_i^\ast$ \\
    \>\>\> else \\
    \>\>\>\> $z_i^{(t + 1)} = z_i^{(t)}$ \\
    \>\> update $T_{t + 1}$ according to the cooling schedule \\
    \>\> $t = t + 1$ \\
    \> $(\bx_\cO, \ba_\cO) = \bz^{(t)}$ \\
    \\
    {\bf Outputs:} \\
    \> state-action pair $(\bx_\cO, \ba_\cO)$
  \end{tabbing}
  }
  \vspace{-0.15in}
  \rule{\textwidth}{0.01in}
  \caption{Pseudo-code implementation of the MCMC-HALP oracle
  $\cO_\mathrm{MCMC}$. The symbol $\unifpdf{0}{1}$ denotes the
  uniform distribution on the interval $[0, 1]$. Since the
  \mbox{testing for} violated constraints (Figure
  \ref{fig:HALP solver}) is inexpensive, our implementation of
  the MCMC-HALP solver in Section \ref{sec:experiments} tests
  all constraints $\bz^{(t)}$ generated by the Markov chain and
  not only the last one. Therefore, the separation oracle
  $\cO_\mathrm{MCMC}$ returns more than one constraint per chain.}
  \label{fig:MCMC-HALP separation oracle}
\end{figure}

\bigskip \noindent {\bf Proposal distribution:} We take advantage
of the factored character of $\bZ$ and adopt the following
proposal distribution \shortcite{geman84stochastic}:
\begin{align}
  q(\bz^\ast \mid \bz) = \left\{
  \begin{array}{l l}
    p(z^\ast_i \mid \bz_{- i}) &
    \mathrm{\hbox{if }} \bz^\ast_{- i} = \bz_{- i} \\
    0 &
    \mathrm{otherwise}
  \end{array}
  \right.,
  \label{eq:proposal distribution}
\end{align}
where $\bz_{- i}$ and $\bz^\ast_{- i}$ are value assignments to
all variables but $Z_i$ in the original and proposed states. If
$Z_i$ is a discrete variable, its conditional:
\begin{align}
  p(z_i^\ast \mid \bz_{- i}) =
  \frac{p(z_1, \dots, z_{i - 1}, z_i^\ast, z_{i + 1}, \dots, z_{n + m})}
  {\sum_{z_i} p(z_1, \dots, z_{i - 1}, z_i, z_{i + 1}, \dots, z_{n + m})}
  \label{eq:discrete conditional}
\end{align}
can be derived in a closed form. If $Z_i$ is a continuous
variable, a closed form of its cumulative density function is
unlikely to exist. To sample from the conditional, we embed
another MH step within the original chain. In the experimental
section, we use the Metropolis algorithm with the acceptance
probability:
\begin{align}
  A(z_i, z_i^\ast) =
  \min \set{1,
  \frac{p(z_i^\ast \mid \bz_{- i})}
  {p(z_i \mid \bz_{- i})}},
  \label{eq:continuous acceptance probability}
\end{align}
where $z_i$ and $z_i^\ast$ are the original and proposed values of
the variable $Z_i$. Note that sampling from both conditionals can
be performed in the space of $\tau^\bw(\bz)$ and locally.

\bigskip Finally, by assuming that $\bz^\ast_{- i} = \bz_{- i}$
(Equation \ref{eq:proposal distribution}), we derive a
non-homogenous Markov chain with the acceptance probability:
\begin{align}
  A(\bz, \bz^\ast)
  \ & = \ \min \set{1,
  \frac{p^{1 / T_t}(\bz^\ast) q(\bz \mid \bz^\ast)}
  {p^{1 / T_t}(\bz) q(\bz^\ast \mid \bz)}} \nonumber \\
  \ & = \ \min \set{1,
  \frac{p^{1 / T_t}(z^\ast_i \mid \bz^\ast_{- i})
  p^{1 / T_t}(\bz^\ast_{- i}) p(z_i \mid \bz^\ast_{- i})}
  {p^{1 / T_t}(z_i \mid \bz_{- i})
  p^{1 / T_t}(\bz_{- i}) p(z^\ast_i \mid \bz_{- i})}} \nonumber \\
  \ & = \ \min \set{1,
  \frac{p^{1 / T_t}(z^\ast_i \mid \bz_{- i})
  p^{1 / T_t}(\bz_{- i}) p(z_i \mid \bz_{- i})}
  {p^{1 / T_t}(z_i \mid \bz_{- i})
  p^{1 / T_t}(\bz_{- i}) p(z^\ast_i \mid \bz_{- i})}} \nonumber \\
  \ & = \ \min \set{1,
  \frac{p^{1 / T_t - 1}(z_i^\ast \mid \bz_{- i})}
  {p^{1 / T_t - 1}(z_i \mid \bz_{- i})}},
  \label{eq:annealing acceptance probability}
\end{align}
which converges to the vicinity of the most violated constraint.
\shortciteA{yuan04annealed} proposed a similar chain for finding
the MAP configuration of random variables in Bayesian networks.

\subsubsection{Constraint Satisfaction}
\label{sec:MCMC-HALP constraint satisfaction}

If the MCMC-HALP separation oracle $\cO_\mathrm{MCMC}$ (Figure
\ref{fig:MCMC-HALP separation oracle}) converges to a violated
constraint (not necessarily the most violated) in polynomial time,
the ellipsoid method is guaranteed to solve HALP formulations in
polynomial time \shortcite{bertsimas97introduction}. Unfortunately,
convergence of our chain within arbitrary precision requires an
exponential number of steps \shortcite{geman84stochastic}. Although
the bound is loose to be of practical interest, it suggests that the
time complexity of proposing violated constraints dominates the time
complexity of solving relaxed HALP formulations. Therefore, the
oracle $\cO_\mathrm{MCMC}$ should search for violated constraints
efficiently. Convergence speedups that directly apply to our work
include hybrid Monte Carlo (HMC) \shortcite{duane87hybrid},
Rao-Blackwellization \shortcite{casella96raoblackwellisation}, and
slice sampling \shortcite{higdon98auxiliary}.

\section{Experiments}
\label{sec:experiments}

Experimental section is divided in three parts. First, we show
that HALP can solve a simple HMDP problem at least as efficiently
as alternative approaches. Second, we demonstrate the scale-up
potential of our framework and compare several approaches to
satisfy the constraint space in HALP (Section \ref{sec:HALP
constraint space}). Finally, we argue for solving our constraint
satisfaction problem in the domains of continuous variables
without discretizing them.

All experiments are performed on a Dell Precision 380 workstation
with 3.2GHz Pentium 4 CPU and 2GB RAM. Linear programs are solved
by the simplex method in the LP\_SOLVE package. The expected
return of policies is estimated by the Monte Carlo simulation of
100 trajectories. The results of randomized methods are
additionally averaged over 10 randomly initialized runs. Whenever
necessary, we present errors on the expected values. These errors
correspond to the standard deviations of measured quantities. The
discount factor $\gamma$ is 0.95.

\subsection{A Simple Example}
\label{sec:experiments simple example}

To illustrate the ability of HALP to solve factored MDPs, we
compare it to $\cL_2$ (Figure \ref{fig:L2 VI}) and grid-based
value iteration (Section \ref{sec:solving HMDPs}) on the 4-ring
topology of the network administration problem (Example
\ref{ex:CMDP admin}). Our experiments are conducted on uniform and
non-uniform grids of varying sizes. Grid points are kept fixed for
all compared methods, which allows for their fair comparison. Both
value iteration methods are iterated for 100 steps and terminated
earlier if their Bellman error drops below $10^{-6}$. Both the
$\cL_2$ and HALP methods approximate the optimal value function
$\Vstar$ by a linear combination of basis functions, one for each
computer $X_i$ ($f_i(\bx) \! = \! x_i$), and one for every
connection $X_i \rightarrow X_j$ in the ring topology ($f_{i
\rightarrow j}(\bx) \! = \! x_i x_j$). We assume that our basis
functions are sufficient to derive a one-step lookahead policy
that reboots the least efficient computer. We believe that such a
policy is close-to-optimal in the ring topology. The constraint
space in the complete HALP formulation is approximated by its
MC-HALP and $\eps$-HALP formulations (Sections \ref{sec:MC-HALP}
and \ref{sec:e-HALP}). The state relevance density function
$\psi(\bx)$ is uniform. Our experimental results are reported in
Figure \ref{fig:CMDP admin results}.

To verify that our solutions are non-trivial, we compare them to
three heuristic policies: dummy, random, and server. The dummy
policy $\pi_\mathrm{dummy}(\bx) = a_5$ always takes the dummy action
$a_5$. Therefore, it establishes a lower bound on the performance of
any administrator. The random policy behaves randomly. The server
policy $\pi_\mathrm{server}(\bx) = a_1$ protects the server $X_1$.
The performance of our heuristics is shown in Figure \ref{fig:CMDP
admin results}. Assuming that we can reboot all computers at each
time step, a utopian upper bound on the performance of any
\mbox{policy $\pi$} can be derived as:
\begin{align}
  \E{\pi}{\sum_{t = 0}^\infty \gamma^t R(\bx_t, \pi(\bx_t))}
  \ & \leq \ \frac{1}{1 - \gamma} \max_{\bx, a}
  \E{P(\bx' \mid \bx, a)}{\max_{a'} R(\bx', a')} \nonumber \\
  \ & = \ \frac{1}{1 - \gamma} \max_{\bx, a}
  \int_{\bx'} 2 P(x_1' \mid \bx, a) x_1'^2 +
  \sum_{j = 2}^4 P(x_j' \mid \bx, a) x_j'^2 \ud \bx' \nonumber \\
  \ & \leq \ \frac{5}{1 - \gamma} \int_{x'}
  \betapdf(x' \mid 20, 2) x'^2 \ud x' \nonumber \\
  \ & \approx \ 83.0.
  \label{eq:CMDP admin upper bound}
\end{align}
We do not analyze the quality of HALP solutions with respect to
the optimal value function $\Vstar$ (Section \ref{sec:HALP error
bounds}) because this one is unknown.

\begin{figure}[t]
  \centering
  {\small
  \begin{tabular}{|l@{\ }|r@{\ }|r@{\ }|r@{\ }|r@{\ }|r@{\ }|r@{\ }|r@{\ }|r@{\ }|r@{\ }|}
    \multicolumn{4}{c}{} &
    \multicolumn{6}{c}{\textbf{Uniform $\eps$-grid}} \\ \cline{5-10}
    \multicolumn{4}{c|}{} &
    \multicolumn{2}{|c|}{\textbf{$\eps$-HALP}} &
    \multicolumn{2}{|c|}{\textbf{$\cL_2$ VI}} &
    \multicolumn{2}{|c|}{\textbf{Grid-based VI}} \\ \cline{3-10}
    \multicolumn{2}{c|}{} &
    $\eps$ & $N$ & Reward & Time & Reward & Time & Reward & Time \\ \cline{3-10}
    \multicolumn{2}{c|}{} &
    $1$ & $8$ & $52.1 \pm 2.2$ & $< 1$ &
    $52.1 \pm 2.2$ & $2$ & & \\
    \multicolumn{2}{c|}{} &
    $1 / 2$ & $91$ & $52.1 \pm 2.2$ & $< 1$ &
    $52.1 \pm 2.2$ & $7$ & $47.6 \pm 2.2$ & $< 1$ \\
    \multicolumn{2}{c|}{} &
    $1 / 4$ & $625$ & $52.1 \pm 2.2$ & $< 1$ &
    $52.1 \pm 2.2$ & $55$ & $51.5 \pm 2.2$ & $20$ \\
    \multicolumn{2}{c|}{} &
    $1 / 8$ & $6\ 561$ & $52.1 \pm 2.2$ & $2$ &
    $52.1 \pm 2.2$ & $577$ & $52.0 \pm 2.3$ & $2\ 216$ \\ \cline{3-10}
    \multicolumn{8}{c}{} \\
    \multicolumn{4}{c}{} &
    \multicolumn{6}{c}{\textbf{Non-uniform grid}} \\ \cline{5-10}
    \multicolumn{2}{c}{\textbf{Heuristics}} &
    \multicolumn{2}{c}{} &
    \multicolumn{2}{|c|}{\textbf{MC-HALP}} &
    \multicolumn{2}{|c|}{\textbf{$\cL_2$ VI}} &
    \multicolumn{2}{|c|}{\textbf{Grid-based VI}} \\ \cline{1-2} \cline{4-10}
    Policy & Reward &
    & $N$ & Reward & Time & Reward & Time & Reward & Time \\ \cline{1-2} \cline{4-10}
    Dummy & $25.0 \pm 2.8$ & &
    $10$ & $45.2 \pm 5.1$ & $< 1$ &
    $45.9 \pm 5.8$ & $1$ & $47.5 \pm 2.8$ & $< 1$ \\
    Random & $42.1 \pm 3.3$ & &
    $50$ & $50.2 \pm 2.4$ & $< 1$ &
    $51.8 \pm 2.2$ & $4$ & $48.7 \pm 2.5$ & $< 1$ \\
    Server & $47.6 \pm 2.2$ & &
    $250$ & $51.5 \pm 2.4$ & $< 1$ &
    $51.9 \pm 2.2$ & $22$ & $50.4 \pm 2.3$ & $2$ \\
    Utopian & $83.0 \qquad \: \;$ & &
    $1\ 250$ & $51.8 \pm 2.3$ & $< 1$ &
    $51.9 \pm 2.2$ & $110$ & $51.6 \pm 2.2$ & $60$ \\ \cline{1-2} \cline{4-10}
  \end{tabular}
  }
  \caption{Comparison of three approaches to solving hybrid MDPs
  on the 4-ring topology of the network administration problem
  (Example \ref{ex:CMDP admin}). The methods are compared on
  uniform and non-uniform grids of varying size ($N$) by
  the expected discounted reward of policies and their computation
  time (in seconds).}
  \label{fig:CMDP admin results}
\end{figure}

Based on our results, we draw the following conclusions. First,
grid-based value iteration is not practical for solving hybrid
optimization problems of even small size. The main reason is the
space complexity of the method, which is quadratic in the number
of grid points $N$. If the state space is discretized uniformly,
$N$ is exponential in the number of state variables. Second, the
quality of the HALP policies is close to the $\cL_2$ VI policies.
This result is positive since $\cL_2$ value iteration is commonly
applied in approximate dynamic programming. Third, both the
$\cL_2$ and HALP approaches yield better policies than grid-based
value iteration. This result is due to the quality of our value
function estimator. Its extremely good performance for $\eps = 1$
can be explained from the monotonicity of the reward and basis
functions. Finally, the computation time of the $\cL_2$ VI
policies is significantly longer than the computation time of the
HALP policies. Since a step of $\cL_2$ value iteration (Figure
\ref{fig:L2 VI}) is as hard as formulating a corresponding relaxed
HALP, this result comes at no surprise.

\subsection{Scale-up Potential}
\label{sec:experiments scale-up potential}

To illustrate the scale-up potential of HALP, we apply three
relaxed HALP approximations (Section \ref{sec:HALP constraint
space}) to solve two irrigation network problems of varying
complexity. These problems are challenging for state-of-the-art
MDP solvers due to the factored state and action spaces.

\begin{figure}[t]
  \centering
  \begin{tabular}{@{\!\!\!}c@{\!\!\!}c@{\!\!\!}c@{\!\!\!}}
    \includegraphics[width=1.71in, bb=0in 0in 12.361in 13.681in]{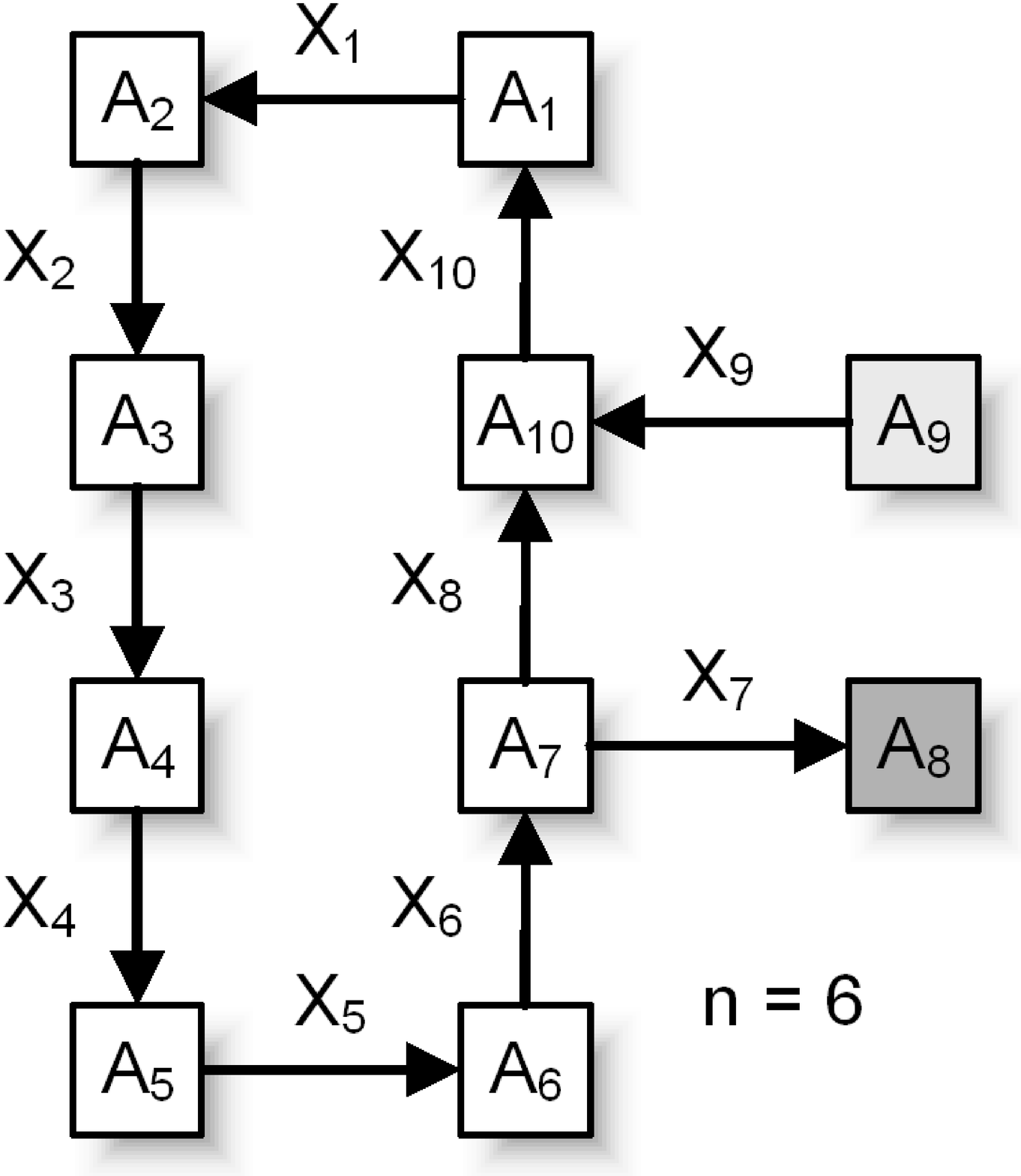} &
    \includegraphics[width=1.71in, bb=0in 0in 12.361in 13.681in]{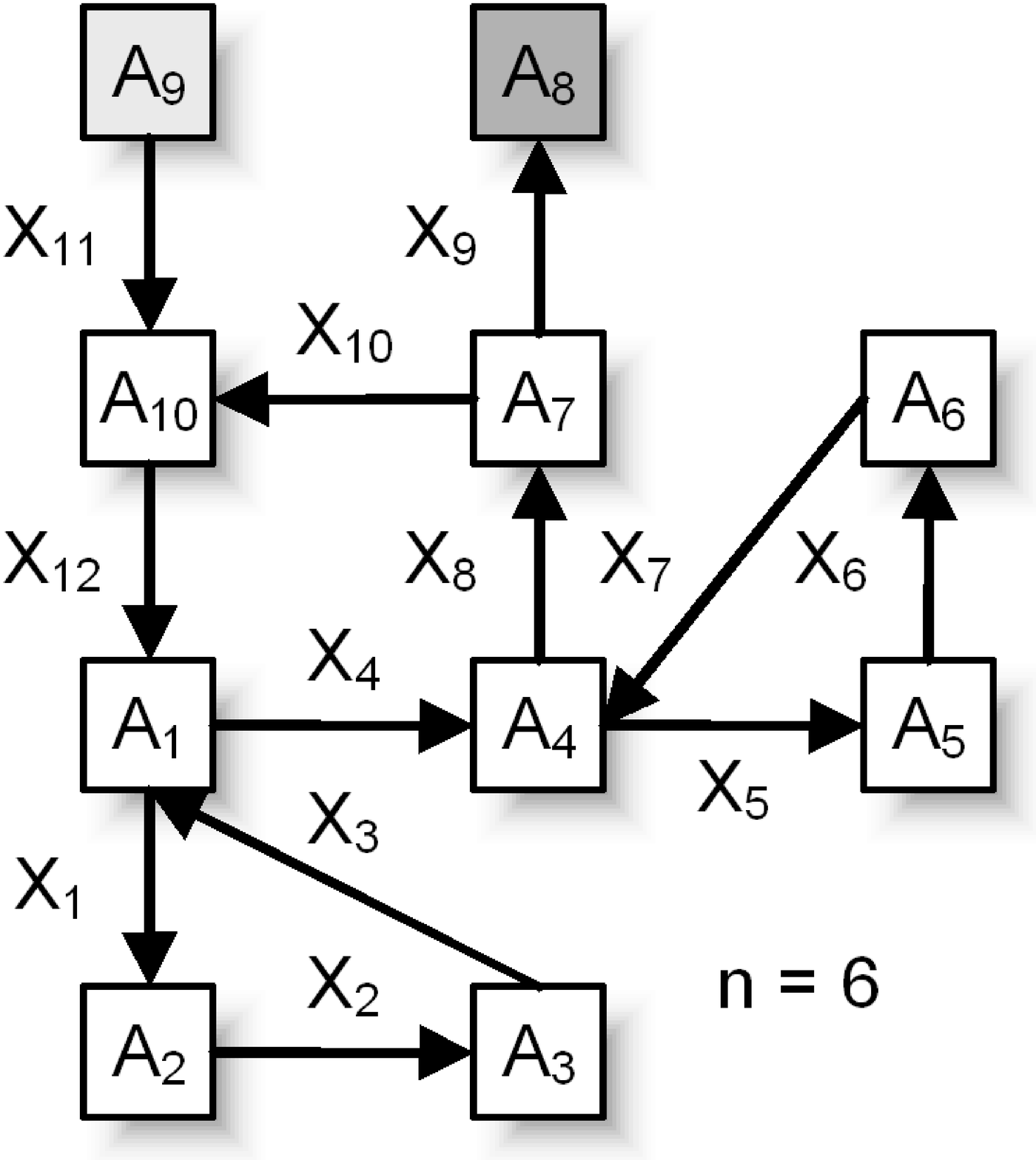} &
    \includegraphics[width=2.61in, bb=0in 0in 18.889in 13.681in]{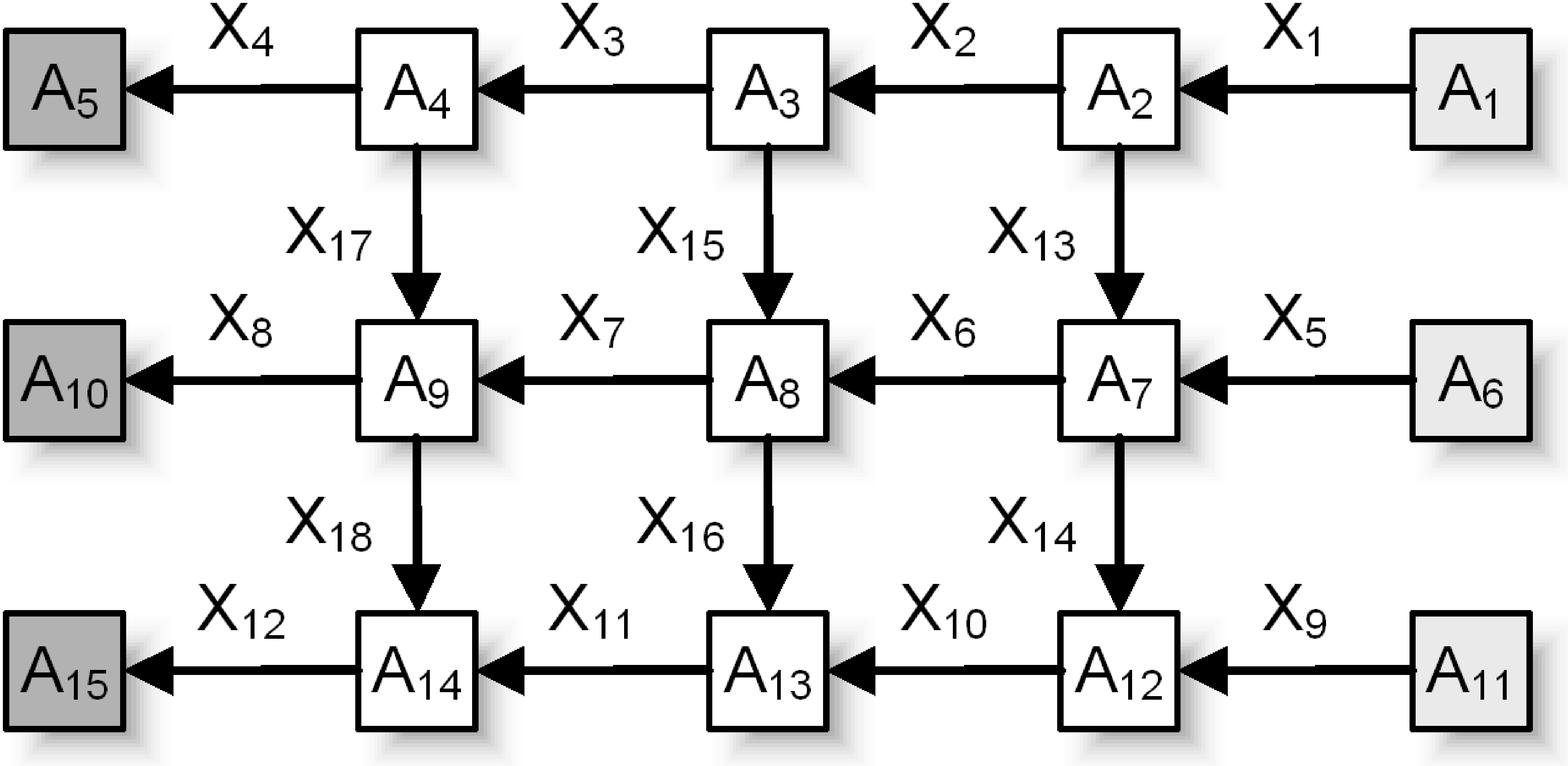} \\
    $\ \ $\textbf{(a)} &
    $\ \ $\textbf{(b)} &
    \textbf{(c)}$\ \ $
  \end{tabular}
  \caption{Illustrations of three irrigation network topologies:
  \textbf{a.} 6-ring, \textbf{b.} 6-ring-of-rings, and \textbf{c.}
  $3 \times 3$ grid. Irrigation channels and their regulation
  devices are represented by arrows and rectangles. Inflow and
  outflow nodes are colored in light and dark gray. The ring and
  ring-of-rings networks are parameterized by the total number of
  regulation devices except for the last four ($n$).}
  \label{fig:irrigation network topologies}
\end{figure}

\begin{figure}[t]
  \centering
  \begin{tabular}{@{\!\!\!}c@{\qquad}c@{\!\!\!}}
    \includegraphics[width=1.71in, bb=0in 0in 12.361in 16.944in]{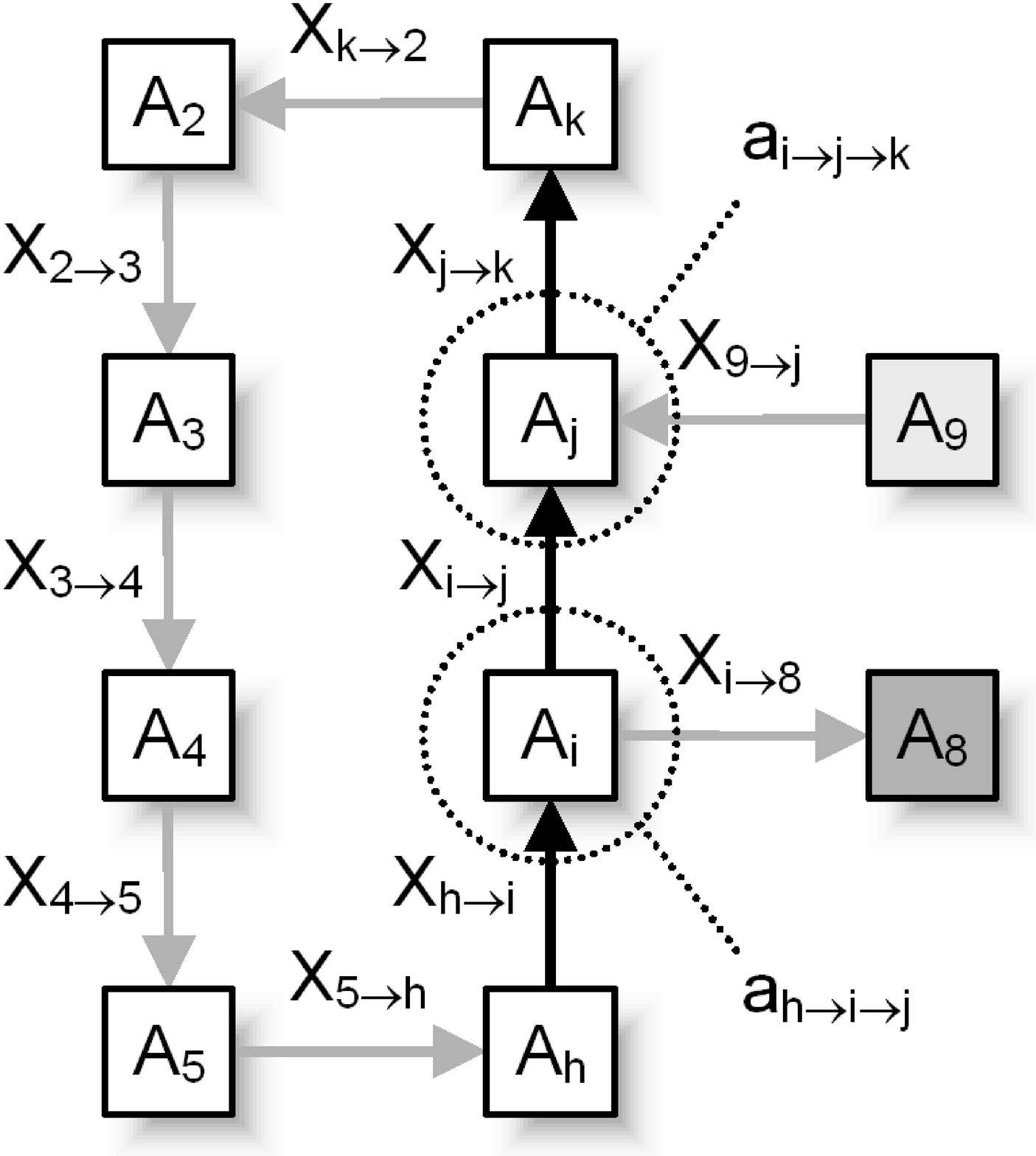} &
    \includegraphics[width=4in, bb=2in 4in 7in 7in]{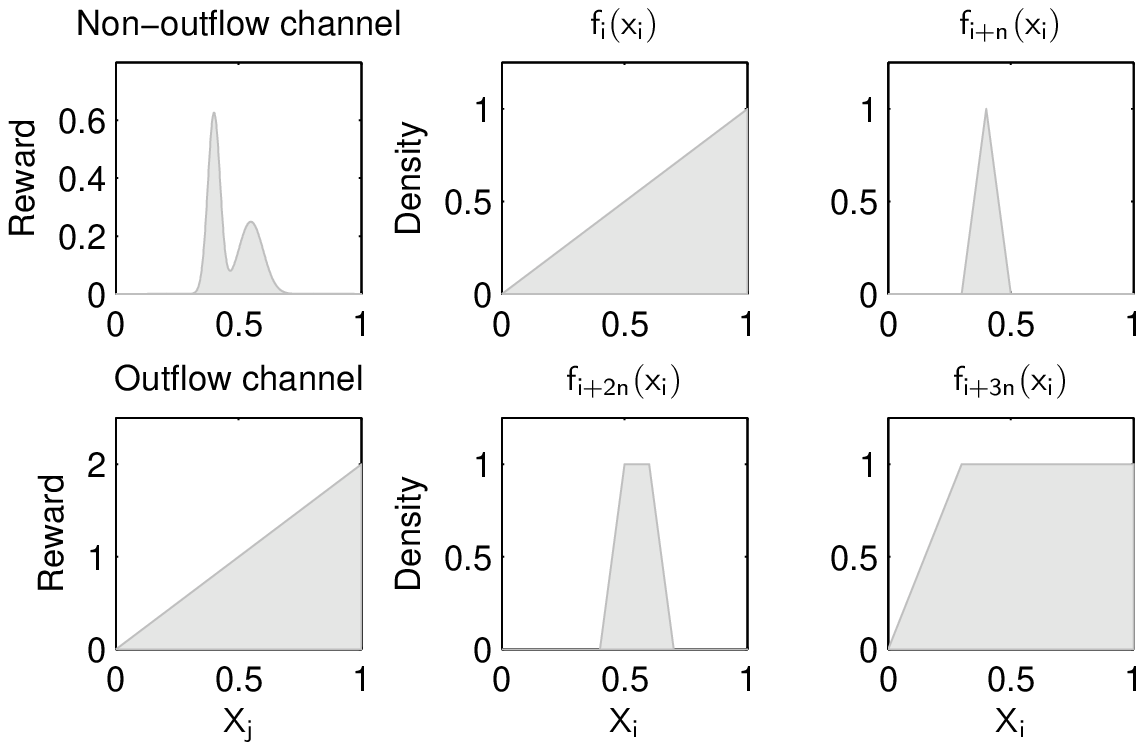} \\
    $\ \ \ \! \!$\textbf{(a)} &
    \textbf{(b)}$\qquad \qquad \qquad \qquad \qquad \quad \! \! \!$
    \textbf{(c)}$\qquad \qquad \ \ \ $
  \end{tabular}
  \caption{\textbf{a.} Indexing used in the description of
  the transition function in Example
  \ref{ex:irrigation network}. The parameters $h$, $i$, $j$, and
  $k$ are equal to 6, 7, 10, and 1, respectively. \textbf{b.}
  Univariate reward functions over water levels $X_j$ (Example
  \ref{ex:irrigation network}).\hspace{-0.0075in} \textbf{c.\hspace{-0.0075in}}
  Univariate basis functions over water levels $X_i$.}
  \label{fig:irrigation network model}
\end{figure}

\begin{example}[Irrigation network operator]
\label{ex:irrigation network} An irrigation network is a system of
irrigation channels connected by regulation devices (Figure
\ref{fig:irrigation network topologies}). The goal of an irrigation
network operator is to route water between the channels to optimize
water levels in the whole system. The optimal levels are determined
by the type of a planted crop. For simplicity of exposition, we
assume that all irrigation channels are oriented and of the same
size.

This optimization problem can be formulated as a factored MDP. The
state of the network is completely observable and represented by
$n$ continuous variables $\bX = \set{X_1, \dots, X_n}$, where the
variable $X_i$ denotes the water level in the $i$-th channel. At
each time step, the irrigation network operator regulates $m$
devices $A_i$ that pump water between every pair of their inbound
and outbound channels. The operation modes of these devices are
described by discrete action variables $\bA = \set{A_1, \dots,
A_m}$. Inflow and outflow devices (no inbound or outbound
channels) are not controlled and just pump water in and out of the
network.

The transition model reflects water flows in the irrigation
network and is encoded locally by conditioning on the operation
modes $\bA$:
\begin{align*}
  P(X_{i \rightarrow j}' = x \mid \Parents(X_{i \rightarrow j}'))
  \ & \propto \ \betapdf(x \mid \alpha, \beta) \quad
  \begin{array}{|l l}
    & \alpha = 46 \mu_{i \rightarrow j}' + 2 \\
    & \beta = 46 (1 - \mu_{i \rightarrow j}') + 2
  \end{array} \\
  \mu_{i \rightarrow j}' \ & = \ \mu_{i \rightarrow j} + \sum_h
  \I{a_{h \rightarrow i \rightarrow j}}{A_i}
  \min (1 - \mu_{i \rightarrow j},
  \min (x_{h \rightarrow i}, \tau_i)) \\
  \mu_{i \rightarrow j} \ & = \ x_{i \rightarrow j} - \sum_k
  \I{a_{i \rightarrow j \rightarrow k}}{A_j}
  \min (x_{i \rightarrow j}, \tau_j)
\end{align*}
where $X_{i \rightarrow j}$ represents the water level between the
regulation devices $A_i$ and $A_j$, $\I{a_{h \rightarrow i
\rightarrow j}}{A_i}$ and $\I{a_{i \rightarrow j \rightarrow
k}}{A_j}$ denote the indicator functions of water routing actions
$a_{h \rightarrow i \rightarrow j}$ and $a_{i \rightarrow j
\rightarrow k}$ at the devices $A_i$ and $A_j$, and $\tau_i$ and
$\tau_j$ are the highest tolerated flows through these devices. In
short, this transition model conserves water mass in the network
and adds some variance to the resulting state $X_{i \rightarrow
j}'$. The introduced indexing of state and action variables is
explained on the 6-ring irrigation network in Figure
\ref{fig:irrigation network model}a. In the rest of the paper, we
assume an inflow of 0.1 to any inflow device $A_i$ ($\tau_i =
0.1$), an outflow of 1 from any outflow device $A_j$ ($\tau_j =
1$), and the highest tolerated flow of $1 / 3$ at the remaining
devices $A_k$ ($\tau_k = 1 / 3$).

The reward function $R(\bx, \ba) = \sum_j R_j(x_j)$ is factored
along individual irrigation channels and described by the
univariate function:
\begin{align*}
  R_j(x_j) = 2 x_j
\end{align*}
for each outflow channel (one of its regulation devices must be
outflow), and by the function:
\begin{align*}
  R_j(x_j) = \frac{\normalpdf(x_j \mid 0.4, 0.025)}{25.6} +
  \frac{\normalpdf(x_j \mid 0.55, 0.05)}{32}
\end{align*}
for the remaining channels (Figure \ref{fig:irrigation network
model}b). Therefore, we reward both for maintaining optimal water
levels and pumping water out of the irrigation network. Several
examples of irrigation network topologies are shown in Figure
\ref{fig:irrigation network topologies}.
\end{example}

\noindent Similarly to Equation \ref{eq:CMDP admin upper bound},
we derive a utopian upper bound on the performance of any policy
$\pi$ in an arbitrary irrigation network as:
\begin{align}
  \E{\pi}{\sum_{t = 0}^\infty \gamma^t R(\bx_t, \pi(\bx_t))}
  \ \leq & \ \ \frac{1}{1 - \gamma}
  \Bigg[0.2 n_\mathrm{in} + (n - n_\mathrm{out}) \nonumber \\
  & \ \ \max_x \int_{x'} \betapdf(x' \mid 46 x + 2, 46 (1 - x) + 2) R(x')
  \ud x'\Bigg],
  \label{eq:irrigation network upper bound}
\end{align}
where $n$ is the total number of irrigation channels,
$n_\mathrm{in}$ and $n_\mathrm{out}$ denote the number of inflow
and outflow channels, respectively, and $R(x') \! = \!
\normalpdf(x' \! \mid 0.4, 0.025) / 25.6 + \normalpdf(x' \! \mid
0.55, 0.05) / 32$. We do not analyze the quality of HALP solutions
with respect to the optimal value function $\Vstar$ (Section
\ref{sec:HALP error bounds}) because this one is unknown.

\begin{figure}[t]
  \centering
  {\small
  \begin{tabular}{|l@{}r@{\ }|r@{\ }|r@{\ }|r@{\ }|r@{\ }|r@{\ }|r@{\ }|r@{\ }|r@{\ }|r@{\ }|} \cline{3-11}
    \multicolumn{2}{c|}{\textbf{Ring}} & \multicolumn{3}{|c|}{$n = 6$} &
    \multicolumn{3}{|c|}{$n = 12$} &
    \multicolumn{3}{|c|}{$n = 18$} \\ \cline{3-11}
    \multicolumn{2}{c|}{\textbf{topology}} & OV & Reward & Time &
    OV & Reward & Time &
    OV & Reward & Time \\ \hline
    \textbf{$\eps$-HALP} & $1 / 4$ &
    $24.3$ & $34.6 \pm 2.0$ & $11$ &
    $36.2$ & $53.9 \pm 2.7$ & $44$ &
    $48.0$ & $74.3 \pm 2.9$ & $87$ \\
    $\quad \eps =$ & $1 / 8$ &
    $55.4$ & $39.6 \pm 2.5$ & $41$ &
    $88.1$ & $61.5 \pm 3.5$ & $107$ &
    $118.8$ & $84.3 \pm 3.8$ & $178$ \\
    & $1 / 16$ &
    $59.1$ & $40.3 \pm 2.6$ & $281$ &
    $93.2$ & $62.6 \pm 3.4$ & $665$ &
    $126.1$ & $86.3 \pm 3.8$ & $1\ 119$ \\ \hline
    \textbf{MCMC} & $10$ &
    $60.9$ & $30.3 \pm 4.9$ & $38$ &
    $86.3$ & $47.6 \pm 6.3$ & $62$ &
    $109.5$ & $56.8 \pm 7.4$ & $87$ \\
    $\quad N =$ & $50$ &
    $70.1$ & $40.2 \pm 2.6$ & $194$ &
    $110.3$ & $62.4 \pm 3.5$ & $328$ &
    $148.8$ & $85.0 \pm 3.6$ & $483$ \\
    & $250$ &
    $70.7$ & $40.2 \pm 2.6$ & $940$ &
    $112.0$ & $63.0 \pm 3.4$ & $1\ 609$ &
    $151.7$ & $85.4 \pm 3.8$ & $2\ 280$ \\ \hline
    \textbf{MC} & $10^2$ &
    $16.2$ & $25.0 \pm 5.1$ & $< 1$ &
    $16.9$ & $41.9 \pm 5.6$ & $< 1$ &
    $17.2$ & $51.8 \pm 8.8$ & $< 1$ \\
    $\quad N =$ & $10^4$ &
    $40.8$ & $37.9 \pm 2.8$ & $10$ &
    $52.8$ & $58.8 \pm 3.5$ & $18$ &
    $63.8$ & $75.9 \pm 6.6$ & $31$ \\
    & $10^6$ &
    $51.2$ & $39.4 \pm 2.7$ & $855$ &
    $67.1$ & $60.3 \pm 3.8$ & $1\ 415$ &
    $81.1$ & $82.9 \pm 3.8$ & $1\ 938$ \\ \hline
    Utopian & &
    & $49.1 \qquad \: \;$ & &
    & $79.2 \qquad \: \;$ & &
    & $109.2 \qquad \: \;$ & \\ \hline
    \multicolumn{11}{c}{} \\ \cline{3-11}
    \multicolumn{2}{c|}{\textbf{Ring-of-rings}} & \multicolumn{3}{|c|}{$n = 6$} &
    \multicolumn{3}{|c|}{$n = 12$} &
    \multicolumn{3}{|c|}{$n = 18$} \\ \cline{3-11}
    \multicolumn{2}{c|}{\textbf{topology}} & OV & Reward & Time &
    OV & Reward & Time &
    OV & Reward & Time \\ \hline
    \textbf{$\eps$-HALP} & $1 / 4$ &
    $28.4$ & $40.4 \pm 2.5$ & $85$ &
    $44.1$ & $66.5 \pm 3.2$ & $382$ &
    $59.8$ & $93.0 \pm 3.8$ & $931$ \\
    $\quad \eps =$ & $1 / 8$ &
    $65.4$ & $47.5 \pm 3.0$ & $495$ &
    $107.9$ & $76.1 \pm 4.1$ & $2\ 379$ &
    $148.8$ & $105.3 \pm 4.2$ & $5\ 877$ \\
    & $1 / 16$ &
    $68.9$ & $47.0 \pm 2.9$ & $4\ 417$ &
    $113.1$ & $77.3 \pm 4.2$ & $19\ 794$ &
    $156.9$ & $107.8 \pm 4.1$ & $53\ 655$ \\ \hline
    \textbf{MCMC} & $10$ &
    $66.9$ & $35.3 \pm 6.1$ & $60$ &
    $94.6$ & $54.4 \pm 9.4$ & $107$ &
    $110.6$ & $47.8 \pm 13.2$ & $157$ \\
    $\quad N =$ & $50$ &
    $80.9$ & $47.1 \pm 2.9$ & $309$ &
    $131.9$ & $76.6 \pm 3.6$ & $571$ &
    $181.4$ & $104.6 \pm 4.4$ & $859$ \\
    & $250$ &
    $81.7$ & $47.2 \pm 2.9$ & $1\ 522$ &
    $134.1$ & $77.3 \pm 3.5$ & $2\ 800$ &
    $186.0$ & $106.6 \pm 3.9$ & $4\ 291$ \\ \hline
    \textbf{MC} & $10^2$ &
    $13.7$ & $31.0 \pm 4.9$ & $< 1$ &
    $15.4$ & $46.1 \pm 6.4$ & $< 1$ &
    $16.8$ & $66.6 \pm 9.4$ & $1$ \\
    $\quad N =$ & $10^4$ &
    $44.3$ & $43.3 \pm 3.2$ & $12$ &
    $59.0$ & $68.9 \pm 5.4$ & $26$ &
    $71.5$ & $92.2 \pm 6.8$ & $49$ \\
    & $10^6$ &
    $55.8$ & $45.1 \pm 3.1$ & $1\ 026$ &
    $75.1$ & $74.3 \pm 3.8$ & $1\ 738$ &
    $92.0$ & $103.1 \pm 4.2$ & $2\ 539$ \\ \hline
    Utopian & &
    & $59.1 \qquad \: \;$ & &
    & $99.2 \qquad \: \;$ & &
    & $139.3 \qquad \: \;$ & \\ \hline
  \end{tabular}
  }
  \caption{Comparison of three HALP solvers on two
  irrigation network topologies of varying sizes ($n$). The
  solvers are compared by the objective value of a relaxed HALP
  (OV), the expected discounted reward of a corresponding policy,
  and computation time (in seconds). The $\eps$-HALP, MCMC-HALP,
  and MC-HALP solvers are parameterized by the resolution of
  $\eps$-grid ($\eps$), the number of MCMC chains ($N$), and the
  number of samples ($N$). Note that the quality of policies
  improves with higher grid resolution ($1 / \eps$) and larger
  sample size ($N$). Upper bounds on their expected returns are
  shown in the last rows of the tables.}
  \label{fig:ring and rors results}
\end{figure}

\begin{figure}[t]
  \centering
  $\qquad$\textbf{Ring topology}
  \includegraphics[width=6.4in, bb=0.45in 4.0in 8.45in 7.0in]{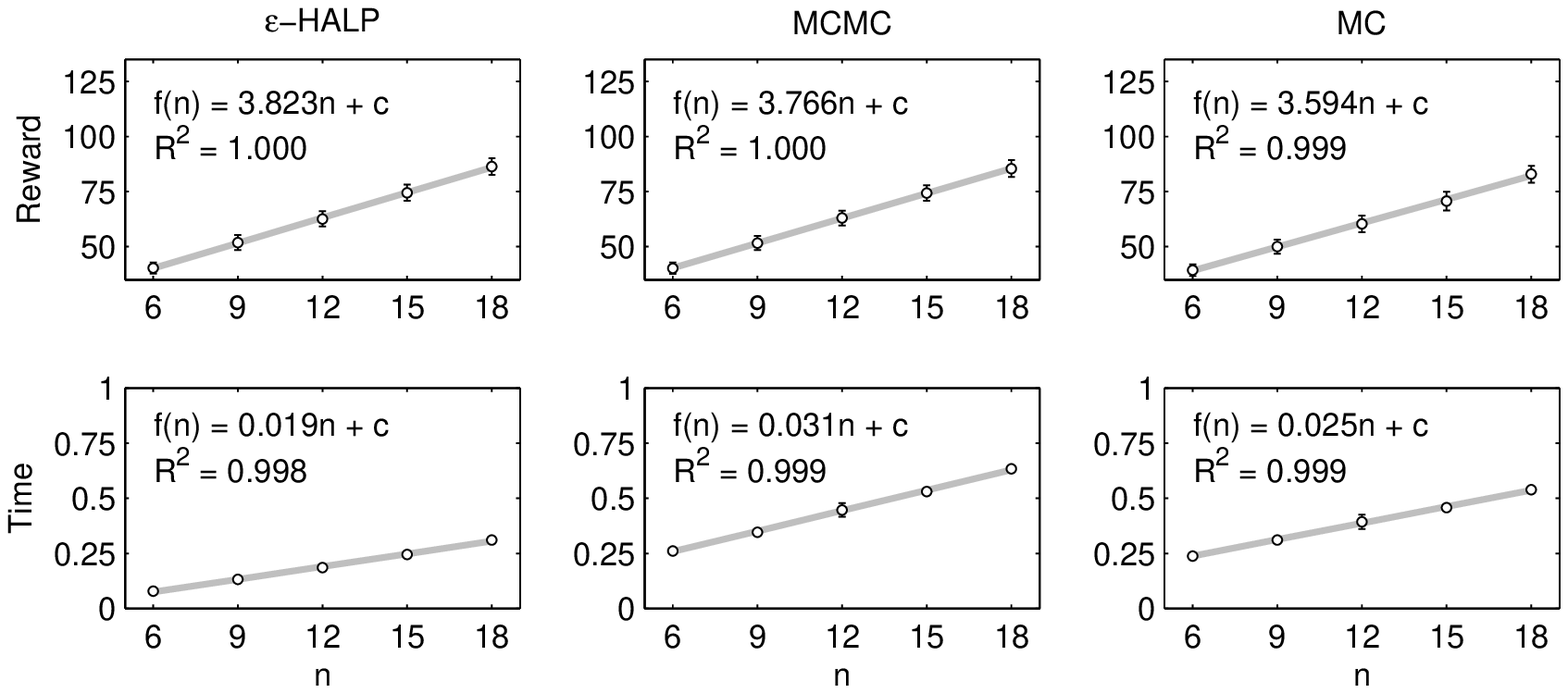} \\
  $\quad \ \ \ $\textbf{Ring-of-rings topology}
  \includegraphics[width=6.4in, bb=0.45in 4.0in 8.45in 7.0in]{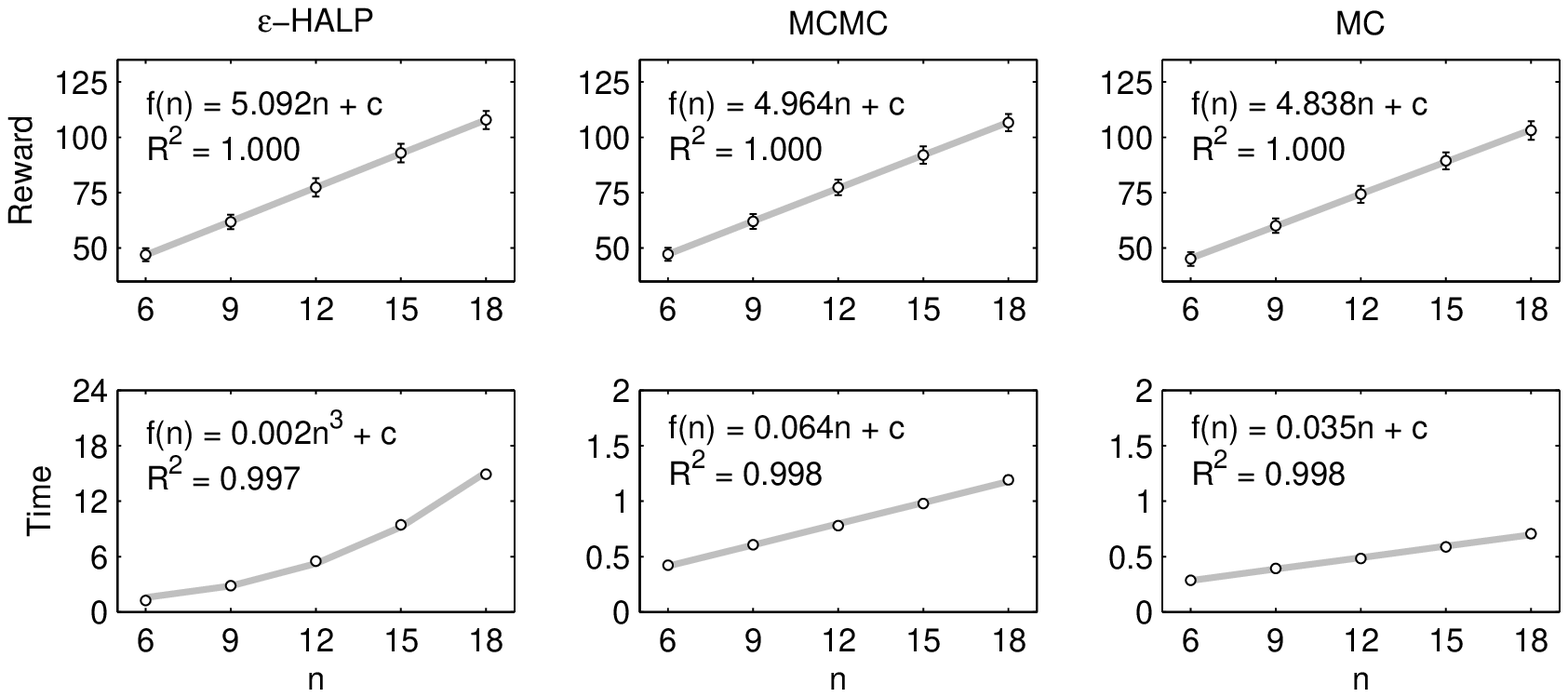}
  \caption{Scale-up potential of the $\eps$-HALP, MCMC-HALP, and
  MC-HALP solvers on two irrigation network topologies of varying
  sizes ($n$). The graphs show the expected discounted reward of
  policies and their computation time (in hours) as functions of
  $n$. The HALP solvers are parameterized by the resolution of
  $\eps$-grid ($\eps = 1 / 16$), the number of MCMC chains
  ($N = 250$), and the number of samples ($N = 10^6$). Note that
  all trends can be approximated by a polynomial $f(n)$ (gray
  line) with a high degree of confidence (the coefficient of
  determination $R^2$), where $c$ denotes a constant independent
  of $n$.}
  \label{fig:ring and rors trends}
\end{figure}

\begin{figure}[t]
  \centering
  \includegraphics[width=6.4in, bb=0.45in 4.3in 8.45in 6.8in]{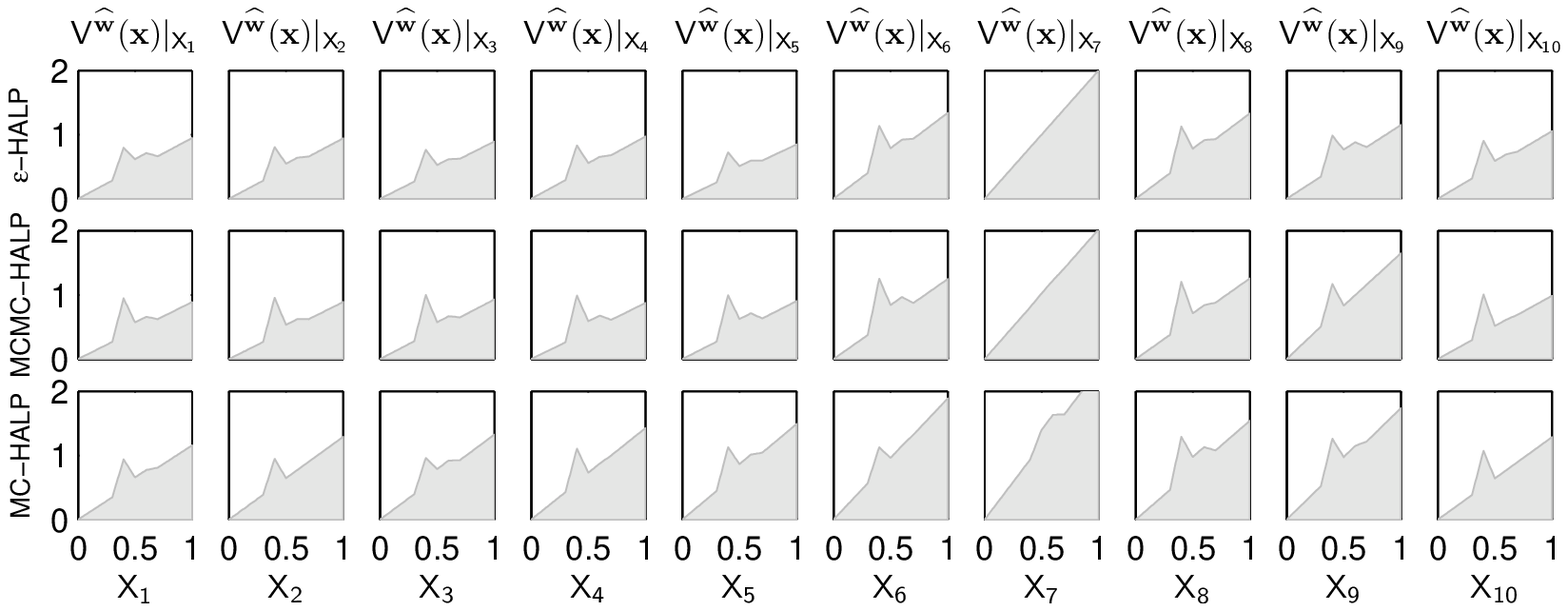}
  \caption{Univariate projections
  $\Vbwhat(\bx) |_{X_j} \! = \sum_{i : X_j = X_i} \what_i f_i(x_i)$
  of approximate value functions $\Vbwhat$ on the 6-ring
  irrigation network problem
  (Figure \ref{fig:irrigation network topologies}a). These
  functions are learned from 40 basis functions
  (Figure \ref{fig:irrigation network model}c) by the $\eps$-HALP,
  MCMC-HALP, and MC-HALP solvers. The solvers are parameterized by
  the resolution of $\eps$-grid ($\eps = 1 / 16$), the number of
  MCMC chains ($N = 250$), and the number of samples ($N = 10^6$).
  Note that the univariate projections $\Vbwhat(\bx) |_{X_j}$ are
  very similar. The proximity of their greedy policies can be
  explained based on this observation.}
  \label{fig:ring value functions}
\end{figure}

In the rest of the section, we illustrate the performance of three
HALP approximations, MC-HALP, $\eps$-HALP, and MCMC-HALP (Section
\ref{sec:HALP constraint space}), on the ring and ring-of-rings
topologies (Figure \ref{fig:irrigation network topologies}) of the
irrigation network problem. The constraints in the MC-HALP
formulation are sampled uniformly at random. This establishes a
baseline for all HALP approximations. The $\eps$-HALP and
MCMC-HALP formulations are generated iteratively by the cutting
plane method. The MCMC oracle $\cO_\mathrm{MCMC}$ is simulated for
500 steps from the initial temperature $c = 0.2$, which leads to a
decreasing cooling schedule from $T_0 = 0.2$ to $T_{500} \approx
0.02$. These parameters are selected empirically to demonstrate
the characteristics of the oracle $\cO_\mathrm{MCMC}$ rather than
to maximize its performance. The value function $\Vstar$ is
approximated by a linear combination of four univariate piecewise
linear basis functions for each channel (Figure
\ref{fig:irrigation network model}c). We assume that our basis
functions are sufficient to derive a one-step lookahead policy
that routes water between the channels if their water levels are
too high or too low (Figure \ref{fig:irrigation network model}b).
We believe that such a policy is close-to-optimal in irrigation
networks. The state relevance density function $\psi(\bx)$ is
uniform. Our experimental results are reported in Figures
\ref{fig:ring and rors results}--\ref{fig:ring value functions}.

Based on the results, we draw the following conclusions. First,
all HALP approximations scale up in the dimensionality of solved
problems. As shown in Figure \ref{fig:ring and rors trends}, the
return of the policies grows linearly in $n$. Moreover, the time
complexity of computing them is polynomial in $n$. Therefore, if a
problem and its approximate solution are structured, we take
advantage of this structure to avoid an exponential blowup in the
computation time. At the same time, the quality of the policies is
not deteriorating with increasing problem size $n$.

Second, the MCMC solver ($N = 250$) achieves the highest objective
values on all solved problems. Higher objective values are
interpreted as closer approximations to the constraint space in
HALP since the solvers operate on relaxed formulations of HALP.
Third, the quality of the MCMC-HALP policies ($N = 250$) surpasses
the MC-HALP policies ($N = 10^6$) while both solvers consume
approximately the same computation time. This result is due to the
informative search for violated constraints in the MCMC-HALP
solver. Fourth, the quality of the MCMC-HALP policies ($N = 250$)
is close to the $\eps$-HALP policies ($\eps = 1 / 16$) although
there is a significant difference between their objective values.
Further analysis shows that the shape of the value functions is
similar (Figure \ref{fig:ring value functions}) and they differ
the most in the weight of the constant basis function $f_0(\bx)
\equiv 1$. Note that increasing $w_0$ does not affect the quality
of a greedy policy for $\Vbw$. However, this trick allows the
satisfaction of the constraint space in HALP (Section
\ref{sec:HALP error bounds}).

Finally, the computation time of the $\eps$-HALP solver is
seriously affected by the topologies of the irrigation networks,
which can be explained as follows. For a small $\eps$ and large
$n$, the time complexity of formulating cost networks for the ring
and ring-of-rings topologies grows by the rates of $\ceils{1 /
\eps + 1}^2$ and $\ceils{1 / \eps + 1}^3$, respectively. Since the
$\eps$-HALP method consumes a significant amount of time by
constructing cost networks, its quadratic (in \mbox{$\ceils{1 /
\eps + 1}$) time} complexity on the ring topology worsens to cubic
(in $\ceils{1 / \eps + 1}$) on the ring-of-rings topology. On the
other hand, a similar cross-topology comparison of the MCMC-HALP
solver shows that its computation times differ only by a
multiplicative factor of 2. This difference is due to the
increased complexity of sampling $p(z_i^\ast \mid \bz_{- i})$,
which results from more complex local dependencies in the
ring-of-rings topology and not its treewidth.

Before we proceed, note that our relaxed formulations (Figure
\ref{fig:ring and rors results}) have significantly less
constraints than their complete sets (Section \ref{sec:cutting
plane method}). For instance, the MC-HALP formulation ($N = 10^6$)
on the 6-ring irrigation network problem is originally established
by $10^6$ randomly sampled constraints. Based on our empirical
results, the constraints can be satisfied greedily by a subset of
400 constraints on average \shortcite{kveton04heuristic}.
Similarly, the oracle $\cO_\mathrm{MCMC}$ in the MCMC-HALP
formulation ($N = 250$) iterates through $250 \times 500 \times
(10 + 10) = 2,500,000$ state-action configurations (Figure
\ref{fig:MCMC-HALP separation oracle}). However, corresponding LP
formulations involve only 700 constraints on average.

\subsection{The Curse of Treewidth}
\label{sec:experiments treewidth}

\begin{figure}[t]
  \centering
  {\small
  \begin{tabular}{|r@{\ }|r@{\ }|r@{\ }|r@{\ }|}
    \multicolumn{4}{c}{\textbf{$\eps$-HALP}} \\ \hline
    $\eps$ & OV & Reward & Time \\ \hline
    $1$ & $30.4$ & $48.3 \pm 3.0$ & $9$ \\
    $1 / 2$ & $42.9$ & $58.7 \pm 3.1$ & $342$ \\
    $1 / 4$ & $49.1$ & $61.9 \pm 3.1$ & $9\ 443$ \\ \hline
  \end{tabular}
  \ \
  \begin{tabular}{|r@{\ }|r@{\ }|r@{\ }|r@{\ }|}
    \multicolumn{4}{c}{\textbf{MCMC}} \\ \hline
    $N$ & OV & Reward & Time \\ \hline
    $10$ & $45.3$ & $43.6 \pm 6.5$ & $83$ \\
    $50$ & $116.2$ & $72.2 \pm 3.6$ & $458$ \\
    $250$ & $118.5$ & $73.2 \pm 3.7$ & $2\ 012$ \\ \hline
  \end{tabular}
  \ \
  \begin{tabular}{|r@{\ }|r@{\ }|r@{\ }|r@{\ }|}
    \multicolumn{4}{c}{\textbf{MC}} \\ \hline
    $N$ & OV & Reward & Time \\ \hline
    $10^2$ & $12.8$ & $56.6 \pm 4.5$ & $< 1$ \\
    $10^4$ & $49.9$ & $53.4 \pm 5.9$ & $19$ \\
    $10^6$ & $71.7$ & $70.3 \pm 3.9$ & $1\ 400$ \\ \hline
  \end{tabular}
  }
  \caption{Comparison of three HALP solvers on the $3 \times 3$
  grid irrigation network problem (Figure
  \ref{fig:irrigation network topologies}). The solvers are
  compared by the objective value of a relaxed HALP (OV), the
  expected discounted reward of a corresponding policy, and
  computation time (in seconds). The $\eps$-HALP, MCMC-HALP, and
  MC-HALP solvers are parameterized by the resolution of
  $\eps$-grid ($\eps$), the number of MCMC chains ($N$), and the
  number of samples ($N$). Note that the quality of policies
  improves with higher grid resolution ($1 / \eps$) and larger
  sample size ($N$). An upper bound on the expected returns is
  87.2.}
  \label{fig:grid results}
\end{figure}

In the ring and ring-of-rings topologies, the treewidth of the
constraint space (in continuous variables) is 2 and 3,
respectively. As a result, the oracle $\cO_\eps$ can perform
variable elimination for a small $\eps$, and the $\eps$-HALP
solver returns close-to-optimal policies. Unfortunately, small
treewidth is atypical in real-world domains. For instance, the
treewidth of a more complex $3 \times 3$ grid irrigation network
(Figure \ref{fig:irrigation network topologies}c) is 6. To perform
variable elimination for $\eps = 1 / 16$, the separation oracle
$\cO_\eps$ requires the space of $\ceils{1 / \eps + 1}^7 \!
\approx \! 2^{28}$, which is at the memory limit of existing PCs.
To analyze the behavior of our separation oracles (Section
\ref{sec:HALP constraint space}) in this setting, we repeat our
experiments from Section \ref{sec:experiments scale-up potential}
on the $3 \times 3$ grid irrigation network.

Based on the results in Figure \ref{fig:grid results}, we conclude
that the time complexity of the $\eps$-HALP solver grows by the
rate of $\ceils{1 / \eps + 1}^7$. Therefore, approximate
constraint space satisfaction (MC-HALP and MCMC-HALP) generates
better results than a combinatorial optimization on an
insufficiently discretized $\eps$-grid ($\eps$-HALP). This
conclusion is parallel to those in large structured optimization
problems with continuous variables. We believe that a combination
of exact and approximate steps delivers the best tradeoff between
the quality and complexity of our solutions (Section
\ref{sec:MCMC-HALP}).

\section{Conclusions}
\label{sec:conclusions}

Development of scalable algorithms for solving real-world decision
problems is a challenging task. In this paper, we presented a
theoretically sound framework that allows for a compact
representation and efficient solutions to hybrid factored MDPs. We
believe that our results can be applied to a variety of
optimization problems in robotics, manufacturing, or financial
mathematics. This work can be extended in several interesting
directions.

First, note that the concept of closed-form solutions to the
expectations terms in HALP is not limited to the choices in
Section \ref{sec:HALP expectation terms}. For instance, we can
show that if $P(x)$ and $f(x)$ are normal densities,
$\E{P(x)}{f(x)}$ has a closed-form solution
\shortcite{kveton06solving}. Therefore, we can directly reason
with normal transition functions instead of approximating them by
a mixture of beta distributions. Similar conclusions are true for
piecewise constant, piecewise linear, and gamma transition and
basis functions. Note that our efficient solutions apply to any
approach to solving hybrid factored MDPs that approximates the
optimal value function by a linear combination of basis functions
(Equation \ref{eq:linear value function}).

Second, the constraint space in HALP (\ref{eq:HALP}) $\Vbw -
\cT^\ast \Vbw \geq 0$ exhibits the same structure as the
constraint space in approximate policy iteration (API)
\shortcite{guestrin01maxnorm,patrascu02greedy} $\maxnorm{\Vbw -
\cT^\ast \Vbw} \leq \eps$, where $\eps$ is a variable subject to
minimization. As a result, our work provides a recipe for solving
API formulations in hybrid state and action domains. In
discrete-state spaces, \shortciteA{patrascu02greedy} and
\shortciteA{guestrin03thesis} showed that API returns better
policies than ALP for the same set of basis functions. Note that
API is more complex than ALP because every step of API involves
satisfying the constraint $\maxnorm{\Vbw - \cT^\ast \Vbw} \leq
\eps$ for some fixed $\eps$.

Third, automatic learning of basis functions seems critical for
the application of HALP to real-world domains.
\shortciteA{patrascu02greedy} analyzed this problem in
discrete-state spaces and proposed a greedy approach to learning
basis functions. \shortciteA{kveton06learning} generalized these
ideas and showed how to learn parametric basis functions in hybrid
spaces. We believe that a combination of the greedy search with a
state space analysis
\shortcite{mahadevan05samuel,mahadevan06value} can yield even
better basis functions.

Finally, we proposed several bounds (Section \ref{sec:HALP error
bounds} and \ref{sec:e-HALP error bounds}) that may explain the
quality of the complete and relaxed HALP formulations. In future,
we plan to empirically evaluate their tightness on a variety of
low-dimensional hybrid optimization problems
\shortcite{bresina02planning,munos02variable} with known optimal
value functions.

\section*{Acknowledgment}

This work was supported in part by National Science Foundation
grants CMS-0416754 and ANI-0325353. The first author was supported
by Andrew Mellon Predoctoral Fellowships for the academic years
2004-06. The first author also recognizes support from Intel
Corporation in the summer 2005 and 2006.

\appendix

\section{Proofs}
\label{sec:proofs}

{\bf Proof of Proposition \ref{prop:HALP lower bound}:} The
Bellman operator $\cT^\ast$ is known to be a contraction mapping.
Based on its monotonicity, for any value function $V$, $V \geq
\cT^\ast V$ implies $V \geq \cT^\ast V \geq \dots \geq \Vstar$.
Since constraints in the HALP formulation (\ref{eq:HALP}) enforce
$\Vbwtilde \geq \cT^\ast \Vbwtilde$, we conclude $\Vbwtilde \geq
\Vstar$. \qed

\bigskip \noindent {\bf Proof of Proposition \ref{prop:HALP L1}:}
Based on Proposition \ref{prop:HALP lower bound}, we note that the
constraint $\Vbw \geq \cT^\ast \Vbw$ guarantees that $\Vbw \geq
\Vstar$. Subsequently, our claim is proved by realizing:
\begin{align*}
  \arg\min_\bw \E{\psi}{\Vbw} =
  \arg\min_\bw \E{\psi}{\Vbw - \Vstar}
\end{align*}
and
\begin{align*}
  \E{\psi}{\Vbw - \Vstar}
  \ & = \ \Eabs{\psi}{\Vbw - \Vstar} \\
  \ & = \ \Eabs{\psi}{\Vstar - \Vbw} \\
  \ & = \ \normw{\Vstar - \Vbw}{1, \psi}.
\end{align*}
The proof generalizes from the discrete-state case
\shortcite{defarias03linear} without any alternations. \qed

\bigskip \noindent {\bf Proof of Theorem
\ref{thm:HALP simple bound}:} Similarly to Theorem 2
\shortcite{defarias03linear}, this claim is proved in three steps.
First, we find a point $\bwbar$ in the feasible region of the HALP
such that $\Vbwbar$ is within $O(\epsilon)$ distance from
$\Vbwstar$, where:
\begin{align*}
  \bwstar \ & = \ \arg\min_\bw \maxnorm{\Vstar - \Vbw} \\
  \epsilon \ & = \ \maxnorm{\Vstar - \Vbwstar}.
\end{align*}
Such a point $\bwbar$ is given by:
\begin{align*}
  \bwbar = \bwstar + \frac{(1 + \gamma) \epsilon}{1 - \gamma} e,
\end{align*}
where $e = (1, 0, \dots, 0)$ is an indicator of the constant basis
function $f_0(\bx) \equiv 1$. This point satisfies all
requirements and its feasibility can be handily verified by
solving:
\begin{align*}
  \Vbwbar - \cT^\ast \Vbwbar
  \ & = \ \Vbwstar + \frac{(1 + \gamma) \epsilon}{1 - \gamma} -
  \left(\cT^\ast \Vbwstar +
  \frac{\gamma (1 + \gamma) \epsilon}{1 - \gamma}\right) \\
  \ & = \ \Vbwstar - \cT^\ast \Vbwstar + (1 + \gamma) \epsilon \\
  \ & \geq \ 0,
\end{align*}
where the last step follows from the inequality:
\begin{align*}
  \maxnorm{\Vbwstar - \cT^\ast \Vbwstar}
  \ & \leq \ \maxnorm{\Vbwstar - \Vstar} +
  \maxnorm{\Vstar - \cT^\ast \Vbwstar} \\
  \ & = \ \maxnorm{\Vstar - \Vbwstar} +
  \maxnorm{\cT^\ast \Vstar - \cT^\ast \Vbwstar} \\
  \ & \leq \ (1 + \gamma) \epsilon.
\end{align*}
Subsequently, we bound the max-norm error of $\Vbwbar$ by using
the triangle inequality:
\begin{align*}
  \maxnorm{\Vstar - \Vbwbar}
  \ & \leq \ \maxnorm{\Vstar - \Vbwstar} +
  \maxnorm{\Vbwstar - \Vbwbar} \\
  \ & = \ \left(1 + \frac{1 + \gamma}{1 - \gamma}\right) \epsilon \\
  \ & = \ \frac{2 \epsilon}{1 - \gamma},
\end{align*}
which yields a bound on the weighted $\cL_1$-norm error of
$\Vbwtilde$:
\begin{align*}
  \normw{\Vstar - \Vbwtilde}{1, \psi}
  \ & \leq \ \normw{\Vstar - \Vbwbar}{1, \psi} \\
  \ & \leq \ \maxnorm{\Vstar - \Vbwbar} \\
  \ & \leq \ \frac{2 \epsilon}{1 - \gamma}.
\end{align*}
The proof generalizes from the discrete-state case
\shortcite{defarias03linear} without any alternations. \qed

\bigskip \noindent {\bf Proof of Theorem \ref{thm:HALP bound}:}
Similarly to Theorem \ref{thm:HALP simple bound}, this claim is
proved in three steps: finding a point $\bwbar$ in the feasible
region of the HALP, bounding the max-norm error of $\Vbwbar$,
which in turn yields a bound on the $\cL_1$-norm error of
$\Vbwtilde$. A comprehensive proof for the discrete-state case was
done by \shortciteA{defarias03linear}. This proof generalizes to
structured state and action spaces with continuous variables. \qed

\bigskip \noindent {\bf Proof of Proposition
\ref{prop:polynomial basis function}:} The proposition is proved
in a sequence of steps:
\begin{align*}
  \E{P(x)}{f(x)}
  \ & = \
  \int_x \betapdf(x \mid \alpha, \beta) x^n (1 - x)^m \ud x \\
  \ & = \
  \int_x \frac{\Gamma(\alpha + \beta)}{\Gamma(\alpha) \Gamma(\beta)}
  x^{\alpha - 1} (1 - x)^{\beta - 1} x^n (1 - x)^m \ud x \\
  \ & = \
  \frac{\Gamma(\alpha + \beta)}{\Gamma(\alpha) \Gamma(\beta)}
  \int_x x^{\alpha + n - 1} (1 - x)^{\beta + m - 1} \ud x \\
  \ & = \
  \frac{\Gamma(\alpha + \beta)}{\Gamma(\alpha) \Gamma(\beta)}
  \frac{\Gamma(\alpha + n) \Gamma(\beta + m)}
  {\Gamma(\alpha + \beta + n + m)}
  \int_x \frac{\Gamma(\alpha + \beta + n + m)}
  {\Gamma(\alpha + n) \Gamma(\beta + m)}
  x^{\alpha + n - 1} (1 - x)^{\beta + m - 1} \ud x \\
  \ & = \
  \frac{\Gamma(\alpha + \beta)}{\Gamma(\alpha) \Gamma(\beta)}
  \frac{\Gamma(\alpha + n) \Gamma(\beta + m)}
  {\Gamma(\alpha + \beta + n + m)}
  \underbrace{\int_x \betapdf(x \mid \alpha + n, \beta + m) \ud x}_1.
\end{align*}
Since integration is a distributive operation, our claim
straightforwardly generalizes to the mixture of beta distributions
$P(x)$. \qed

\bigskip \noindent {\bf Proof of Proposition
\ref{prop:piecewise linear basis function}:} The proposition is
proved in a sequence of steps:
\begin{align*}
  \E{P(x)}{f(x)}
  \ & = \
  \int_x \betapdf(x \mid \alpha, \beta)
  \sum_i \I{[l_i, r_i]}{x} (a_i x + b_i) \ud x \\
  \ & = \
  \sum_i \int_{l_i}^{r_i} \betapdf(x \mid \alpha, \beta)
  (a_i x + b_i) \ud x \\
  \ & = \
  \sum_i \left[a_i
  \int_{l_i}^{r_i} \betapdf(x \mid \alpha, \beta) x \ud x +
  b_i \int_{l_i}^{r_i} \betapdf(x \mid \alpha, \beta) \ud x\right] \\
  \ & = \
  \sum_i \left[a_i \frac{\alpha}{\alpha + \beta}
  \int_{l_i}^{r_i} \betapdf(x \mid \alpha + 1, \beta) \ud x +
  b_i \int_{l_i}^{r_i} \betapdf(x \mid \alpha, \beta) \ud x\right] \\
  \ & = \
  \sum_i \left[a_i \frac{\alpha}{\alpha + \beta}
  (F^+(r_i) - F^+(l_i)) + b_i (F(r_i) - F(l_i))\right].
\end{align*}
Since integration is a distributive operation, our claim
straightforwardly generalizes to the mixture of beta distributions
$P(x)$. \qed

\bigskip \noindent {\bf Proof of Proposition
\ref{prop:HALP to relaxed HALP bound}:} This claim is proved in
three steps. First, we construct a point $\bwbar$ in the feasible
region of the HALP such that $\Vbwbar$ is within $O(\delta)$
distance from $\Vbwhat$. Such a point $\bwbar$ is given by:
\begin{align*}
  \bwbar = \bwhat + \frac{\delta}{1 - \gamma} e,
\end{align*}
where $e = (1, 0, \dots, 0)$ is an indicator of the constant basis
function $f_0(\bx) \equiv 1$. This point satisfies all
requirements and its feasibility can be handily verified by
solving:
\begin{align*}
  \Vbwbar - \cT^\ast \Vbwbar
  \ & = \ \Vbwhat + \frac{\delta}{1 - \gamma} -
  \left(\cT^\ast \Vbwhat + \frac{\gamma \delta}{1 - \gamma}\right) \\
  \ & = \ \Vbwhat - \cT^\ast \Vbwhat + \delta \\
  \ & \geq \ 0,
\end{align*}
where the inequality $\Vbwhat - \cT^\ast \Vbwhat \geq - \delta$
holds from the $\delta$-infeasibility of $\bwhat$. Since the
optimal solution $\bwtilde$ is feasible in the relaxed HALP, we
conclude $\E{\psi}{\Vbwhat} \leq \E{\psi}{\Vbwtilde}$.
Subsequently, this inequality yields a bound on the weighted
$\cL_1$-norm error of $\Vbwbar$:
\begin{align*}
  \normw{\Vstar - \Vbwbar}{1, \psi}
  \ & = \ \Eabs{\psi}{\Vbwhat + \frac{\delta}{1 - \gamma} - \Vstar} \\
  \ & = \ \E{\psi}{\Vbwhat} + \frac{\delta}{1 - \gamma} -
  \E{\psi}{\Vstar} \\
  \ & \leq \ \E{\psi}{\Vbwtilde} + \frac{\delta}{1 - \gamma} -
  \E{\psi}{\Vstar} \\
  \ & = \ \normw{\Vstar - \Vbwtilde}{1, \psi} +
  \frac{\delta}{1 - \gamma}.
\end{align*}
Finally, we combine this result with the triangle inequality:
\begin{align*}
  \normw{\Vstar - \Vbwhat}{1, \psi}
  \ & \leq \ \normw{\Vstar - \Vbwbar}{1, \psi} +
  \normw{\Vbwbar - \Vbwhat}{1, \psi} \\
  \ & \leq \ \normw{\Vstar - \Vbwtilde}{1, \psi} +
  \frac{2 \delta}{1 - \gamma},
\end{align*}
which leads to a bound on the weighted $\cL_1$-norm error of
$\Vbwhat$. \qed

\bibliographystyle{theapa}
\bibliography{References}

\end{document}